\documentclass{article}
\usepackage{amsfonts}
\usepackage{amsmath}
\usepackage{amsthm}
\usepackage{color}
\usepackage{graphicx}
\usepackage{wasysym}
\usepackage{fullpage}
\usepackage{setspace}
\usepackage{charter}

\usepackage{hyperref} 
\hypersetup{backref, colorlinks=true, citecolor=blue, linkcolor=blue}
\newcommand{\rref}[1]{\hyperref[#1]{\ref*{#1}}}

\newtheorem{theorem}{Theorem}[section]
\newtheorem{lemma}[theorem]{Lemma}

\newtheorem{corollary}[theorem]{Corollary}

\theoremstyle{definition}
\newtheorem{definition}[theorem]{Definition}

\newcommand{\R}{\mathbb R}

\usepackage[round, authoryear]{natbib}
\begin{document}

\title{Stability of Density-Based Clustering
\author{Alessandro Rinaldo\thanks{Email: {\tt arinaldo@stat.cmu.edu}}\\ 
Department of Statistics\\
Carnegie Mellon University\\
Pittsburgh, PA, 15213-3890 USA  
\and
Aarti Singh\thanks{Email: {\tt aarti@cs.cmu.edu}}\\
Machine Learning Department\\
Carnegie Mellon University\\
Pittsburgh, PA, 15213-3890 USA  
\and
Rebecca Nugent\thanks{Email: {\tt rnugent@stat.cmu.edu}}\\
Department of Statistics\\
Carnegie Mellon University\\
Pittsburgh, PA, 15213-3890 USA
\and
Larry Wasserman\thanks{Email: {\tt larry@stat.cmu.edu}}\\
Department of Statistics and\\ 
Machine Learning Department\\
Carnegie Mellon University\\
Pittsburgh, PA, 15213-3890 USA  
}
\date{}
}

\maketitle

\begin{abstract}%
High density clusters can be characterized by the connected 
components of a level set $L(\lambda) = \{x:\ p(x)>\lambda\}$ of the underlying 
probability density function $p$ generating the data, at some 
appropriate level $\lambda\geq 0$. The complete hierarchical clustering 
can be characterized by a cluster tree 
${\cal T}= \bigcup_{\lambda} L(\lambda)$. 
In this paper, we study the behavior of a density level set estimate  $\widehat L(\lambda)$
and cluster tree estimate $\widehat{\cal{T}}$ based on a kernel density estimator
with kernel bandwidth $h$. We define two notions of instability to measure the variability 
of $\widehat L(\lambda)$ and $\widehat{\cal{T}}$ as a function of $h$, and investigate the theoretical properties
of these instability measures.
\end{abstract}



\tableofcontents

\section{Introduction}
\label{section::intro}

A common approach to identifying high density clusters is 
based on using level sets of the density function
(\cite{hartigan}, \cite{rigollet}).
Let $X_1,\ldots, X_n$ be a random sample
from a distribution $P$ on $\mathbb{R}^d$
with density $p$.
For $\lambda > 0$ define the level set
$L(\lambda) = \{x:\ p(x)> \lambda\}$.
Assume that $L(\lambda)$
can be decomposed into disjoint, connected sets:
$L(\lambda) = \bigcup_{j=1}^{N(\lambda)} C_j$.
We refer to 
${\cal C}_\lambda = \{C_1, \ldots, C_{N(\lambda)}\}$
as the {\em density clusters} at level $\lambda$.
We call the collection of clusters
\begin{equation}
{\cal T} = \bigcup_{\lambda \geq 0} {\cal C}_\lambda
\end{equation}
the {\em cluster tree}.
Note that ${\cal T}$ does indeed have a tree structure:
if $A,B\in {\cal T}$ then either,
$A\subset B$, or $B\subset A$ or $A\cap B= \emptyset$.
The tree summarizes the cluster structure of
the distribution; see \cite{nugent}.

\begin{figure}[!ht]
\begin{center}
\includegraphics[width=8cm,height=8cm]{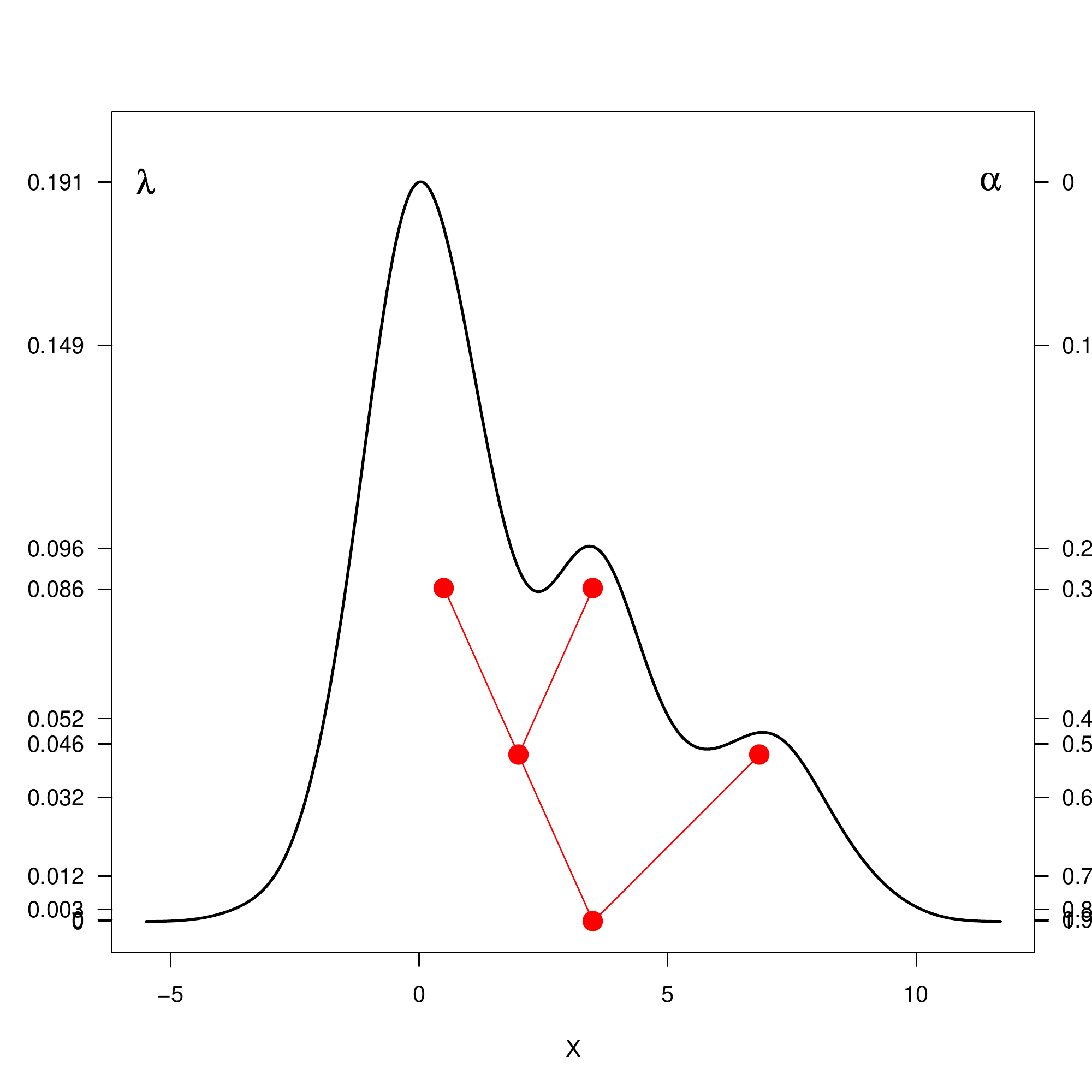}
\vspace{-4mm}
\caption{The cluster tree for a density 
estimate of a sample from the mixture $(4/7)N(0,1)+(2/7)N(3.5,1)+(1/7)N(7,1)$; 
the tree is indexed by both $\lambda$ (left) and $\alpha$ (right).}
\label{fig::intro}
\end{center}
\end{figure}

It is also possible to index the level sets by 
probability content. For $0 < \alpha < 1$, define the level set
$M(\alpha) = L(\lambda_\alpha)$ where
\begin{equation}
\lambda_\alpha = \sup \{\lambda:\ P(L(\lambda)) \geq \alpha\}.
\end{equation}
If the density does not contain any jumps or flat parts, then there is
a one-to-one correspondence between the level sets indexed by the
density level and the probability content.  The cluster tree obtained
from the clusters of $M(\alpha)$ for $0\leq \alpha \leq 1$ is
equivalent to ${\cal T}$.  Relabeling the tree in terms of $\alpha$
may be convenient because $\alpha$ is more interpretable than
$\lambda$, but the tree is the same.  Figure \ref{fig::intro} shows the cluster tree
for a density estimate of a mixture of three normals (using a
reference rule bandwidth).  The cluster tree's two splits and
subsequent three leaves correspond to the density estimate's modes.
The tree is also indexed by $\lambda$, the density estimate's height,
on the left and $\alpha$, the probability content, on the right.  For
example, the second split corresponds to $\lambda = 0.086$ and 
$\alpha = 0.257$.  We note here that determining the true clusters for even
this seemingly simple univariate distribution is not trivial for all
$\lambda$; in particular, values of $\lambda$ near $0.04$ and  $0.09$ will give
ambiguous results.

In this paper we study some properties of clusters defined by density level sets and cluster trees.
In particular, we consider their estimators based on a kernel density estimate
and show how the bandwidth $h$ of the kernel affects the risk of these estimators.
Then we investigate the notion of stability for density-based clustering. Specifically, we propose
two measures of instability.
The first, denoted by $\Xi_n(h)$, measures the instability of
a given level set.
The second, denoted by $\Gamma_n(h)$,
is a more global measure of instability.

\begin{figure}[!ht]
\begin{center}
\includegraphics[width=9cm,height=11cm]{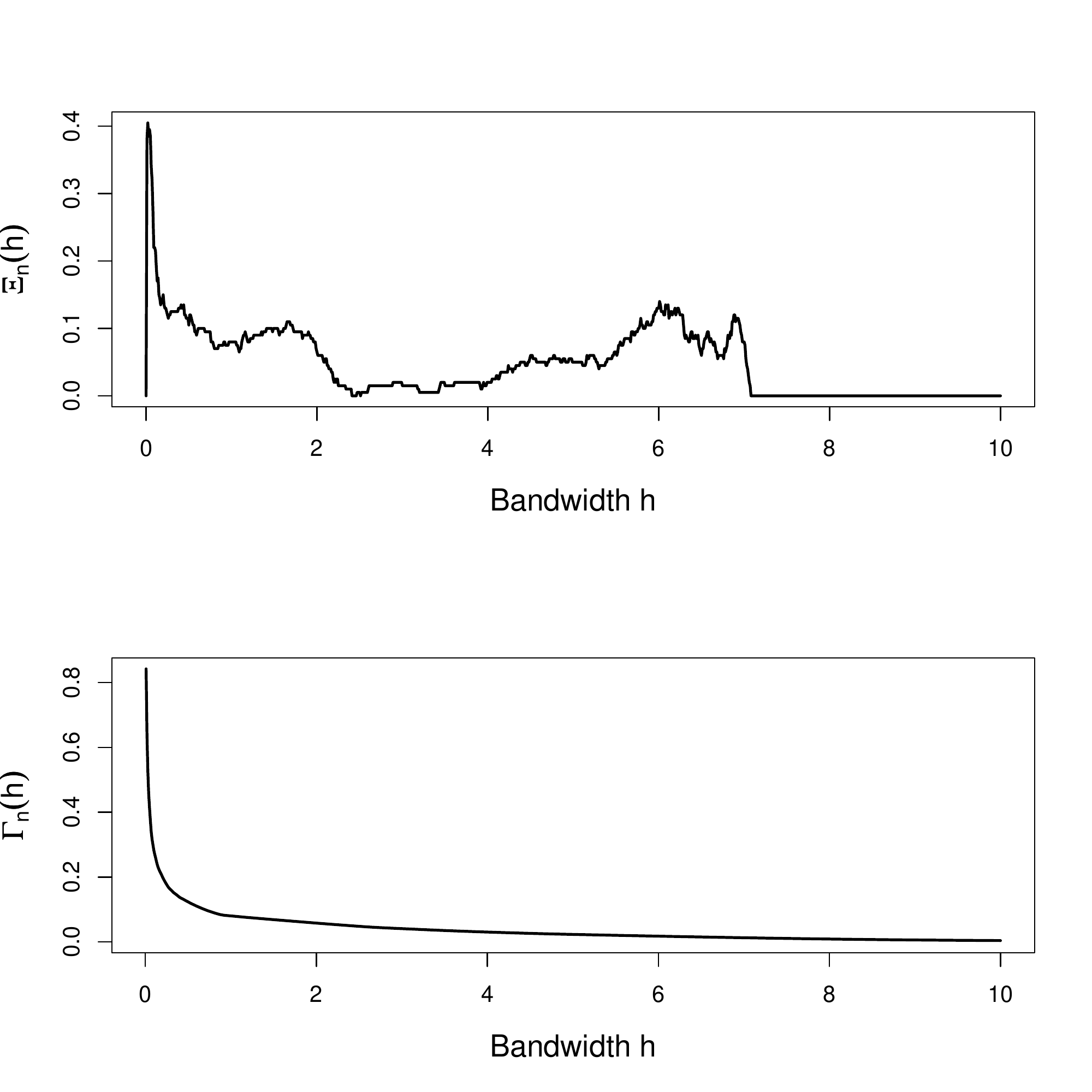}
\vspace{-4mm}
\caption{Plots of the fixed-$\lambda$ instability (top) $\Xi_n(h)$ for $\lambda = 0.09$ and of the total variation instability $\Gamma_n(h)$ (bottom) for the mixture distribution in Figure \ref{fig::intro} as functions of the bandwidth $h$.}
\label{fig::simple}
\end{center}
\end{figure}

Investigation of the stability properties
of density clusters is the main focus of the paper.
Stability has become an increasingly popular tool
for choosing tuning parameters in clustering;
see 
\cite{lux}, \cite{lange},
\cite{bendavid},
\cite{benhur},
\cite{carlsson},
\cite{Meinshausen},
\cite{Buhmann},
and \cite{rinaldo}.
The basic idea is 
this: clustering procedures inevitably depend on one or more tuning parameters.
If we choose a good
value of the tuning parameter,
then we expect that the clusters
from different subsets
of the data should be similar.
While this idea sounds simple,
the reality is rather complex.
Figure \ref{fig::simple} shows a plot of $\Xi_n$ and $\Gamma_n$
for our example.
We see that $\Xi_n(h)$ is a complicated function of $h$ while
$\Gamma_n(h)$ is much simpler.
Our results will explain this behavior.

In Section \ref{section::prelim} we state some notation, assumptions on the density, and
discuss the kernel density estimate.
In Section
\ref{section::estimate} we construct plug-in estimates $\widehat L(\lambda)$ of the level
set $L(\lambda)$, $\hat{\cal T}$ of the cluster tree ${\cal T}$, and $\widehat M(\alpha)$ of the level
set indexed by probability content $M(\alpha)$.  
In Section \ref{section::stability} we
define and study a notion of the stability of $\widehat L(\lambda)$ and extend it to
$\hat{\cal T}$. We also 
consider an alternative version of our results when
the level sets are indexed by probability content.
We then describe another notion of stability of cluster trees based on total variation that leads to
a constructive procedure for selecting the kernel bandwidth.
In Section \ref{section::examples} we consider some numerical examples.
Section \ref{section::discussion} contains a discussion of the
results and the 
proofs are in Section \ref{section::proofs}.
Throughout, we use symbols like $c,c_1,c_2,\ldots , C,C_1,
C_2,\ldots,$ to denote various positive constants whose value can
change in different expressions.


\section{Preliminaries}
\label{section::prelim}

\subsection{Notation}
For $x \in \mathbb{R}^d$, let $\| x \|$ denote its euclidean norm.
Let $B(x,\epsilon) = \{y:\ ||x-y||\leq \epsilon\} \subset \mathbb{R}^d$
denote a ball centered at $x$ with radius $\epsilon$.
For any set $A \subset \mathbb{R}^d$ and any $\epsilon \geq 0$, let
$A\oplus\epsilon = \bigcup_{x\in A} B(x,\epsilon)$.
Let $v_d = \frac{ \pi^{d/2}}{\Gamma\left(\frac{d}{2} +1\right)}$
denote the volume of the unit ball.
The Hausdorff distance between two sets $A$ and $B$ is
\begin{equation}
d_\infty(A,B) = \inf\{\epsilon:\ A\subset (B\oplus\epsilon)\ \ {\rm and}\ \ B \subset (A\oplus\epsilon)\}.
\end{equation}
Finally, let $A^c$ denote the complement of set $A$ and let
$$
A \Delta B = (A \cap B^c) \bigcup (A^c \cap B)
$$
denote the symmetric set difference.

We will be considering samples of $n$ independent and identically distributed random vectors from an unknown distribution $P$ on $\mathbb{R}^d$. If $X$ and $Y$ are such samples, we will denote with $\mathbb{P}_{X,Y}$ the probability measures associated to them and with $\mathbb{E}_{X,Y}$ the corresponding expectation operator. Thus, if $\mathcal{A}$ is an event depending on $X$ and $Y$, we will write $\mathbb{P}_{X,Y}(\mathcal{A})$ for its probability.
Finally, for a sample $X = (X_1,\dots ,X_n)$, we will denote with $\hat{P}_X$ the empirical measure associated with it; explicitly, for any measurable set $A \subset \mathbb{R}^d$,
\[
\widehat P_X(A) = \frac1{n}\sum^n_{i=1}I(X_i \in A).
\]
\subsection{Assumptions}

We will use the following assumptions on the density $p$:

\begin{itemize}
\item[(A0)] \textit{Compact Support -} The support $S$ of $p$ is compact.
\item[(A1)] \textit{Lipschitz Density -}
Assume that 
\begin{equation}
p\in \Sigma(A) \equiv \Biggl\{p:\ |p(x) - p(y)| \leq A ||x-y||,\ \ {\rm for\ all\ }x,y\in S\Biggr\}
\end{equation}
for some $A>0$. 
\item[(A2)] \textit{Local density regularity -} 
For a given density level of interest $\lambda$, there exist constants 
$0 < \kappa_1 \leq \kappa_2 < \infty$ and $0< \epsilon_0$ such that, for all $\epsilon < \epsilon_0$,
\begin{equation}
\kappa_1 \epsilon \leq P(\{ x \colon |p(x) - \lambda| \leq \epsilon\}) \leq \kappa_2 \epsilon.
\end{equation}
It is possible to formulate this condition more generally in terms of
power of $\epsilon$, that is $\epsilon^a$.
But, as argued in \cite{rinaldo}, the above statement typically holds with $a = 1$ for almost all $\lambda$.
\end{itemize}


Some of the results will only require a subset of these assumptions, and this will be explicitly mentioned in the statement of the result.
Assumptions (A1) and (A2) characterize the density regularity - (A1) implies that the density cannot change
drastically anywhere, while (A2) implies that the density cannot be too flat or steep locally 
around the level set. Also notice that (A0) and (A1) together imply that the density $p$ is bounded
by some positive constant $p_{\max} < \infty$.
These assumptions are stronger than
necessary, but they simplify the proofs. Notice in particular, that assumption (A2) can rule out the case of sharp clusters, in which $S$ is the disjoint union of a finite number of compact sets over which $p$ is bounded from below by a positive constant.

\subsection{Estimating the Density}\label{sec:est.density}

To estimate the density $p$ based on the i.i.d. sample $X = (X_1,\ldots,X_n)$,
we use the kernel density estimator
\begin{equation}
\hat{p}_{h,X}(u) = \frac{1}{n}\sum_{i=1}^n \frac{1}{h^d}
K\left(\frac{||u-X_i||}{h}\right), \quad u \in \mathbb{R}^d,
\end{equation}
where $K$ is a symmetric kernel with compact support
and $h>0$ is the bandwidth.
Let
$p_{h}(u) = \mathbb{E}_X[\hat p_{h,X}(u)]$.
Note that
$p_h$ is the density of
$$
P_h = P \oplus \mathbb{K}_h
$$
where $\oplus$ denotes convolution and $\mathbb{K}_h$ denotes the probability measure of 
a random variable with density $K_h(t) = h^{-d} K(||t||/h)$.

We impose the following conditions on $p_h$: 
\begin{itemize}
\item[(B0)] The support of $P_h$ is compact for all $h \geq 0$.
\item[(B1)] $p_h\in\Sigma(A)$.
\item[(B2)] 
For a given density level $\lambda$, there exist positive constants $\kappa_1 
\leq \kappa_2$, $\epsilon_0$ and $H$ bounded away from $0$ and $\infty$, such that, for all $0\leq \epsilon < \epsilon_0$,
$$
\kappa_1 \epsilon \leq \inf_{0\leq h\leq H} P\left(\{ x \colon |p_h(x)-\lambda|\leq \epsilon\}\right) \leq
\sup_{0\leq h\leq H} P(\{x \colon |p_h(X)-\lambda|\leq \epsilon \}) \leq \kappa_2 \epsilon.
$$
\item[(B3)] For a given $\alpha$, there exist positive constants
$\kappa_3$, $\eta_0$ and $H$ bounded away from $0$ and $\infty$, such that, for all $0\leq \eta < \eta_0$,
\[
\sup_{0\leq h\leq H}d_\infty(M_h(\alpha),M_h(\alpha+\eta)) \leq \kappa_3 |\eta|.
\]
Here $M_h(\alpha) = \{u: p_h(u) > \lambda_\alpha\}$.
\end{itemize}

We remark that condition (B0) follows from (A0) and the compactness of the kernel, while (B1) follows directly from (A1) (for a formal argument, see the end of the proof of Lemma \ref{lem:alpha}). We state them as assumptions for clarity.

The more stringent conditions (B2) and (B3) are used only for some specific results from Section \ref{sec:fixed.lambda} and Section \ref{sec::alpha}, respectively. This will be explicitly mentioned in the statement of such  results. 
In particular, condition (B2) is needed in order to explicitly state the behavior of the instability measure we 
define below. We believe this assumption follows from Condition (A2) on the true density $p$
and using kernels with compact support, 
however for technical convenience we state it as an assumption. 
This assumption 
holds for all density levels that are not too close to a local maxima or minima of the density.
Assumption (B3) characterizes the regularity of the level sets of $p_h$ and essentially states that the boundary
of these level sets is well-behaved and not space-filling. 

Our analysis depends crucially on the quantity $\|  \hat{p}_{h,X} - p_h  \|_\infty = 
\sup_{u \in \mathbb{R}^d} |\hat{p}_{h,X}(u) - p_h(u)|$, for which we use a probabilistic 
upper bound that follows from the arguments in \cite{gine},
under the following assumption on the kernel $K$.
\begin{itemize}
\item[(VC)] The class of functions 
$$
\mathcal{F}_h = \left\{K\left(\frac{x-\cdot}{h}\right), x\in\R^d, h>0\right\}
$$
satisfies, for some positive number $A$ and $v$
$$
\sup_P N({\cal F}_h, L_2(P), \epsilon\|F\|_{L_2(P)}) \leq \left(\frac{A}{\epsilon}\right)^v,
$$
where $N(T; d; \epsilon)$ denotes the $\epsilon$-covering number of the metric space $(T, d)$, $F$ is the envelope function
of ${\cal F}$ and the supremum is taken over the set of all probability measures $P$ on $\R^d$. The quantities $A$ and
$v$ are called the VC characteristics of ${\cal F}$.
\end{itemize}
Assumption (VC) holds for a large class of kernels, including, any compactly supported polynomial kernel
and the Gaussian kernel. 
The lemma below follows from
\cite{gine} \citep[see also][]{rinaldo}.

\begin{lemma}
\label{prop::density}
Assume that the kernel satisfies the VC property, and that 
\begin{equation}\label{eq:p}
\sup_{t \in \mathbb{R}^d} \sup_{h > 0} \int_{\mathbb{R}^d} K_h^2(t - x) d P(x) < B < \infty.
\end{equation}
There exist positive constants $K_1$, $K_2$ and $C$, which depends on $B$ and the VC characteristic of $K$ such that the following hold:
\begin{enumerate}
\item For every $\epsilon > 0$ and $h>0$, there exists $n(\epsilon,h)$ such that, for all $n \geq n(\epsilon,h)$
\begin{equation}
\label{eq:gine_prob}
\mathbb{P}_X \left( \| \widehat{p}_{h,X} - p_h \|_\infty >  \epsilon  \right) \leq K_1 e^{-K_2 n \epsilon^2 h^d}.
\end{equation}
\item Let $h_n \rightarrow 0$ as $n \rightarrow \infty$ in such a way that 
\begin{equation}\label{eq:hn}
\frac{n h^d_n}{\log n} \rightarrow \infty. 
\end{equation}
Then, there exist a constant $K_3$ and a number $n_0$ such that, setting $\epsilon_n = \sqrt{\frac{K_3 \log n}{n h_n^d}}$, 
\begin{equation}
\label{eq:gine}
\mathbb{P}_X \left( \| \widehat{p}_{h_n,X} - p_{h_n} \|_\infty >  \epsilon_n  \right) \leq \frac{1}{n}
\end{equation}
for all $n \geq n_0$.
\end{enumerate}
The numbers $n(\epsilon,h)$ and $n_0$ depend also on the VC characteristic of $K$ and on $B$. Furthermore, $n(\epsilon,h)$ is decreasing in both $\epsilon$ and $h$.
\end{lemma}

This result requires virtually no assumptions on $p$ and only minimal assumptions about the kernel,
which are satisfied for all the usual kernels.

The constrain in equation \eqref{eq:hn}, which in general cannot be dispensed with, has a subtle but important implication for our later results on instability. In fact, it implies that the bandwidth parameter $h_n$ is only allowed to vanish at a slower rate than $\left( \frac{\log n}{n} \right)^{1/d}$. As a result, our measures of instability defined in Sections \ref{sec:fixed.lambda} and \ref{sec::alpha} can be reliably estimated for values of the bandwidth $h >> \left( \frac{\log n}{n} \right)^{1/d}$. Indeed, the threshold value $\left( \frac{\log n}{n} \right)^{1/d}$ is of the same order of magnitude of the minimal spacing among the points in a sample of size $n$ form $P$. See \cite{deheuvels} and, in particular, Lemma \ref{lemma:xi_smallh} below.


\section{Estimating the level set and cluster tree}
\label{section::estimate}

For a given density level $\lambda$ and kernel bandwidth $h$, the estimated level set is
$\hat{L}_{h,X}(\lambda) = \{x:\ \hat{p}_{h,X}(x) > \lambda\}$.
The clusters (connected components) of
$\hat{L}_{h,X}(\lambda)$ are denoted by
$\hat{\cal C}_{h,\lambda}$ and the estimated cluster tree is
\begin{equation}
\hat{\cal T}_h = \bigcup_{\lambda \geq 0}\hat{\cal C}_{h,\lambda}.
\end{equation}

\subsection{Fixed $\lambda$}

We measure the quality of $\hat{L}_{h,X}(\lambda)$ 
as an estimator of $L(\lambda)$ 
using
the following loss function:
\begin{equation}
{\cal L}(h,X,\lambda) = \int_{L(\lambda) \Delta \hat L_{h,X}(\lambda)} p(u) du 
\end{equation}
where we recall that
$\Delta$ denotes symmetric set difference. 
The performance of plug-in estimators of density level sets 
has been studied earlier, but we state the results here in a form that provides insights into the performance of instability measures
proposed in the next section.

\begin{theorem}
\label{thm:risk.lambda}
Assume that the density $p$ satisfies the conditions (A0) and (A1) and that the 
kernel $K$ satisfies $\int K(z)dz = 1$ and $\int \|z\|K(z)dz< D$.
For any sequence $h_n = \omega((\log n/n)^{1/d})$, let
\[
\epsilon_n = \sqrt{\frac{K_3 \log n}{n h_n^d}}
\]
and
\[
r_{h_n,\epsilon_n,\lambda} = P\left( \{ u \colon |p(u) - \lambda| < A D h_n +  \epsilon_n\} \right).
\]
Then, for all large $n$,
\[
\mathbb{P}_X \left( {\cal L}(h_n,X,\lambda) \leq r_{h_n,\epsilon_n,\lambda} \right) \geq 1 - \frac{1}{n}.
\]
If the assumption (A2) holds for density level $\lambda$, then for all large $n$,
\[
\mathbb{P}_X \Bigl( {\cal L}(h_n,X,\lambda) \leq \kappa_2(A D h_n + \epsilon_n) \Bigr) \geq 1 - \frac{1}{n}.
\]
\end{theorem}

The following corollary characterizes the optimal scaling of the bandwidth parameter $h_n$
that balances the approximation and estimation errors.

\begin{corollary}
\label{cor::opt_h}
The value of $h$ that minimizes the bound on ${\cal L}$
is 
\begin{equation}
h_n^* = c \left(\frac{n}{\log n}\right)^{-\frac1{d+2}}
\end{equation}
where $c>0$ is an appropriate constant.
\end{corollary}

\subsection{Fixed $\alpha$}
\label{sec::alpha}

Often it is more natural to define the high-density clusters or level sets by the probability mass 
contained in the high-density region, instead of the density level. 
The level set estimator indexed by the probability content $\alpha \in (0,1)$ is given as
$$
\widehat M_{h,X}(\alpha) = \widehat L_{h,X}(\widehat \lambda_{h,\alpha,X})
$$
where
\begin{equation}
\label{eq:hat.lambda}
\widehat\lambda_{h,\alpha,X} = \sup\Biggl\{\lambda :\ 
\hat{P}_X (\{u:\ \hat{p}_{h,X}(u) > \lambda\}) \geq \alpha\Biggr\},
\end{equation}
$\hat{p}_{h,X}$ is the kernel density 
estimate computed using the data $X$ with bandwidth $h$.
This estimator was 
studied by \cite{cadre}, though using different 
techniques and in different settings than ours.

Let $\alpha \in (0,1)$ be fixed and define
$$
\lambda_{h,\alpha} = \sup\{\lambda:\ P(p_h(X) > \lambda)\geq \alpha\}.
$$
We first show that the deviation $|\lambda_{h,\alpha} - \lambda_\alpha|$
is of order $h$, uniformly over $\alpha$, under the very general assumption that
the true density $p$ is Lipschitz.

\begin{lemma}
\label{lemma::lam_dev}
Assume the true density $p$ satisfies the conditions (A0) and (A1). Then, for any $h>0$,
\begin{equation}
\sup_{\alpha \in (0,1)} |\lambda_{h,\alpha}- \lambda_\alpha| \leq A D h,
\end{equation}
where $D = \int_{\mathbb{R}^d} \| z \| K(z)  dz$.
\end{lemma}

\noindent\textbf{Remark:} 
More generally, if $p$ is assumed to be H\"{o}lder continuous with parameter $\beta$ then, under additional mild 
integrability conditions on $K$, it can be shown that $|\lambda_{h,\alpha}- \lambda_\alpha| = O(h^\beta)$, uniformly in $\alpha$.\\

\noindent The following lemma bounds the deviation of $|\hat \lambda_{h,\alpha,X} - \lambda_{h,\alpha}|$. 


\begin{lemma}
\label{lem:alpha}
Assume that the true density satisfies (A0)-(A1) and the density level sets of $p_h$ corresponding to probability content $\alpha$ 
satisfy (B3). Then, for any $0< h \leq H$, any $\epsilon < \eta_0 - 1/n$, and for all large $n$,
\begin{equation}\label{eq:alpha.conc}
\mathbb{P}_X \left(
|\hat\lambda_{h,\alpha,X}- \lambda_{h,\alpha}| \geq 
\epsilon (A\kappa_3 + 1) +A\kappa_3/n \right) \leq K_1 e^{-K_2 n h^d \epsilon^2} + 8 n e^{-n \epsilon^2/32},
\end{equation}
where $A$ is the Lipschitz constant and $\kappa_3$ is the constant in (B3).
\end{lemma}

Using Lemma~\ref{lemma::lam_dev} and Lemma~\ref{lem:alpha}, we immediately obtain the following bound on the 
deviation of the estimated level 
$\widehat \lambda_{h,\alpha,X}$ from the true density level $\lambda_\alpha$ corresponding to probability content $\alpha$.
\begin{corollary}\label{lemma::lambda}
Under the same conditions of Lemma \ref{lem:alpha},
\[
\mathbb{P}_X\left(|\hat\lambda_{h,\alpha,X}- \lambda_{\alpha}| \geq ADh + \epsilon(A\kappa_3+1) +A\kappa_3/n \right) 
\leq K_1 e^{-K_2 n h^d \epsilon^2} + 8ne^{-n\epsilon^2/32}.
\]
\end{corollary}


We now study the performance of the level set estimator indexed by probability content
using the following loss function
$$
{\cal L}(h,X,\alpha) =  P(M(\alpha) \Delta \hat M_{h,X}(\alpha))=
 \int_{M(\alpha) \Delta \hat M_{h,X}(\alpha)} p(u) du.
$$

\begin{theorem}
\label{thm:risk.alpha}
Assume that the density $p$ satisfies conditions the conditions (A0) and (A1) and the level set of $p_h$ 
indexed by probability content $\alpha$ satisfies (B3). For any sequence $ h_n = \omega((\log n/n)^{1/d})$,
let 
\[
\epsilon_n = \sqrt{\frac{K_3 \log n}{n h_n^d}}.
\]
Set
\[
C_{1,n} = A D h_n + \epsilon_n, \quad C_{2,n} = AD h_n + (A\kappa_3 + 1) \epsilon_n +A\kappa_3/n.
\]
and
\[
r_{h_n,\epsilon_n,\alpha} = P \left( \{ u \colon |p(u) - \lambda_\alpha| \leq C_{1,n} + C_{2,n}\}\right).
\]
Then, for $h_n = \omega((\log n/n)^{1/d})$
and $h_n \leq H$, we have for all large $n$, 
\[
\mathbb{P}_X(\mathcal{L}(h_n,X,\alpha) \leq r_{h_n,\epsilon_n,\alpha})  \geq 1 - \frac{2}{n}.
\]
In particular, if the assumption (A2) also holds for density level $\lambda_\alpha$, then for all large $n$, 
$$
\mathbb{P}_X\left(\mathcal{L}(h_n,X,\alpha) \leq \kappa_2(C_{1,n}+ C_{2,n})\right)  \geq 1 - \frac{2}{n}.
$$
\end{theorem}

\begin{corollary}
The value of $h$ that minimizes the upper bound on ${\cal L}$ is
\begin{equation}
h^*_{n,\alpha} = c \left(\frac{n}{\log n}\right)^{-\frac1{d+2}}
\end{equation}
where $c>0$ is a constant.
\end{corollary}

\section{Stability}
\label{section::stability}

The loss ${\cal L}$ is a useful theoretical measure of clustering accuracy.
Balancing the terms in the upper bound on the loss gives an indication of the
optimal scaling behavior of $h$.
But estimating the loss is difficult and the value of the 
constant $c$ in the expression for $h_n^*$ is unknown.
Thus, in practice, we need an alternate method to determine $h$.
Instead of minimizing the loss, we consider using the 
stability of $\hat L_{h,X}(\lambda)$ and $\hat{\cal T}_h$
to choose $h$.
As we discussed in
the introduction, stability ideas have been used for clustering before.
But the behavaior of stability measures can be quite complicated.
For example, in the context of k-means clustering and related methods,
\cite{bendavid}
showed that 
minimizing instability leads to 
poor clustering.
On the other hand,
\cite{rinaldo}
showed that, for density-based clustering, 
stability-based methods can sometimes lead to good results.
This motivates us to take a deeper look at stability 
for density clustering.

In this section, we investigate
two measures of stability
which we denote by $\Xi_n(h)$ and $\Gamma_n(h)$.
The measure $\Xi_n(h)$ is the stability of a fixed level set,
as a function of $h$.
We will see that $\Xi_n$ has surprisingly complex behavior.
See Figure \ref{fig::simple}.
First of all,
$\Xi_n(0) =0$.
This is an artifact and is due to the fact that the level sets get small as $h\to 0$.
As $h$ increases, $\Xi_n(h)$ first increases and then gets smaller.
Once it gets small enough, the level sets have become stable
and we have reached a good value of $h$.
However, after this point, $\Xi_n(h)$ continues to rise and fall.
The reason is that, as $h$ gets larger, $p_h(x)$ decreases.
Every time we reach a value of $h$ such that a mode of $p_h$
has height $\lambda$, $\Xi_n(h)$ will increase. 
$\Xi_n(h)$ is thus a non-monotonic function
whose mean and variance become large at particular values of $h$.
This behavior will be made explicit in the theory 
and simulations that follow.
As a practical matter, 
we can exclude all values of $h$
before the first local maximum of $\Xi_n(h)$.
Then, a reasonable choice of $h$ is the smallest value for which $\Xi_n(h)$ is less than
some pre-specified level $\beta$.

The second stability measure $\Gamma_n(h)$ is a more global
measure of stability.
When $\Gamma_n(h)$ is small, the whole cluster tree is stable.
It turns out that the behavior of $\Gamma_n(h)$ is much simpler.
It is monotonically decreasing as a function of $h$.
In this case we can choose $h$ to be the smallest $h$ for which
$\Gamma_n(h) \leq \beta$.

The motivation for this choice of $h$ is the following.
We cannot estimate loss exactly.
But we can use the instability to estimate variability.
Our choice of $h$ corresponds to making the bias as small as possible
while maintaining control over the 
variability.
This is very much in the spirit of the Neyman-Pearson approach to hypothesis testing
where one tries to make the power of a test as large as possible
while controlling the probability of false positives.
Put another way,
$P_h = P\oplus \mathbb{K}_h$ has
a blurred version of the shape information in $P$.
{\em We are choosing the smallest $h$ such that
the shape information in $P_h$
can be reliably recovered.}

Before getting into the details, which turn out to be somewhat technical,
here is a very loose description of the results.
For large $h$, $\Gamma_n(h) \approx 1/\sqrt{n h^d}$.
On the other hand,
$\Xi_n(h)$ tends to oscillate up and down
corresponding to the presence of modes of the density.
In regions where it is small, it also behaves like
$1/\sqrt{n h^d}$.

\subsection{Level Set Stability}
\label{sec:fixed.lambda}

In this section we focus on a single level set indexed by the density level $\lambda$.
Fix some $\lambda \geq 0$.
Consider two independent samples
$X=(X_1,\ldots, X_n)$ and 
$Y=(Y_1,\ldots, Y_n)$.
Let
\begin{eqnarray}
\xi(h) &=& \mathbb{E}_{XY}\left( P\left(\hat{L}_{h,X}(\lambda)\Delta \hat{L}_{h,Y}(\lambda)\right)\right).
\end{eqnarray}
Thus, $\xi(h)$ measures the disagreement between level sets based on two samples.

The definition of $\xi$ depends on $P$ which, of course, we do not know.
To estimate $\xi(h)$ we proceed as follows.
For simplicity, assume that the sample size is $3n$.
We randomly split the data into three pieces $(X,Y,Z)$ each of size $n$.
Let $\hat{p}_{h,X}$ be the density estimator constructed from $X$ and
$\hat{p}_{h,Y}$ be the density estimator constructed from $Y$.
Let
$\hat{P}_Z$ denote the empirical distribution of $Z$.
The sample instability statistic is
\begin{eqnarray}
\Xi_n(h) &=&\hat{P}_Z (\hat{L}_{h,X}(\lambda) \Delta \hat{L}_{h,Y}(\lambda)),
\end{eqnarray}
and its expectation is
\[
\xi(h) = \mathbb{E}_{X,Y,Z}[\Xi_n(h)].
\]



\noindent Note that since we are using the empirical distribution $\hat{P}_Z$, the sample instability can be rewritten as
\begin{eqnarray}
\Xi_n(h) &=& \frac{1}{n} \sum_{i=1}^n I (Z_i \in (\hat{L}_{h,X}(\lambda) \Delta \hat{L}_{h,Y}(\lambda)))\\
&=& \frac{1}{n} \sum_{i=1}^n I({\rm sign}(\hat{p}_{h,X}(Z_i)-\lambda) \neq {\rm sign}(\hat{p}_{h,Y}(Z_i)-\lambda)).
\end{eqnarray}

\noindent For a fixed $\lambda$, we count the fraction of the
observations in $Z$ where $\hat{p}_{h,X}(Z_i) < \lambda <
\hat{p}_{h,Y}(Z_i)$ or $\hat{p}_{h,X}(Z_i) > \lambda >
\hat{p}_{h,Y}(Z_i)$.  This representation is closely tied to the use of
the \emph{sample level sets} to construct the cluster tree
(\cite{nugent}) where each level set is represented only by the
observations associated with its connected components rather than the
feature space.  Using the empirical distribution $\hat{P}_Z$ also
removes the need to determine the exact shape of the density
estimate's level sets.  The top graph of Figure \ref{fig::simple}
shows the sample instability as a function of $h$ for $\lambda = 0.09$
for our example distribution.  Note that the instability initially
drops and then oscillates before dropping to zero at $h = 7.08$,
indicating the multi-modality seen in Figure \ref{fig::intro}.  More
discussion of this example is in Section \ref{section::examples}.

We now present the following simple but important boundary
properties of $\Xi_n$ and $\xi$. The proof is straightforward and is
omitted. 

\begin{lemma}
For fixed $n$, 
$\lim_{h \rightarrow 0} \xi(h) = \lim_{h \rightarrow \infty}\xi(h)=0$, and 
$\lim_{h \rightarrow 0} \Xi_n(h) =
\lim_{h \rightarrow \infty}\Xi_n(h) =0$ a.s.
\end{lemma}

We now study the behavior of the mean function
$\xi(h)$.
Let $u \in \mathbb{R}^d$, $h > 0$ and $\epsilon > 0$, and define
\begin{equation}\label{eq:piu}
\pi_h(u) = \mathbb{P}_X(\hat{p}_{h,X}(u) > \lambda) \quad 
\text{ and } \quad U_{h,\epsilon} = \{ u \colon |p_h(u) - \lambda| < \epsilon\}.
\end{equation}


\begin{theorem}
\label{thm:xi}
Let $u \in \mathbb{R}^d$, $h > 0$ and $\epsilon > 0$.
\begin{enumerate} 
\item  The following identity holds:
\[
\xi(h) = 2  \int_{\mathbb{R}^d}\pi_h(u)(1-\pi_h(u)) d P(u) .
\]
\item For all large $n$,
\begin{equation}\label{eq::double-sided}
r_{h,\epsilon} \ \underline{A}_{h,\epsilon} \leq \xi(h) \leq  r_{h,\epsilon}  \ 
\overline{A}_{h,\epsilon} + 2K_1 e^{- K_2 n h^d \epsilon^2},
\end{equation}
where
$
r_{h,\epsilon} = P(U_{h,\epsilon}),
$
\[
 \overline{A}_{h,\epsilon}  =  \sup_{u \in U_{h,\epsilon}} 2 \pi_h(u) (1 - \pi_h(u))
\]
and
\[
\underline{A}_{h,\epsilon} =  \inf_{u \in U_{h,\epsilon}} 2 \pi_h(u) (1 - \pi_h(u)).
\]
\end{enumerate} 
\end{theorem}

\begin{figure}
\begin{center}
\includegraphics[width=2.5in,height=2.5in]{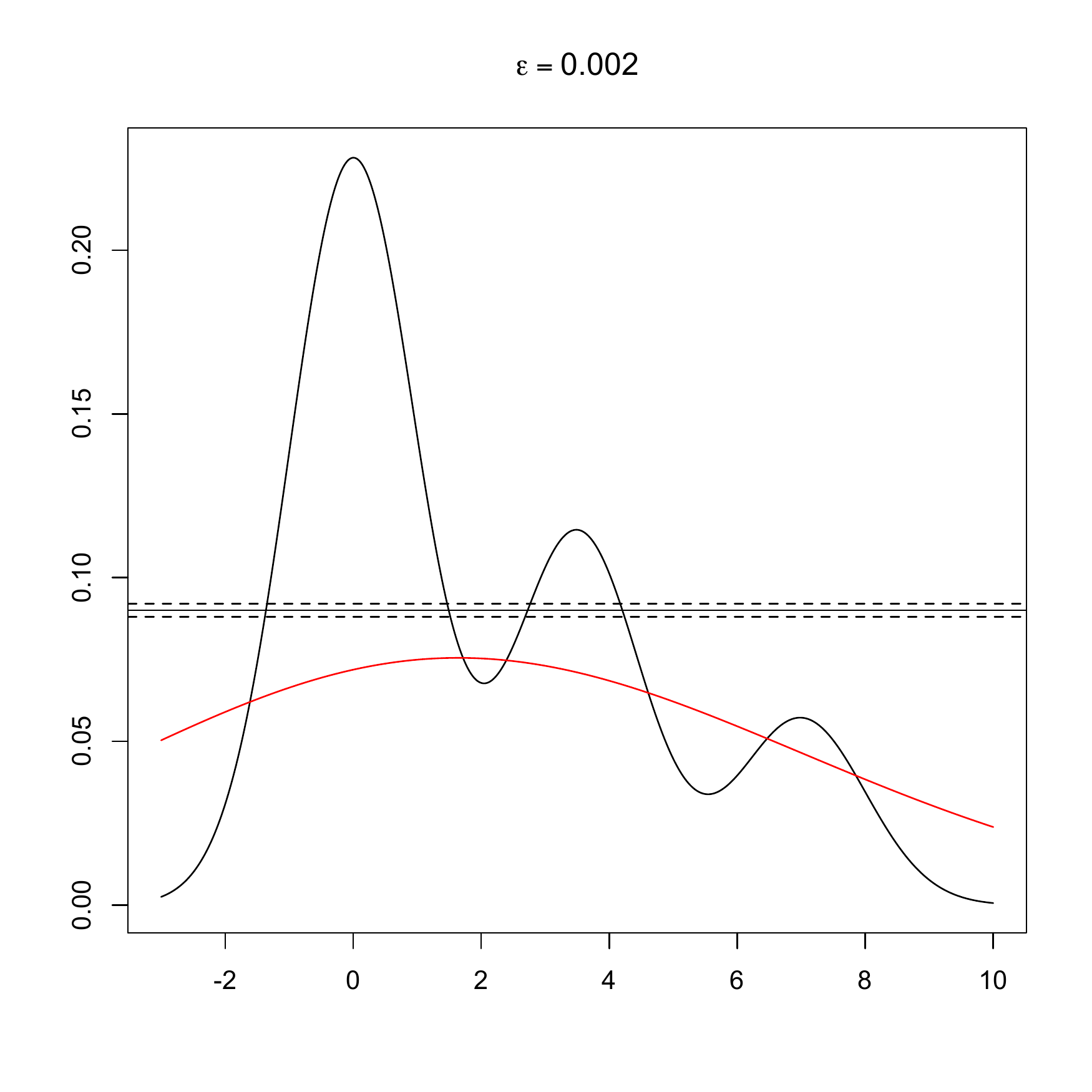}
\includegraphics[width=2.5in,height=2.5in]{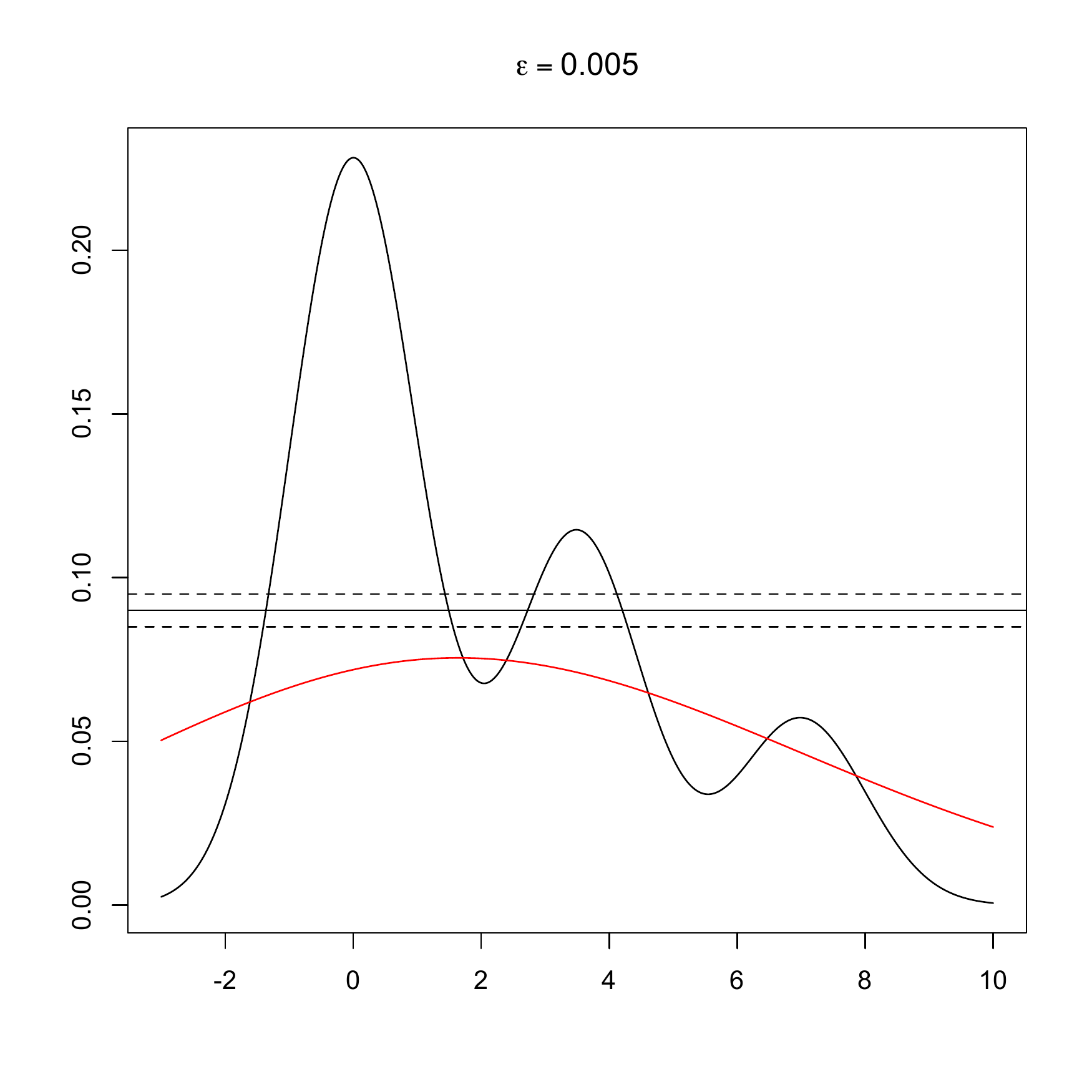}

\includegraphics[width=2.5in,height=2.5in]{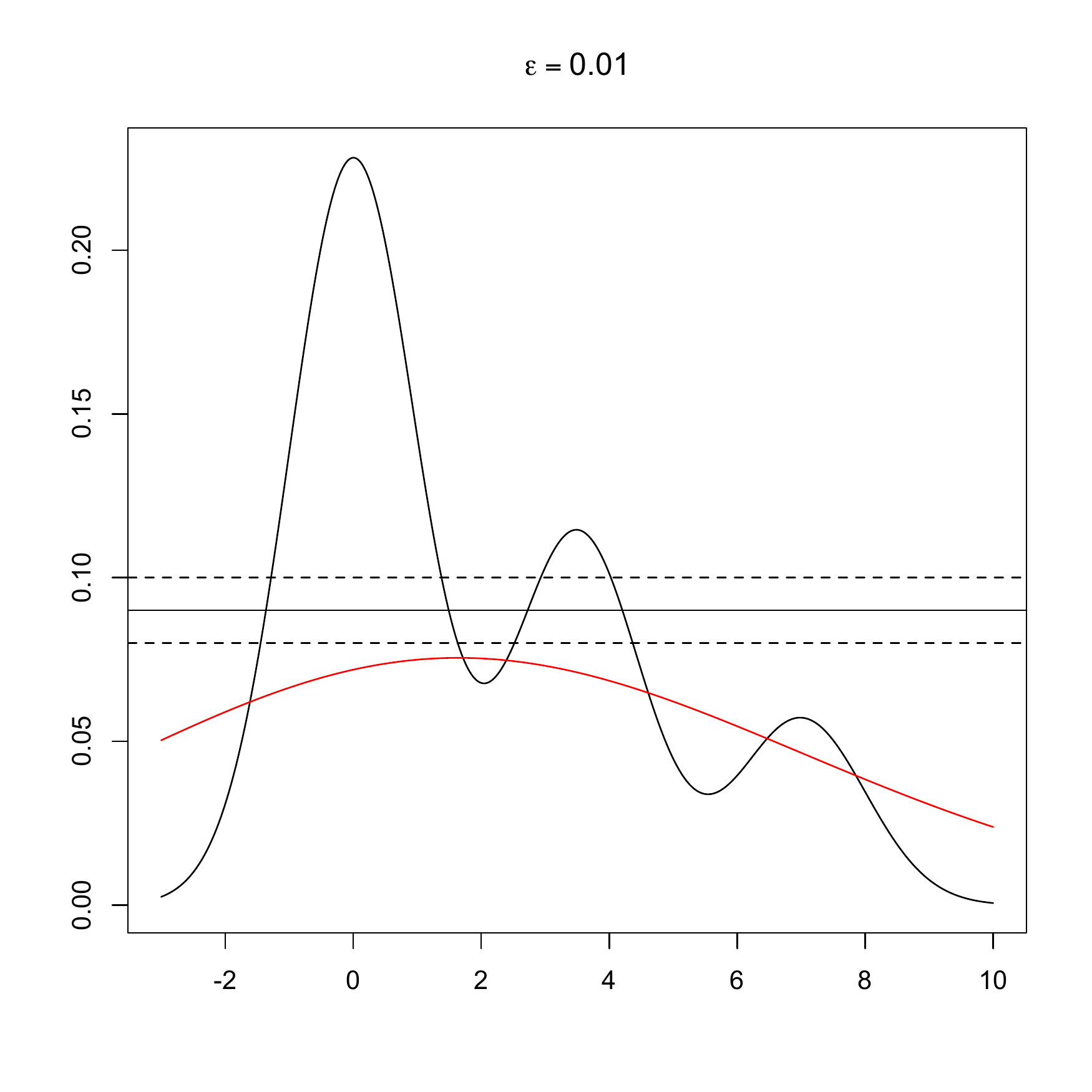}
\includegraphics[width=2.5in,height=2.5in]{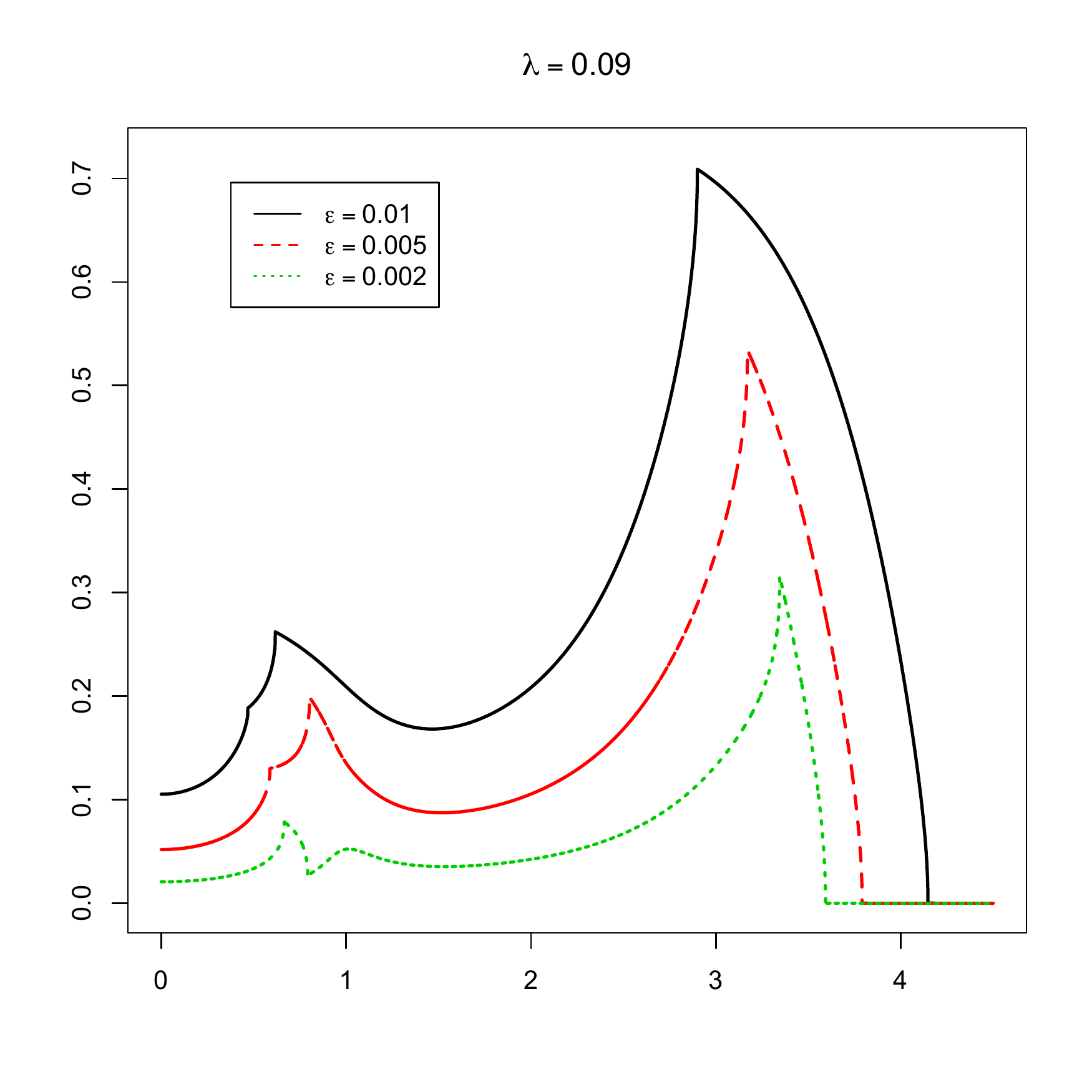}
\end{center}
\caption{Top plots and left bottom plot: two densities $p_h$ corresponding to the mixture distribution of Figure \ref{fig::intro} for $h=0$, i.e. the true density, (in black) and $h=4.5$ (in red);  the horizontal lines indicate the level set value of  $\lambda = 0.09$, $\lambda+\epsilon $ and $\lambda - \epsilon$, for $\epsilon$ equal to $0.02$, $0.05$ and $0.1$. Right bottom plot: probability content values $r_{h,\epsilon}$ as a function of $h \in [0,4.5]$ for the three values of $\epsilon$.}
\label{fig::U}
\end{figure}

Part 2 of the previous theorem implies that the behavior of $\xi$ is essentially captured by the behavior of the probability content $r_{h,\epsilon}$. This quantity is, in general, a complicated function of both $h$ and $\epsilon$. While it is easy to see that, for fixed $h$ and a sufficiently well-behaved density $p$, $r_{h,\epsilon} \rightarrow 0$ as $\epsilon \rightarrow 0$, for fixed $\epsilon$, $r_{h,\epsilon}$ can instead be a non-monotonic function of $h$. See, for example, the bottom right plot in Figure \ref{fig::U}, which displays the values $r_{h,\epsilon}$ as a function of $h \in [0,4.5]$ and for $\epsilon$ equal to  $0.02$, $0.05$ and $0.1$ for the mixture density of Figure \ref{fig::intro}. In particular, the fluctuations of $r_{h,\epsilon}$ as a function of $h$ are related to the values of $h$ for which the critical points of $p_h$ are in the interval $[\lambda - \epsilon, \lambda + \epsilon]$.
The main point to notice is that
$r_{h,\epsilon}$ is a complicated, non-monotonic function of $h$.
This explains why $\Gamma_n(h)$ is non-monotonic in $h$.

As mentioned at the end of section  \ref{sec:est.density}, for values of $h << \left( \frac{\log n}{n}\right)^{1/d}$, smaller than minimal spacing among the sample points, the kernel density estimate $\hat{p}_h$ is no longer a reliable estimate of $p_h$. We describe this effect on the expected instability in our next result.

\begin{lemma}
\label{lemma:xi_smallh}
Fix $\lambda >0$. Then, for any fixed $n$ large enough, $\xi(h) = O(h^d)$ as $h \rightarrow 0$.
\end{lemma}

We now provide an upper and lower bound on the values of $
\overline{A}_{h,\epsilon}$ and $\underline{A}_{h,\epsilon}$,
respectively, under the simplifying assumption that $K$ is the
spherical kernel. Notice that, while $ \overline{A}_{h,\epsilon}$
remains bounded away from $\infty$ for any sequence $\epsilon_n \rightarrow
0$ and $h_n =\omega(n^{-1/d})$, the same is not true for
$\underline{A}_{h,\epsilon}$, which remains bounded away from $0$ as
long as $\epsilon_n = \Theta(\frac{1}{n  h_n^d})$ and  $h_n =\omega(n^{-1/d})$. 


\begin{lemma}\label{prop:lambda.berry-esseen}
Assume that $K$ is the spherical kernel and let  $0 < \epsilon \leq \lambda/2$. For a given $\delta \in (0,1)$, let
\begin{equation}
h(\delta,\epsilon) = \sup \Bigl\{h:\  \sup_{u \in U_{h,\epsilon}} P(B(u,h)) \leq 1-\delta \Bigr\}.
\end{equation}
Then, for all $h \leq h(\delta,\epsilon)$, 
\[
\overline{A}_{h,\epsilon} \leq 
2 \left(1 - \Phi\left( - \sqrt{nh^d}\epsilon  \frac{ 2 v_d}{3 \lambda}\right) + \frac{C(\delta,\lambda)}{\sqrt{nh^d }}\right)^2 ,
\]
and
\[
\underline{A}_{h,\epsilon} \geq 
2 \left(1 - \Phi\left( \sqrt{nh^d}\epsilon \frac{ 2 v_d }{\delta \lambda}\right) - \frac{C(\delta,\lambda)}{\sqrt{nh^d }}\right)^2 ,
\]
where $\Phi$ denote the cumulative distribution function of a standard normal random variable and
\[
C(\delta,\lambda) = \frac{33}{4}\sqrt{\frac{2}{\delta v_d \lambda}}.
\]
\end{lemma}


The dips in Figure \ref{fig::simple}
correspond to values for which $p_h$ does not have a mode at height $\lambda$.
In this case, (B2) holds and we have
$r_{h,\epsilon} = O(\epsilon)$. 
Now choosing $\epsilon \approx \sqrt{\log n /(nh^d)}$ for the upper bound
and $\epsilon \approx \sqrt{1/(nh^d)}$ for the lower bound, we have that
$\overline{A}_{h,\epsilon}$ and
$\underline{A}_{h,\epsilon}$ are bounded, and
the theorem yields
\begin{equation}
\sqrt{ \frac{C_1 }{nh^d}} \leq \xi(h) \leq
\sqrt{ \frac{C_2 \log n}{nh^d}}.
\end{equation}

Next we investigate the extent to which $\Xi_n(h)$ is concentrated
around its mean $\xi(h) = \mathbb{E}[\Xi_n(h)]$. We first point out
that, for any fixed $h$, the variance of the instability can be
bounded by $\xi(h)( 1/2 - \xi(h)) $.

\begin{lemma}
\label{lemma::boundvar}
For any $h>0$, 
\[
\mathrm{Var}[\Xi_n(h)] \leq \xi(h)  \left( \frac{n+1}{2n} - \xi(h) \right) \approx
\xi(h)  \left( \frac{1}{2} - \xi(h) \right) .
\]
\end{lemma}

The previous results highlight the interesting feature that the
empirical instability will be less variable around the values of $h$
for which the expected instability is very small (close to $0$) or
very large (close to $1/2$).

\begin{lemma}\label{lem:conc}
For any $h>0$, $\epsilon > 0$, $\eta \in (0,1)$ let $t$ be such that
\begin{equation}\label{eq:t}
t ( 1- \eta)  \geq  r_{h,\epsilon} + 2 K_1 e^{-K_2 n \epsilon^2 h^d},
\end{equation}
where $r_{h,\epsilon} = P(U_{h,\epsilon})$. Then, for all large $n$,
\begin{equation}\label{eq:Xi.xi}
\mathbb{P}_{X,Y,Z} \left( \left| \Xi_n(h) - \xi(h) \right| > t \right) 
\leq e^{ - n t C_\eta }+ 2 K_1 e^{ - n K_2  h^d \epsilon^2 }
\end{equation}
where
\[
C_\eta = 9  ( 1- \eta)  \left( \frac{3 -2 \eta}{3(1 - \eta)} - \sqrt{\frac{3 - \eta}{3(1 - \eta)}} \right).
\]
\end{lemma}

\subsection{Stability of level sets indexed by probability content}
\label{subsection::fixed-prob-content}

As in the fixed-$\lambda$ case, we assume for simplicity that
the sample has size $3n$ and split it equally in three parts: $X$, $Y$
and $Z$.  We now define the fixed-$\alpha$ instability as
\[
\Xi_n(h,\alpha) = \widehat P_Z(\hat{M}_{h,X}(\alpha) \Delta \hat{M}_{h,Y}(\alpha)),
\]
where
\begin{equation}\label{eq:Lhat}
\hat{M}_{h,X}(\alpha) = \{ x \colon \hat{p}_{h,X}(x) > \hat{\lambda}_{h,\alpha,X}\}, \quad
\end{equation}
with $ \hat{\lambda}_{h,\alpha,X}$ estimated as in \eqref{eq:hat.lambda} using the points in $X$; we similarly estimate $\hat{M}_{h,Y}(\alpha)$. As before, $\hat{P}_Z$ denote the empirical measure arising from $Z$.  Again, we use the observations to represent $\hat{M}_{h,X}$, $\hat{M}_{h,Y}$ as done for $\Xi_n(h)$ for a fixed $\lambda$.  Examples of $\Xi_n(h,\alpha)$ as a function of $h, \alpha$ can be seen in Section \ref{section::examples}.

The expected instability is
\[
\xi(h,\alpha) = \mathbb{E}_{X,Y,Z}[\Xi_n(h,\alpha)].
\]
We begin by studying the behavior of the expected instability.

\begin{theorem}\label{thm:instab.alpha}
Let $u \in \mathbb{R}^d$, $h > 0$ and $\epsilon > 0$, and set 
\[
\pi_{h,\alpha}(u) = \mathbb{P}_X(\hat{p}_{h,X}(u) > \hat{\lambda}_{h,\alpha,X}) \quad \text{ and } U_{h,2 \epsilon,\alpha} = \{ u \colon |p_h(u) - \lambda_{\alpha,h}| \leq 2\epsilon \}.
\]
\begin{enumerate}
\item The expected instability can be expressed as
\[
\xi(h,\alpha) = \mathbb{E}_{X,Y,Z}[\Xi_n(h,\alpha)] = 2 \int_{\mathbb{R}^d} \pi_{h,\alpha}(u) (1 - \pi_{h,\alpha}(u)) dP(u).
\]
\item Let $\epsilon < \eta_0 - 1/n$ and $\tilde{\epsilon} = \epsilon(A\kappa_3 + 1) + A\kappa_3/n$. Then, for all large $n$,
\[
P(U_{h,2\tilde{\epsilon},\alpha}) \underline{A}_{h,\epsilon,\alpha} \leq \xi(h) \leq P(U_{h,2\tilde{\epsilon},\alpha}) \overline{A}_{h,\epsilon,\alpha} + 4 K_1 e^{ - K_2 n h^d \epsilon^2} + 16 n e^{-n \epsilon^2/32},
\]
where
\[
\overline{A}_{h,\epsilon,\alpha} = \sup_{u \in U_{h,2\tilde{\epsilon},\alpha}} 2 \pi_{h,\alpha}(u) (1 - \pi_{h,\alpha}(u))
\]
and
\[
\underline{A}_{h,\epsilon,\alpha} = \inf_{u \in U_{h,2\tilde{\epsilon},\alpha}} 2 \pi_{h,\alpha}(u) (1 - \pi_{h,\alpha}(u)).
\]
\item Assume in addition that $K$ is the spherical kernel and that $\tilde{\epsilon} \leq \inf_{h} \frac{\lambda_{h,\alpha}}{4}$. For a given $\delta \in (0,1)$, let
\begin{equation}
h(\delta,\epsilon,\alpha) = \sup \Bigl\{h:\ \sup_{u \in U_{h,\tilde{\epsilon} ,\alpha}} P(B(u,h)) \leq 1-\delta \Bigr\}.
\end{equation}
Then, for all $h \leq h(\delta,\epsilon,\alpha)$, 
\[
\overline{A}_{h,\epsilon,\alpha} \leq
2 \left(1 - \Phi\left( - 3\sqrt{nh^d}\tilde{\epsilon} \frac{ 2 v_d}{3 \lambda_{h,\alpha}}\right) + \frac{C(\delta,\lambda_{h,\alpha})}{\sqrt{nh^d }}+ 4 K_1 e^{ - K_2 n h^d \epsilon^2} + 16 n e^{-n \epsilon^2/32}\right)^2 ,
\]
and
\[
\underline{A}_{h,\epsilon,\alpha} \geq
2 \left(1 - \Phi\left( 3\sqrt{nh^d}\tilde{\epsilon} \frac{ 2 v_d }{\delta \lambda_{h,\alpha}}\right) - \frac{C(\delta,\lambda_{h,\alpha})}{\sqrt{nh^d }} - 4 K_1 e^{ - K_2 n h^d \epsilon^2} - 16 n e^{-n \epsilon^2/32}\right)^2 ,
\]
where $\Phi$ denote the cumulative distribution function of a standard normal random variable and
\[
C(\delta,\lambda_{h,\alpha}) = \frac{33}{4}\sqrt{\frac{2}{\delta v_d \lambda_{h,\alpha}}}.
\]
\end{enumerate}
\end{theorem}


As for the fluctuations of $\Xi_n(h,\alpha)$ around its mean, we can easily obtain a result similar to the one we obtain in Lemma \ref{lem:conc}.

\begin{lemma}\label{lem:conc2}
For any $h>0$, $\epsilon > 0$, $\eta \in (0,1)$ let $\tilde{\epsilon} = \epsilon(A\kappa_3 + 1) + A\kappa_3/n$ and $t$ be such that
\[
t ( 1- \eta)  \geq  r_{h,\epsilon,\alpha} + 4 K_1 e^{ - K_2 n h^d \epsilon^2} + 16 n e^{-n \epsilon^2/32},
\]
where $r_{h,\epsilon,\alpha} = P(\{ u \colon |p_h(u) - \lambda_{h,\alpha}| \leq 2 \tilde{\epsilon}\})$.
Then, for all large $n$,
\begin{equation}\label{eq:Xi.xi.alpha}
\mathbb{P}_{X,Y,Z} \left( \left| \Xi_n(h,\alpha) - \xi(h,\alpha) \right| > t \right) 
\leq e^{ - n t C_\eta }+ 4 K_1 e^{ - K_2 n h^d \epsilon^2} + 
16 n e^{-n \epsilon^2/32}.
\frac{\log \frac{2}{\delta}}{3n}   \leq \delta + 2 K_1 \exp \left\{ - n K_2  h^d \epsilon^2  \right\}.
\end{equation}
with
\[
C_\eta = 9  ( 1- \eta)  \left( \frac{3 -2 \eta}{3(1 - \eta)} - \sqrt{\frac{3 - \eta}{3(1 - \eta)}} \right).
\]
\end{lemma}

The proof is basically the same as the proof of Lemma \ref{lem:conc},
except that we have to restrict our analysis to the event in
\eqref{eq:Atilde}.
We omit the details.

\subsection{Stability for density cluster trees}

The stability properties of the density tree can be easily derived from the results we have established so far.
To this end, for a fixed $h>0$, define the level set of $p_h$
\[
L_h(\lambda) = \{u:p_h(u) >\lambda\}
\]
and recall the level set estimate
\[
\hat{L}_{h,X}(\lambda) = \{ u \colon \hat{p}_{h,X}(u) > \lambda \}.
\]
Let $N_h(\lambda)$, $\hat{N}_{h,X}(\lambda)$ be the number of connected components of the sets $L_{h}(\lambda)$ and $\hat{L}_{h,X}(\lambda)$, respectively. Notice that $\hat{L}_{h,X}(\lambda)$ is a random set. Also, denote with $C_{1},\ldots,C_{N_h(\lambda)}$ and $\hat{C}_{1},\ldots,\hat{C}_{\hat{N}_{h,X}(\lambda)}$ the connected components of $L_h(\lambda)$ and $\hat{L}_{h,X}(\lambda)$, respectively.

When building cluster trees, the value of the bandwidth $h$ is kept fixed and the values of the level $\lambda$ vary instead. It has been observed empirically \citep[see, e.g.][]{nugent} that the uncertainty of cluster trees depend on the particular value of $\lambda$ at which the tree is observed. In order to characterize the behavior of the density tree, we propose the following definition. 
\begin{definition}
A level set value $\lambda$ is $(h,\epsilon)$-stable, with $\epsilon > 0$ and $h >0$, if
\[
N_h(\lambda) = N_h(\lambda'), \quad \forall \lambda' \in (\lambda - \epsilon, \lambda + \epsilon)
\]
and, for any $\lambda -\epsilon < \lambda_1 < \lambda_2 < \lambda + \epsilon$, 
\[
C_i(\lambda_2) \subseteq C_i(\lambda_1), \quad \forall i=1,\ldots,N_h(\lambda).
\]
\end{definition}
If the level $\lambda$ is $(h,\epsilon)$-stable, then the cluster tree estimate at level $\lambda$ is an accurate estimate of the true cluster tree, in a sense made precise by the following result, whose proof follows easily from the proofs of our previous results and Lemma 2 in \cite{rinaldo}.

\begin{lemma}
If $\lambda$ is $(h,\epsilon)$-stable, then, for all $n$ large enough, with probability at least $1 - \frac{1}{n}$,
\begin{enumerate}
\item $N_h(\lambda) = \hat{N}_{h,X}(\lambda)$;
\item there exists a permutation $\sigma$ on $\{1,\ldots,N_h(\lambda) \}$ such that, for every connected component $C_j$ of $L_h(\lambda - \epsilon)$ there exists one $\hat{C}_{\sigma(j)}$ for which 
\[
C_j \subseteq \hat{C}_{\sigma(j)};
\]
\item $P(\hat{L}_{h,X}(\lambda) \Delta L_{h}(\lambda)) \leq P(\{u:|p_h(u) - \lambda| < \epsilon\})$.
\end{enumerate}

\end{lemma}

{\bf Remarks.}
\begin{enumerate}
\item The values of $\lambda$ which are not $(h,\epsilon)$-stable are the ones for which 
\[
\inf_{u \in U_{\lambda',h,\epsilon}} \| \nabla p_h(u) \| = 0,
\]
for some $\lambda' \in (\lambda - \epsilon, \lambda + \epsilon)$. For those values, the probability of  $N_h(\lambda) \neq \hat{N}_{h,X}(\lambda)$ can be quite large, since the set $\hat{L}_{h,X} \Delta L_h(\lambda)$ may have a relatively large $P$-mass. 
\item Conversely, if $p_h$ is smooth (which is the case if, for instance, the kernel or $p$ are smooth) and $\inf_{u \in U_{\lambda,h,\epsilon}} \| \nabla p_h(u) \| > \delta$, then $\lambda$ is $(h,\epsilon)$-stable for a small enough $\epsilon$.
\end{enumerate}

The above result has a somewhat limited practical value, because the notion of a $(h,\epsilon)$-stable $\lambda$ depends on the unknown density $p_h$. In order to get a better sense of which $\lambda$'s are $(h,\epsilon)$-stable or not, we once again resort to evaluate the instability of the clustering solution via data splitting.
In fact, essentially all of our previous results about instability from section \ref{sec:fixed.lambda} carry over to these new settings by treating $h$ fixed and letting $\lambda$ vary. To express this changes explicitly, we will adopt a slightly different notation for quantities we have already considered. In particular, we let 
\begin{eqnarray*}
U_{\lambda,\epsilon} &=& \{ u \colon |p_h(u) - \lambda| < \epsilon\}\\
 r_{\lambda,\epsilon} &=& P(U_{\lambda,\epsilon})\\
 \pi_\lambda(u) &=& \mathbb{P}_X(\hat{p}_{h,X}(u) > \lambda)\\
 \overline{A}_{\lambda,\epsilon} &=& \sup_{u \in U_{\lambda,\epsilon}} 2\pi_\lambda(u)( 1- \pi_\lambda(u))\\
 \underline{A}_{\lambda,\epsilon} &=& \inf_{u \in U_{\lambda,\epsilon}} 2\pi_\lambda(u)( 1- \pi_\lambda(u)).\\
\end{eqnarray*}


We divide the sample size into three distinct groups, $X$, $Y$ and $Z$, of equal sizes $n$.
Define the instability of the density cluster tree as the random function $T_n \colon \mathbb{R}_{\geq 0} \mapsto [0,1]$ given by
\[
\lambda \rightarrow \hat{\mathbb{P}}_Z (\hat{L}_{h,X}(\lambda) \Delta \hat{L}_{h,Y}(\lambda)).
\]
Also, let
\[
\tau(\lambda) = \mathbb{E}_{X,Y,Z}[T_n(\lambda)].
\]
For any fixed $h$, the behavior of $T_n(\lambda)$ and
$\tau(\lambda)$ is essentially governed by $r_{\lambda,\epsilon}$.
The following result describes some of the properties of the density
tree instability. We omit its proof, because it relies essentially on the
same arguments from the proofs of the results described in section
\ref{sec:fixed.lambda}.

\begin{corollary}
$\;$
\begin{enumerate}
\item For any $\lambda > 0$, the expected density tree instability can be expressed as
\[
\tau(\lambda) =  2 \int \pi_{\lambda}(u) (1 - \pi_{\lambda}(u)) dP(u).
\]
\item For any $\epsilon > 0$ and $\lambda > 0$,
\[
  \underline{A}_{\lambda,\epsilon} r_{\lambda,\epsilon} \leq \tau(\lambda) \leq \underline{A}_{\lambda,\epsilon} r_{\lambda,\epsilon} +2  K_1 e^{ - K_2 n h^d \epsilon^2},
\]
for all $n$ large enough.
\item Assume that $K$ is the spherical kernel. For any $\lambda > 0$, let $0 < \epsilon \leq \frac{\lambda}{2}$ and let 
\[
\delta = 1 - \sup_{u} P(B(u,h)).
\]
Then,
\[
\overline{A}_{\lambda,\epsilon} \leq 
2 \left(1 - \Phi\left( - \sqrt{nh^d}\epsilon  \frac{ 2 v_d}{3 \lambda}\right) + \frac{C(\delta,\lambda)}{\sqrt{nh^d }}\right)^2 ,
\]
and
\[
\underline{A}_{\lambda,\epsilon} \geq 
2 \left(1 - \Phi\left( \sqrt{nh^d}\epsilon \frac{ 2 v_d }{\delta \lambda}\right) - \frac{C(\delta,\lambda)}{\sqrt{nh^d }}\right)^2 ,
\]
where $\Phi$ denote the cumulative distribution function of a standard normal random variable and
\[
C(\delta,\lambda) = \frac{33}{4}\sqrt{\frac{2}{\delta v_d \lambda}}.
\]
\item For any $h>0$, $\epsilon > 0$, $\eta \in (0,1)$ let $t$ by such that
\begin{equation}\label{eq:t2}
t ( 1- \eta)  \geq  r_{\lambda,\epsilon} + 2 K_1 e^{-K_2 n \epsilon^2 h^d},
\end{equation}
Then, for all $n$ that are large enough 
\begin{equation}\label{eq:Xi.xi2}
\mathbb{P}_{X,Y,Z} \left( \left| T_n(\lambda) - \tau(\lambda) \right| > t \right) \leq e^{ - n t C_\eta }+ 2 K_1 e^{ - n K_2  h^d \epsilon^2 }.
\end{equation}
with
\[
C_\eta = 9  ( 1- \eta)  \left( \frac{3 -2 \eta}{3(1 - \eta)} - \sqrt{\frac{3 - \eta}{3(1 - \eta)}} \right).
\]
\end{enumerate}
\end{corollary}


\subsection{Total Variation Stability}

In the previous section, we established stability of the cluster tree
for a fixed $h$ and all levels $\lambda$ that are $(h,\epsilon)$-stable.
To establish stability of the entire cluster tree, we will now consider an even stronger notion of instability.
Let ${\cal B}$ denote all measurable subsets of $\mathbb{R}^d$.
Define the {\em total variation instability} 
$$
\Gamma_n(h) \equiv \sup_{B\in {\cal B}}
\left| \int_B \hat{p}_{h,X}(u)du - \int_B \hat{p}_{h,Y}(u)du\right| =
\frac{1}{2}\int \left| \hat{p}_{h,X}(u) - \hat{p}_{h,Y}(u)\right| du
$$
where the latter equality is a standard identity.
Requiring $\Gamma_n(h)$ to be small is a more demanding type of stability.
In particular, ${\cal B}$ includes all level sets
for all $\lambda$.
Thus, when $\Gamma_n(h)$ is small, the entire cluster tree is stable.
Note that $\Gamma_n(h)$ is easy to interpret: it is the maximum
difference in probability between the two density estimators.
And of course $0 \leq \Gamma_n(h) \leq 1$.  The bottom graph in Figure \ref{fig::simple} 
shows the total variation instability for our example distribution in Figure \ref{fig::intro}.  
Note that $\Gamma_n(h)$ first drops drastically as $h$ increases and then continues to smoothly decrease.

We now discuss the properties of $\Gamma_n(h)$.
Note first that $\Gamma_n(h)\approx 1$ for small $h$
so the behavior as $h$ gets large is
most relevant.

\begin{theorem}
\label{thm::total-var}
Let ${\cal H}_n$ be a finite set of bandwidths such that
$|\mathcal{H}_n| = A n^a$, for some positive $A$ and $a \in (0,1)$.
Fix a $\delta \in (0,1)$.
\begin{enumerate}
\item 
(Upper bound.)
There exists a constant $C$ such that, for all $n$ large enough and such that $\delta > A/n$, 
$$
\mathbb{P}_{X,Y}\left( \Gamma_n(h) \leq t_{h} \ \ {\rm for\ all\ }h\in {\cal H}_n\right)> 1-\delta,
$$
where $t_h = \sqrt{\frac{C \log n}{n h^d}}$.
\item 
(Lower bound.) Suppose that $K$ is the spherical kernel and that the probability distribution $P$ satisfies the conditions
\begin{equation}\label{eq:a1a2}
a_1 h^d v_d \leq \inf_{u \in S}P(B(u,h)) \leq \sup_{u \in S}P(B(u,h)) \leq h^d v_d a_2, \quad \forall h>0,
\end{equation}
for some positive constants $a_1 < a_2$, where $S$ denotes the support of $P$. Let
Let $h_*$ be such that
$\sup_u P(B(u,h_*)) < 1-\delta$.
There exists a $t$, depending on $\delta$ but not on $h$, such that,
for all $h < h_*$ and all $n$ large enough,
$$
\mathbb{P}_{X,Y}\left( \Gamma_n(h) \geq t \sqrt{\frac{ 1}{n h^d}}\right)> 1-\delta.
$$
\item $\Gamma_n(0)=1$ and $\Gamma_n(\infty)=0$.
\end{enumerate}
\end{theorem}

\vspace{.5cm}
\noindent {\bf Remarks.}\\
\begin{enumerate}
\item Note that the upper bound is uniform in $h$ while the lower bound is pointwise in $h$.
Making the lower bound uniform is an open problem.
However, if we place a nonzero lower bound on the bandwidths in ${\cal H}_n$
then the bound could be made uniform. This approach was used in \cite{marron}.
\item Conditions \eqref{eq:a1a2} are quite standard in support set estimation. In particular, when the lower bound holds, the support $S$ is said to be {\it standard}. See, for instance, \cite{cuevascasal04}.
\end{enumerate}

In low dimensions, we can
compute $\Gamma_n(h)$ by
numerically evaluating the integral
\[
\frac{1}{2}\int \left| \hat{p}_{h,X}(u) - \hat{p}_{h,Y}(u)\right| du.
\]
In high dimensions
it may be easier to use
importance sampling as follows.
Let
$g(u) = (1/2)(\hat{p}_{h,X}(u) + \hat{p}_{h,Y}(u))$.
Then
$$
\Gamma_n(h) = 
\frac{1}{2}\int \frac{\left| \hat{p}_{h,X}(u) - \hat{p}_{h,Y}(u)\right|}{g(u)} g(u) du \approx
\frac{1}{N}\sum_{i=1}^N
\frac{\left| \hat{p}_{h,X}(U_i) - \hat{p}_{h,Y}(U_i)\right|}
{\left| \hat{p}_{h,X}(U_i) + \hat{p}_{h,Y}(U_i)\right|}
$$
where
$U_1, \ldots, U_N$ is a random sample sample from $g$.
We can thus estimate $\Gamma_n(h)$ with the following algorithm:

\vspace{.5cm}

\rule{6in}{.1cm}

\begin{enumerate}
\item Draw Bernoulli(1/2) random variables $Z_1,\ldots, Z_N$.
\item Draw $U_1, \ldots, U_N$ as follows:
\begin{enumerate}
\item If $Z_i =1$: draw $X$ randomly from $X_1,\ldots, X_n$.
Draw $W\sim K$. Set $U_i = X + h W$.
\item If $Z_i =0$: draw $Y$ randomly from $Y_1,\ldots, Y_n$.
Draw $W\sim K$. Set $U_i = Y + h W$.
\end{enumerate}
\item Set
$$
\hat\Gamma_n(h) =  
\frac{1}{N}\sum_{i=1}^N
\frac{\left| \hat{p}_{h,X}(U_i) - \hat{p}_{h,Y}(U_i)\right|}
{\left| \hat{p}_{h,X}(U_i) + \hat{p}_{h,Y}(U_i)\right|}.
$$
\end{enumerate}

\rule{6in}{.1cm}

\vspace{.5cm}

It is easy to see that $U_i$ has density $g$ and that
$\hat\Gamma_n(h) -\Gamma_n(h) = O_P(1/\sqrt{N})$
which is negligible for large $N$.


\section{Examples}
\label{section::examples}

\begin{figure}[t]
\begin{center}
\includegraphics[width=2.5in,height=2.5in]{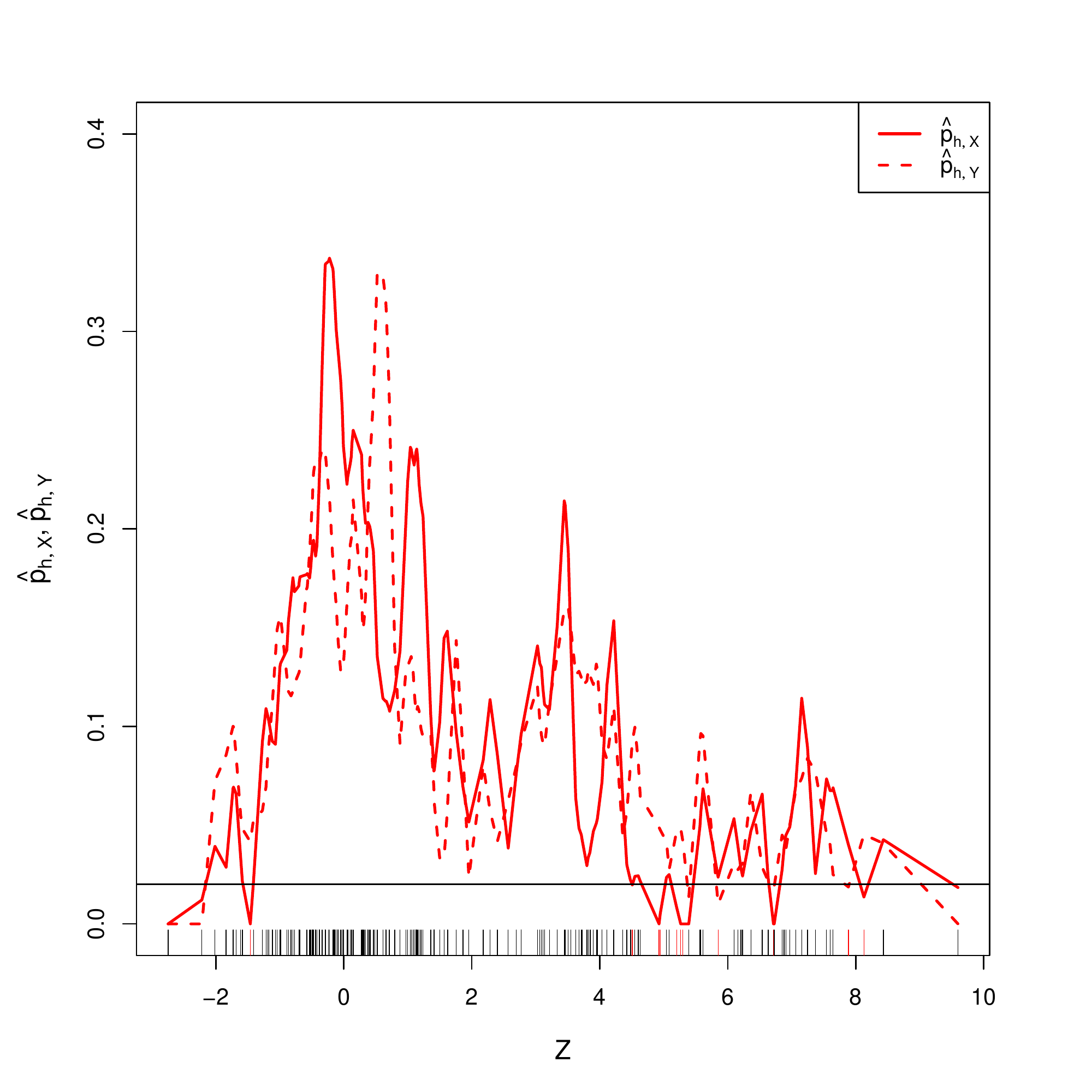}
\includegraphics[width=2.5in,height=2.5in]{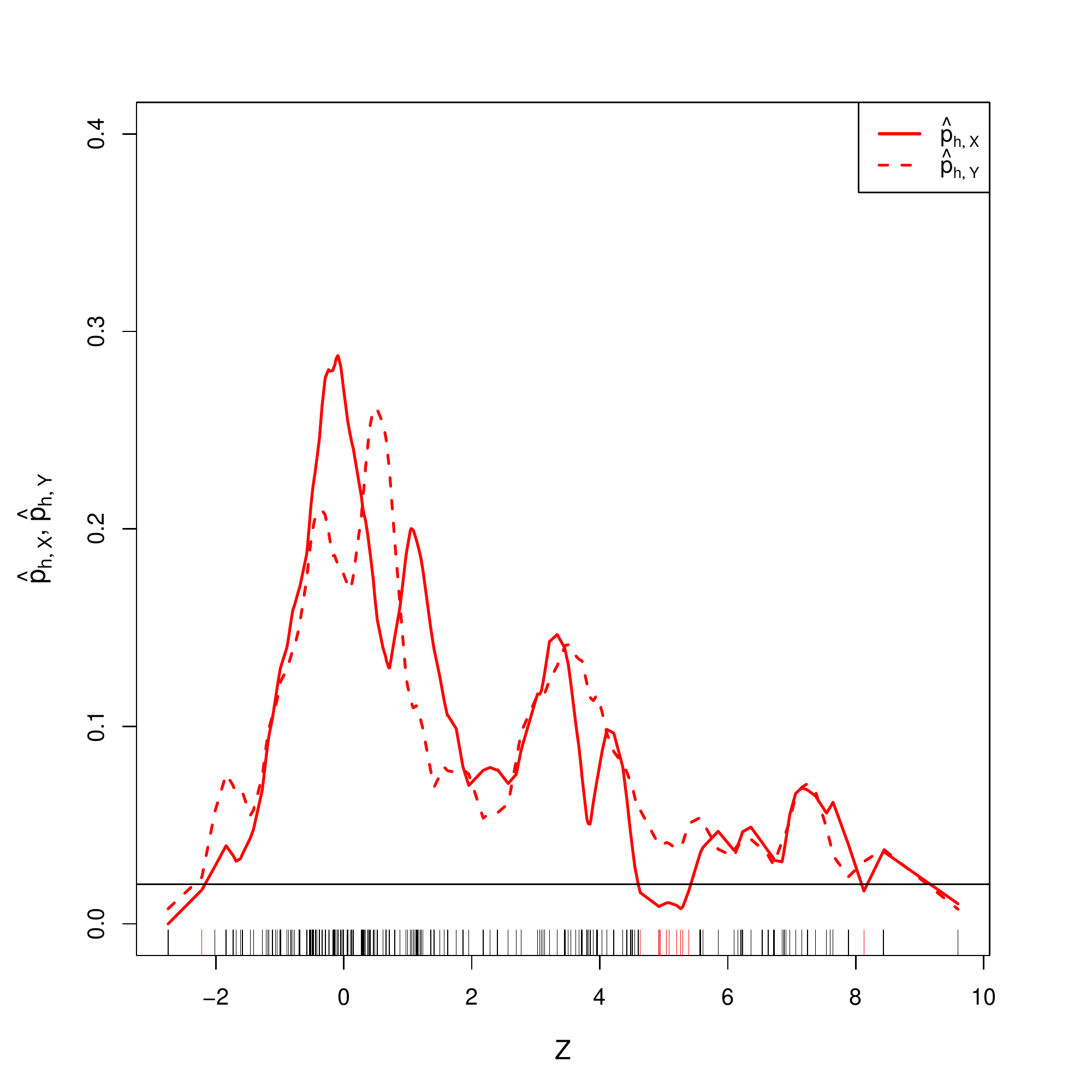}

\includegraphics[width=2.5in,height=2.5in]{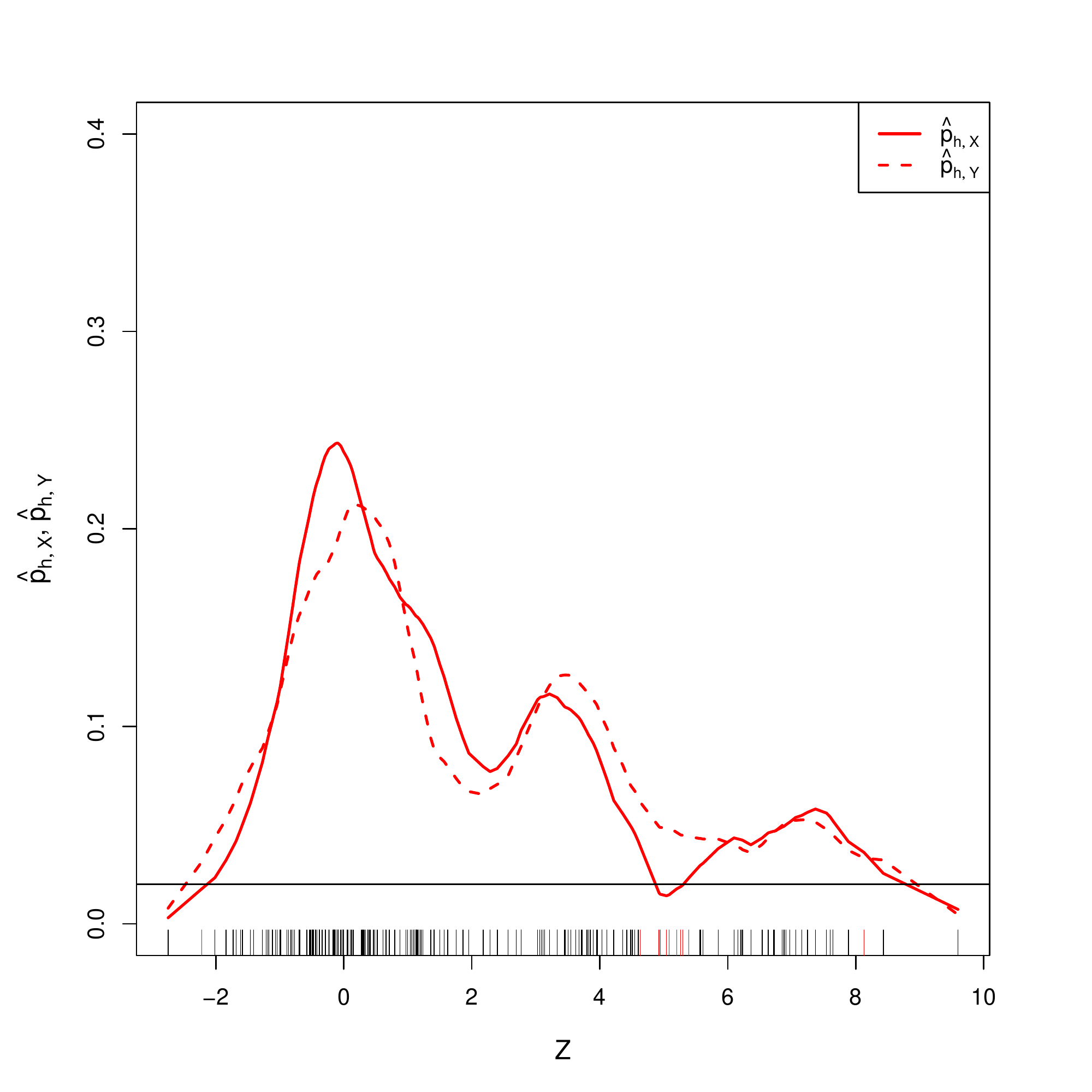}
\includegraphics[width=2.5in,height=2.5in]{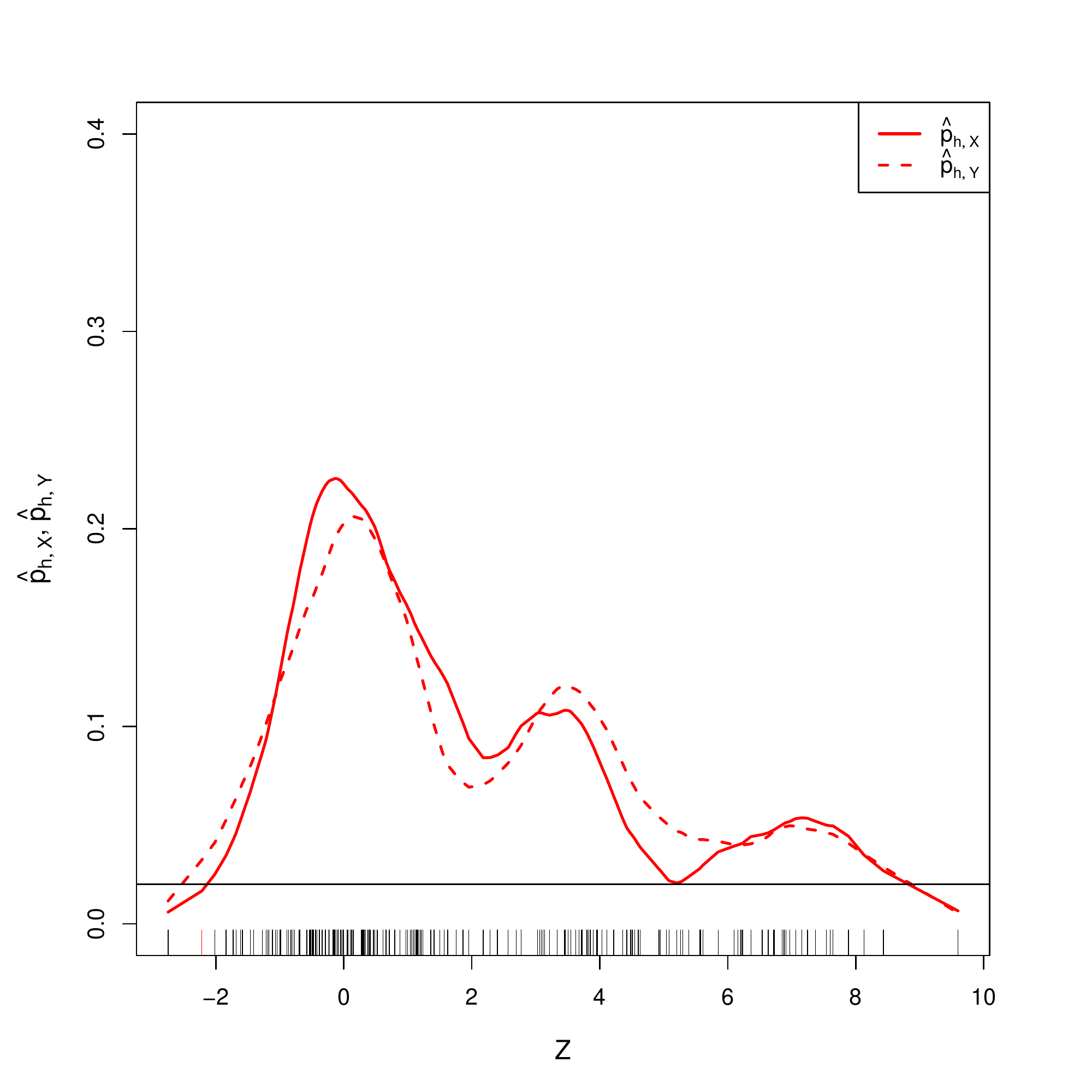}
\end{center}
\vspace{-5mm}
\caption{Comparing $\hat{L}_{h,X}(0.02)$ and $\hat{L}_{h,Y}(0.02)$ with $h = 0.15$ (top left), $h = 0.35$  (top right),  $h = 0.75$ (bottom left) and $h = 0.95$ (bottom right) for data sampled from the mixture distribution of Figure \ref{fig::intro}. The The two kernel density estimates are obtained using the $X$ sample (solid line) and the $Y$ sample (dotted line). Points in the $Z$ sample are showed as short vertical lines on the $x$-axis, and are colored in red when they belong to $\hat{L}_{h,X}(\lambda) \Delta \hat{L}_{h,Y}(\lambda)$.}
\label{fig::uni-lambda02-ex}
\end{figure}

\begin{figure}[ht]
\begin{center}
\includegraphics[width=2.5in,height=2.5in]{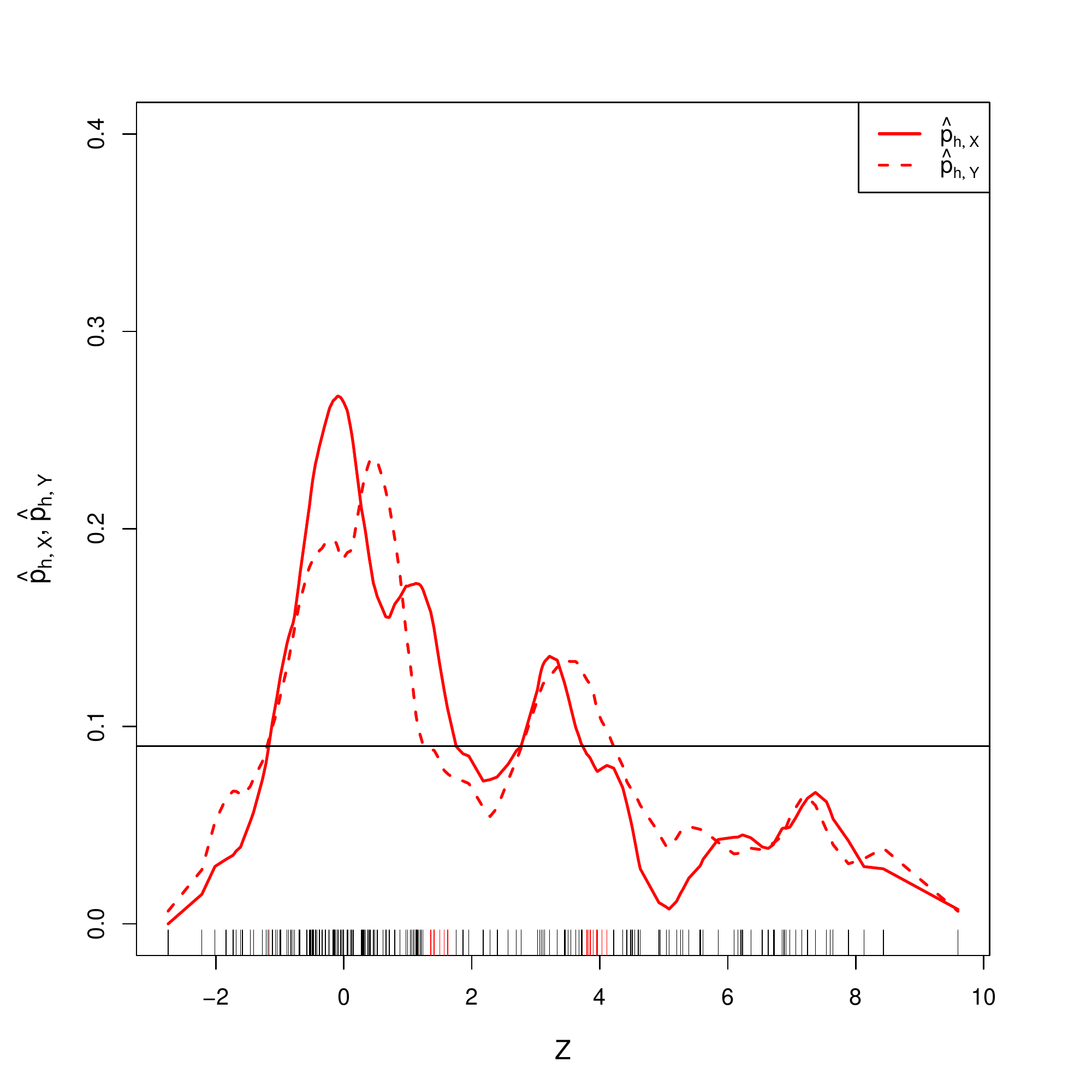}
\includegraphics[width=2.5in,height=2.5in]{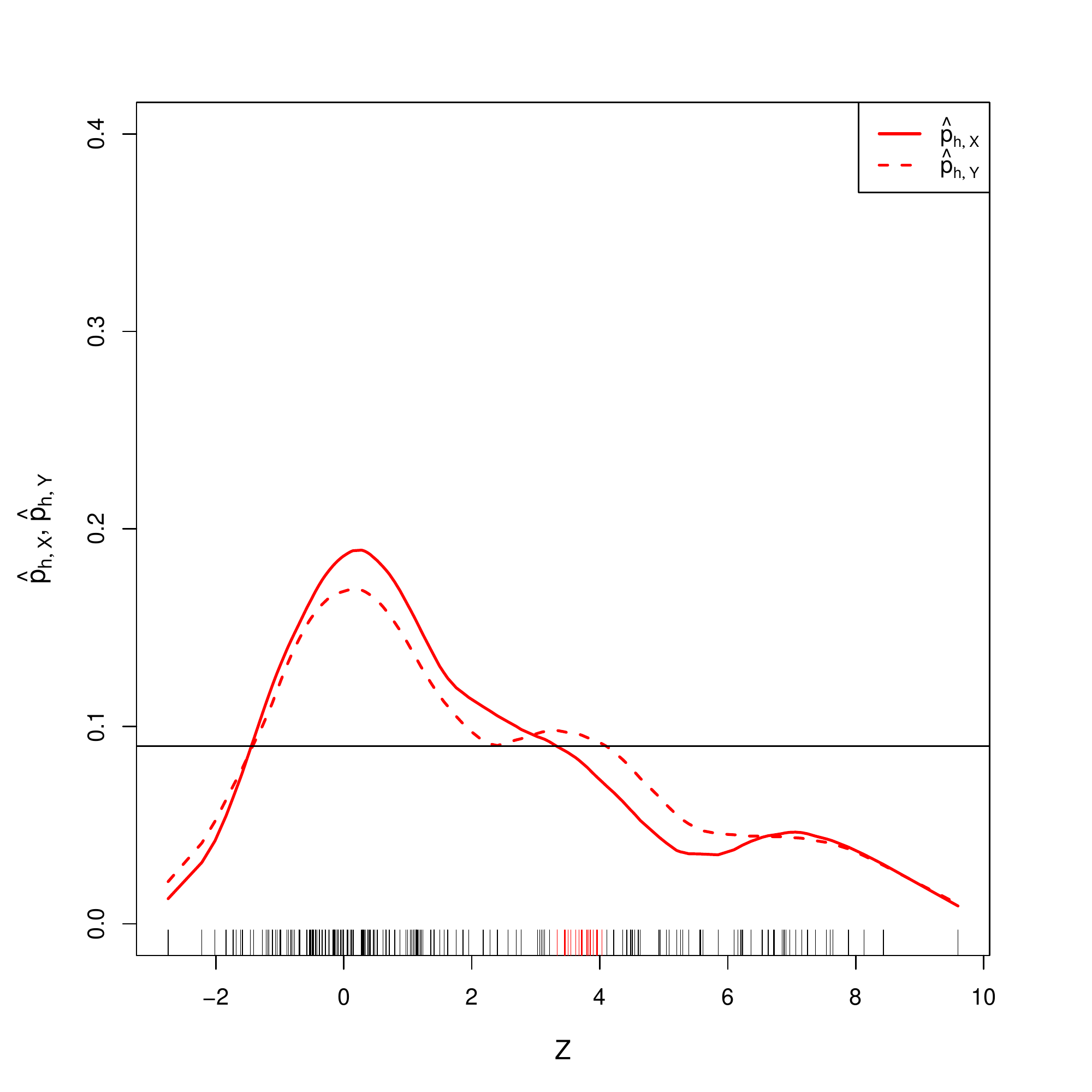}

\includegraphics[width=2.5in,height=2.5in]{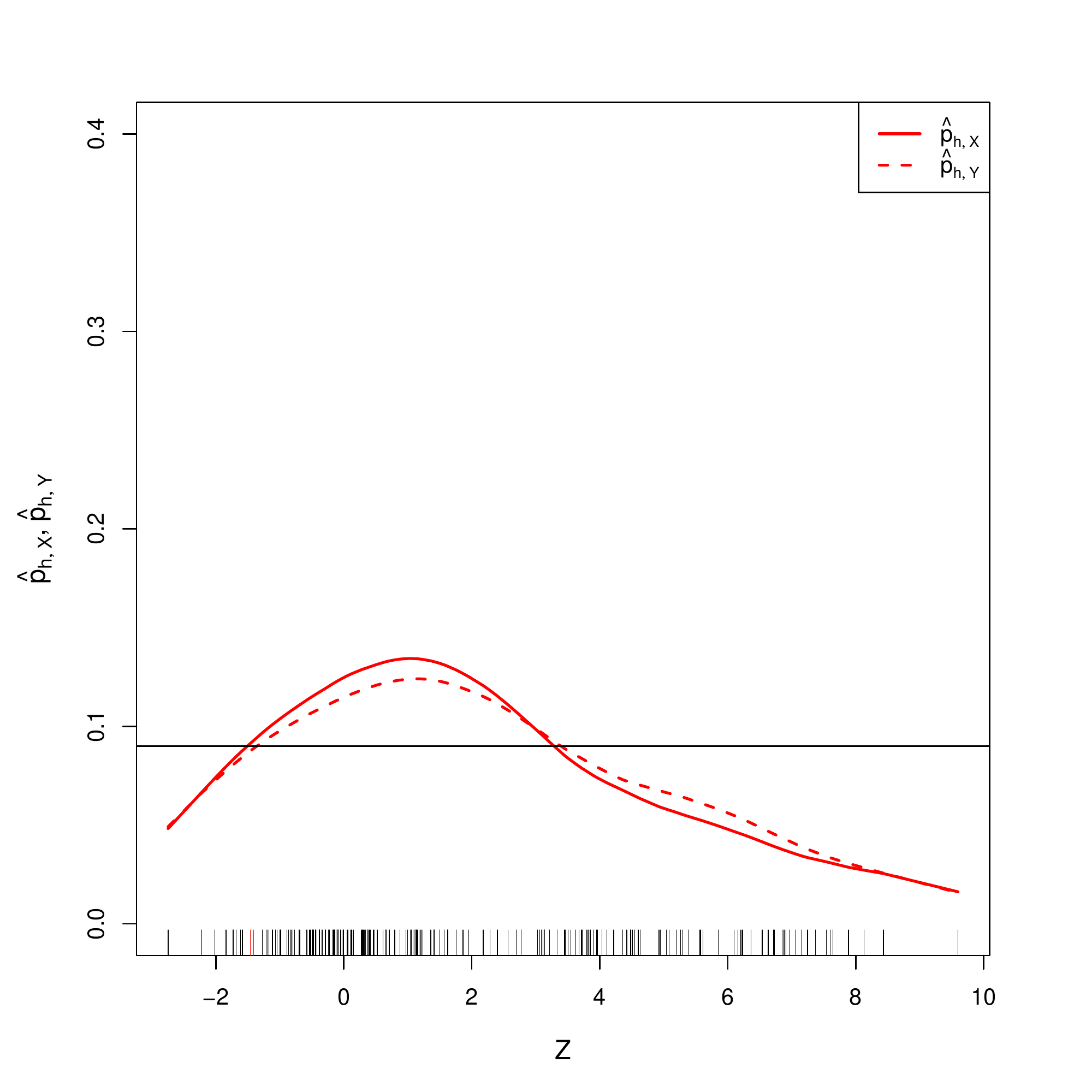}
\includegraphics[width=2.5in,height=2.5in]{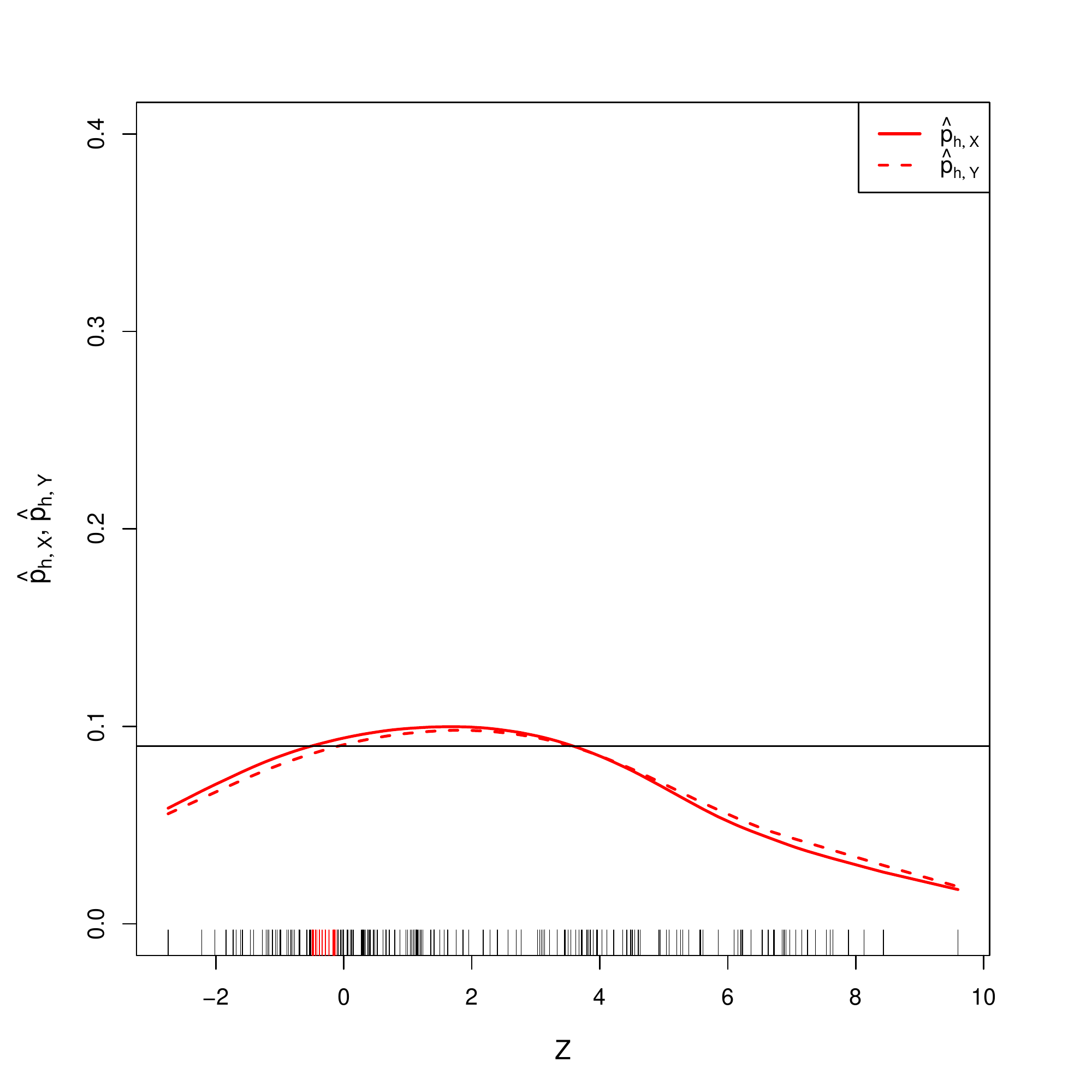}
\end{center}
\vspace{-5mm}
\caption{Comparing $\hat{L}_{h,X}(0.09)$ and $\hat{L}_{h,Y}(0.09)$ for $h = 0.5$ (top left), $h = 1.75$ (top right), $h = 3.75$ (bottom left) and $h = 6$ (bottom right) for data sampled from the mixture distribution of Figure \ref{fig::intro}. The The two kernel density estimates are obtained using the $X$ sample (solid line) and the $Y$ sample (dotted line). Points in the $Z$ sample are showed as short vertical lines on the $x$-axis, and are colored in red when they belong to $\hat{L}_{h,X}(\lambda) \Delta \hat{L}_{h,Y}(\lambda)$.}
\label{fig::uni-lambda09-ex}
\end{figure}

We present results for two examples where, although the dimensionality
is low, estimating the connected components of the true level sets is
surprisingly difficult.  For the first example, we begin by
illustrating how the instability changes for given values of $\lambda,
\alpha$ and then split each data set 200 times to find point-wise
confidence bands for $\Xi_n(h)$ for fixed $\lambda, \alpha$ and for
$\Gamma_n(h)$.  We then present selected results for a bivariate
example.

\subsection{Instability as function of fixed $\lambda$}
\label{subsection:examples-lambda}

Returning to the example distribution in Section \ref{section::intro},
600 observations were sampled from the following mixture of
normals: $(4/7)N(0,1)+(2/7)N(3.5,1)+(1/7)N(7,1)$.  The original sample
is randomly split into three samples of 200.  All kernel density
estimates use the Epanechikov kernel.  We examine the stability at
$\lambda = 0.02$, a height at which the true density's connected
components should be unambiguous, and $\lambda = 0.09$, the height
used in our earlier motivating graphs.

We start by conceptually illustrating the instability for selected values of $h$ in Figures \ref{fig::uni-lambda02-ex}, \ref{fig::uni-lambda09-ex}.  In each subfigure, $\hat{p}_{h,X}, \hat{p}_{h,Y}$ are graphed for the $Z$ set of observations.  Levels $\lambda = 0.02, 0.09$ are marked respectively with a horizontal line.  Those observations in $Z$ that belong to $\hat{L}_{h,X}(\lambda)$ and not to $\hat{L}_{h,Y}(\lambda)$ (or vice versa) are marked in red; the overall fraction of these observations is $\Xi_n(h)$.  In general, we can see that as $h$ increases, the number of the red $Z$ observations decreases.  For $\lambda = 0.02$, note that the location that most contributes to the instability is the valley around $Z = 5$.  Once $h$ is large enough to smooth this valley to have height above $\lambda = 0.02$, the instability is negligible.  Turning to $\lambda = 0.09$ (Figure \ref{fig::uni-lambda09-ex}), even for larger values of $h$, the differences between the two density estimates can be quite large.  When $h$ is large enough such that both density estimates lie entirely below $\lambda = 0.09$, our instability drops to and remains at zero.

\begin{figure}[t]
\begin{center}
\includegraphics[width=2.75in,height=2.75in]{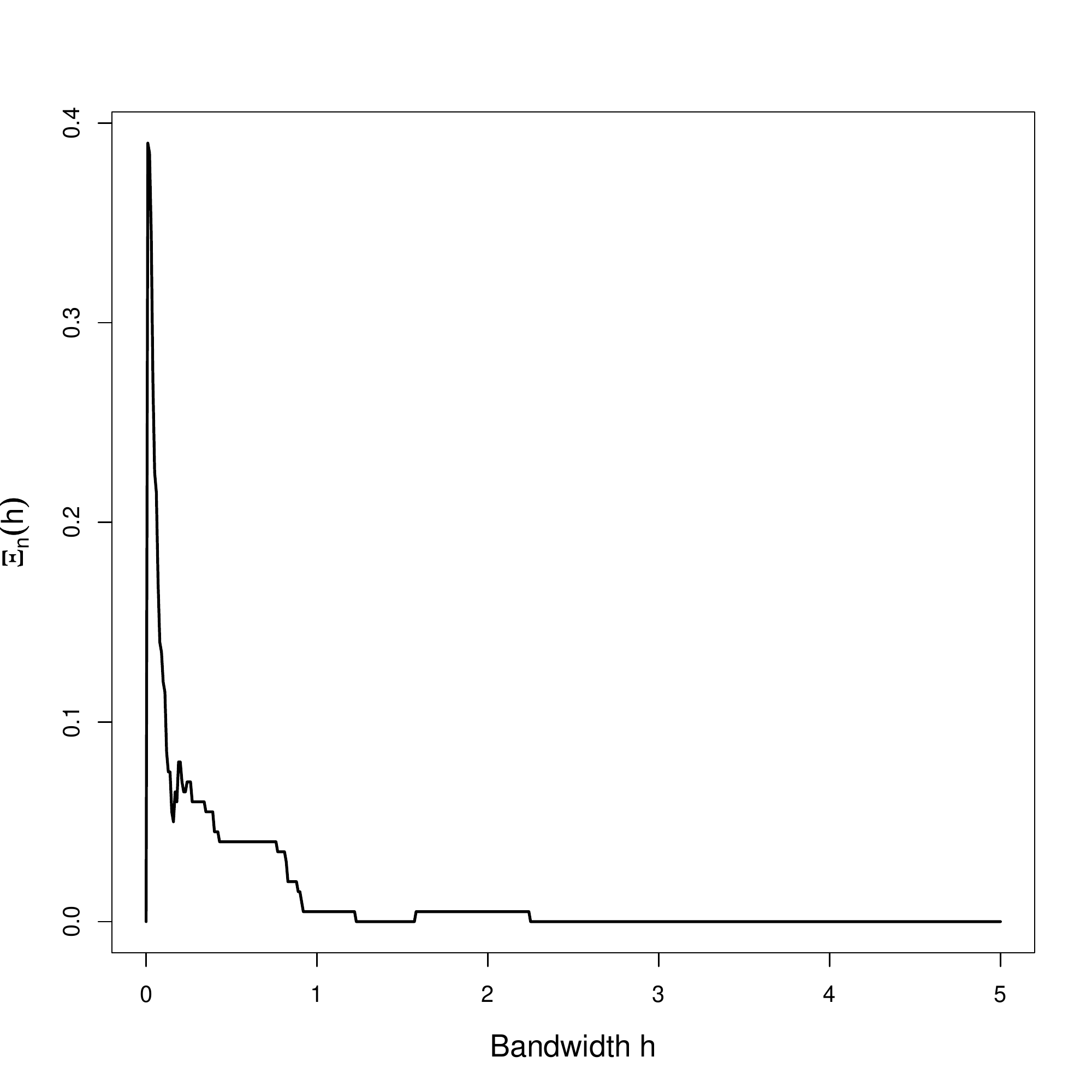}
\includegraphics[width=2.75in,height=2.75in]{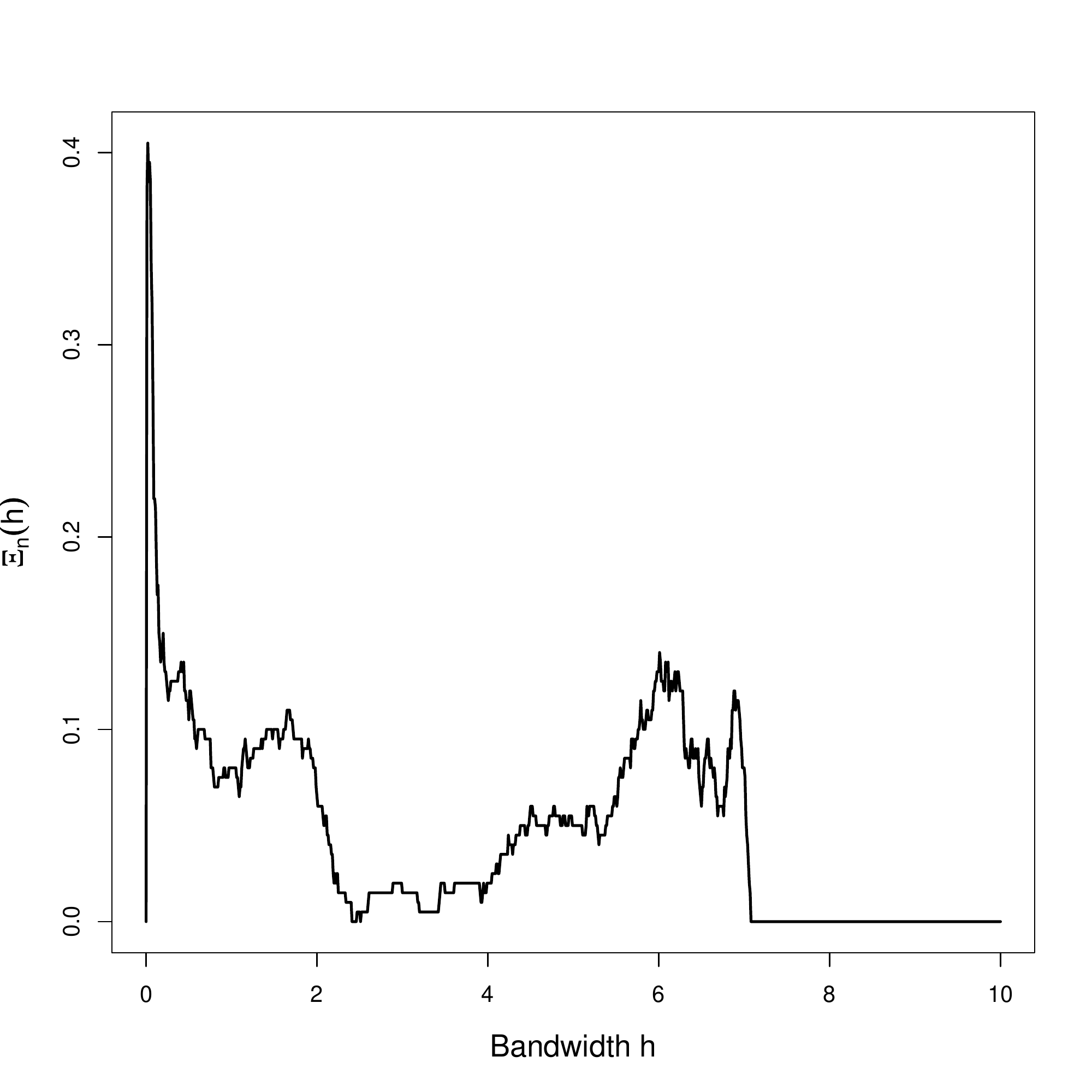}
\end{center}
\vspace{-3mm}
\caption{$\Xi_n(h)$ as a function of the bandwidth $h$ for $\lambda = 0.02$ (left) and $0.09$ (right) for data sampled from the mixture distribution of Figure \ref{fig::intro}.}
\label{fig::uni-lambda}
\end{figure}

Figure \ref{fig::uni-lambda} shows the overall behavior of $\Xi_n(h)$ as a function of $h$.  As expected, for $\lambda = 0.02$, $\Xi_n(h)$ jumps for the first non-zero $h$ and then quickly drops to almost zero by $h = 1$ (Figure \ref{fig::uni-lambda},  left).   At $\lambda = 0.09$, a height with a wide range of possible level sets (depending on the density estimate and the value of $h$), $\Xi_n(h)$ first drops and then oscillates as previously described as $h$ increases, indicating multi-modality (Figure \ref{fig::uni-lambda}, right). 

\subsection{Instability as function of probability content}
\label{subsection::examples-alpha}

\begin{figure}[t]
\begin{center}
\includegraphics[width=2.5in,height=2.5in]{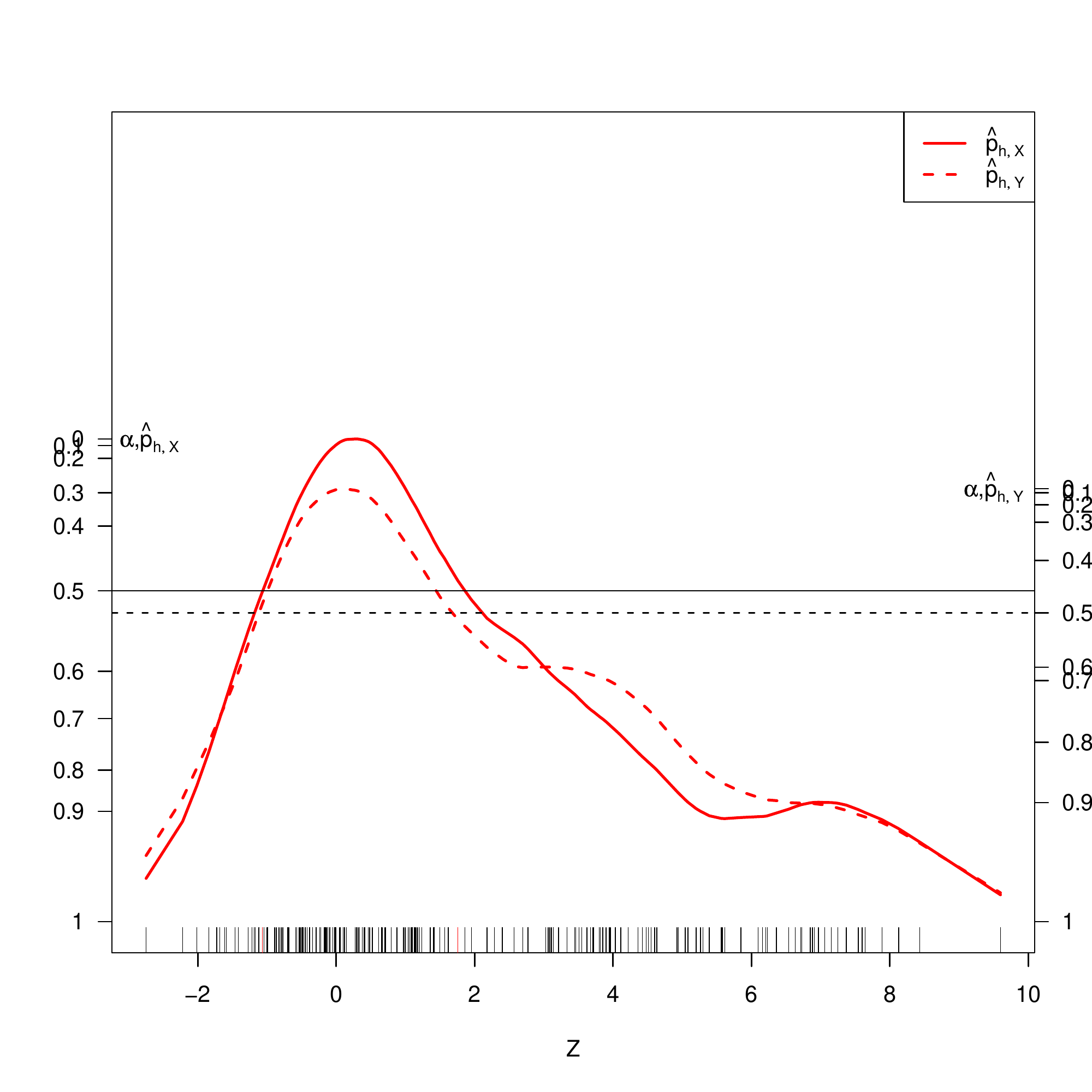}
\includegraphics[width=2.5in,height=2.5in]{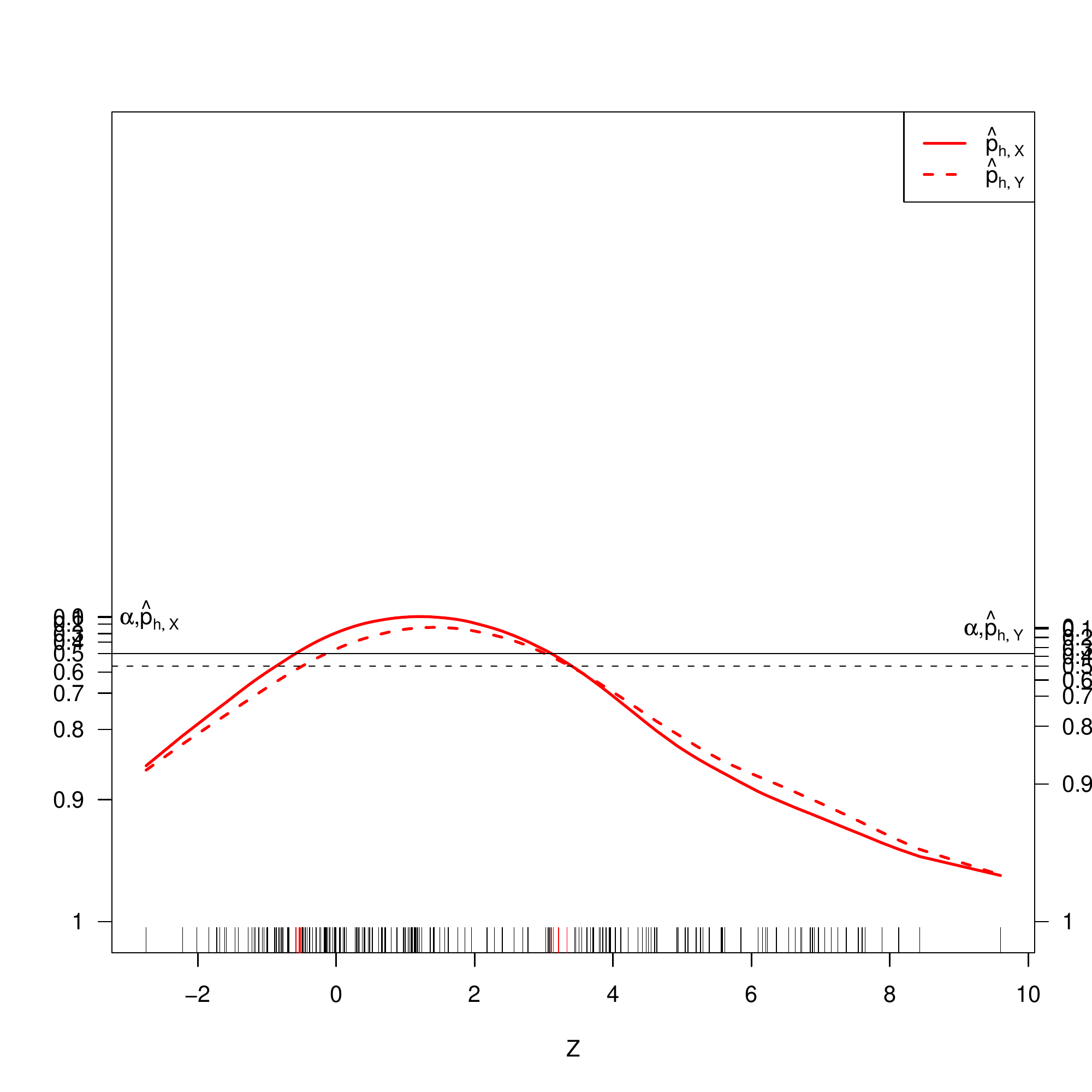}

\includegraphics[width=2.5in,height=2.5in]{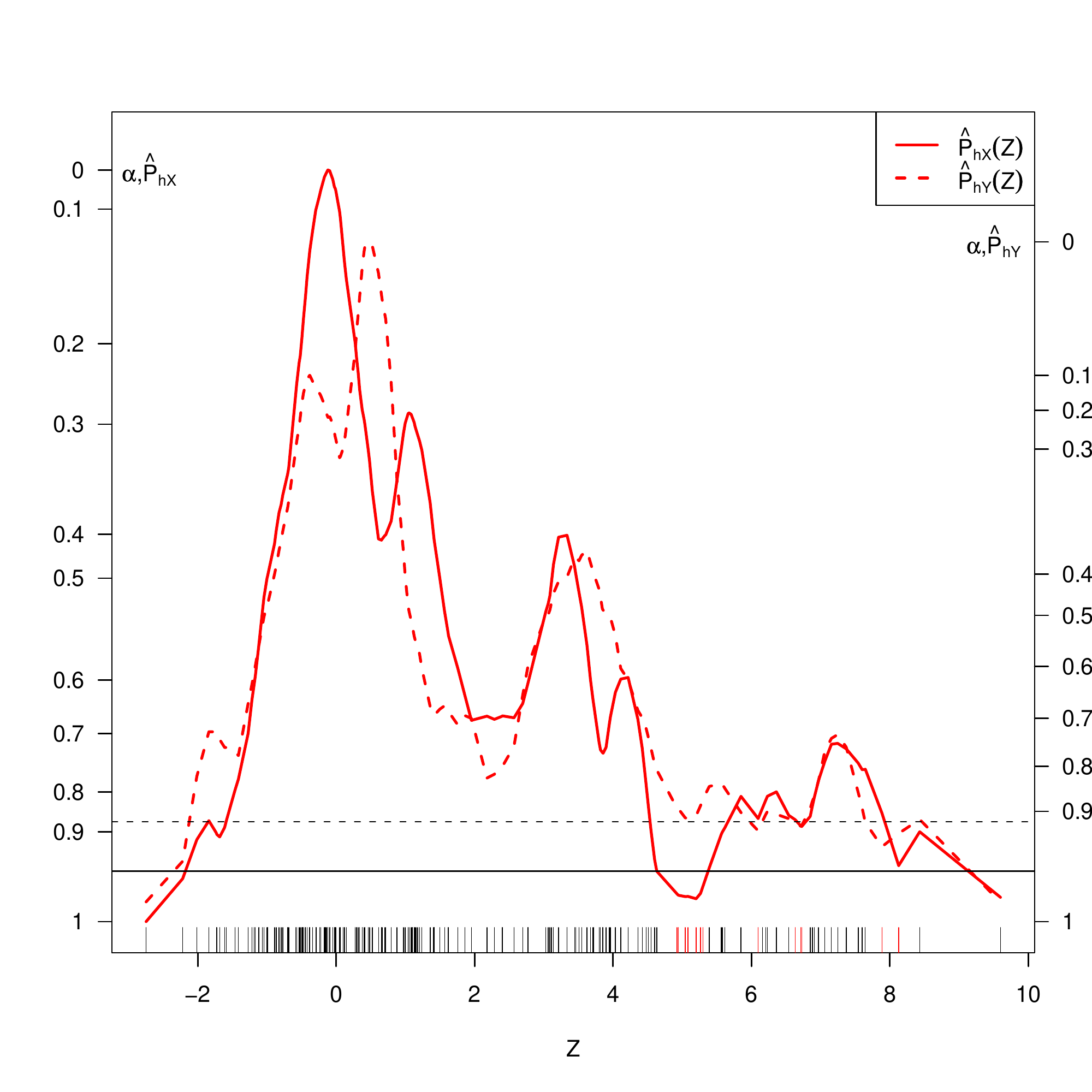}
\includegraphics[width=2.5in,height=2.5in]{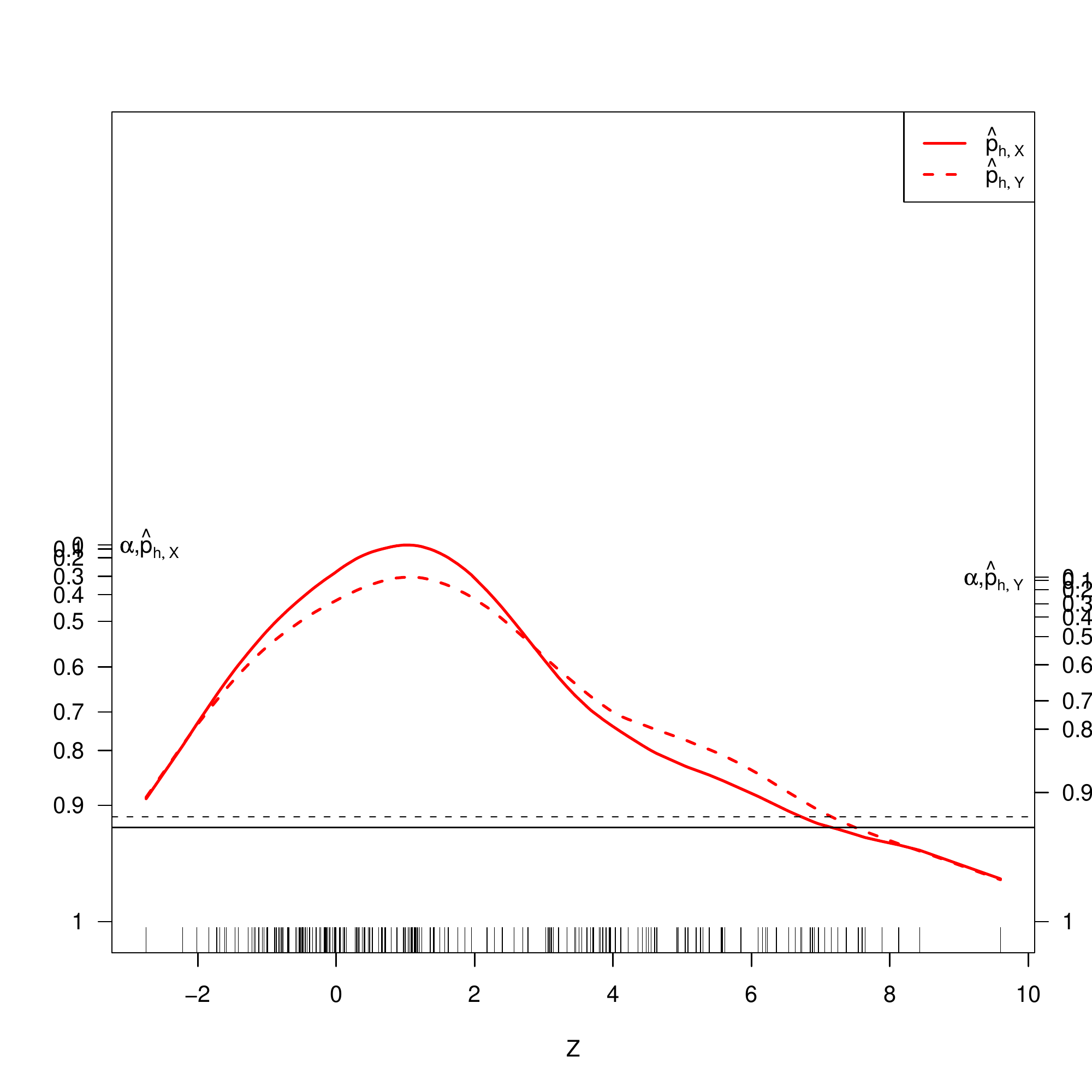}
\end{center}
\vspace{-5mm}
\caption{Top:  comparing $\hat{M}_{h,X}(0.50)$ and  $\hat{M}_{h,Y}(0.50)$ for $h = 2$ (left) and $h = 5$ right). Bottom:  comparing $\hat{M}_{h,X}(0.95)$ and $\hat{M}_{h,Y}(0.95)$ for $h = 0.4$ (left) and $h = 3.5$ (right). The data were sampled from the mixture distribution of Figure \ref{fig::intro}. The The two kernel density estimates are obtained using the $X$ sample (solid line) and the $Y$ sample (dotted line). Points in the $Z$ sample are showed as short vertical lines on the $x$-axis, and are colored in red when they belong to $\hat{M}_{h,X}(\alpha) \Delta \hat{M}_{h,Y}(\alpha)$.}\label{fig::alphacompare}
\end{figure}

In Section \ref{subsection::fixed-prob-content}, we defined $\Xi_n(h,\alpha)$, the sample instability as a function of $h$ and $\alpha$.  
As done before, we conceptually illustrate $\Xi_n(h,\alpha)$ for selected values of $h$ and $\alpha = 0.50$ and  $0.95$ in Figure \ref{fig::alphacompare}.
In each subfigure, $\hat{p}_{h,X}, \hat{p}_{h,Y}$ again are graphed for the $Z$ set of observations.  The probability content of the density estimates are respectively indicated on the left and right axes. The values $\alpha=0.50,0.95$ are also marked with solid and dashed horizontal lines for the two density estimates.  Those observations in $Z$ that belong to $\hat{M}_{h,X}(\alpha)$ and not to $\hat{M}_{h,Y}(\alpha)$ (or vice versa) are marked in red; the overall fraction of these observations is $\Xi_n(h,\alpha)$.  In general, we can see that as $h$ increases (for both values of $\alpha$), the number of red $Z$ observations decreases.  This decrease happens more quickly for higher values of $\alpha$ (as expected).

\begin{figure}[!h]
\begin{center}
\includegraphics[width=2.5in,height=2.5in]{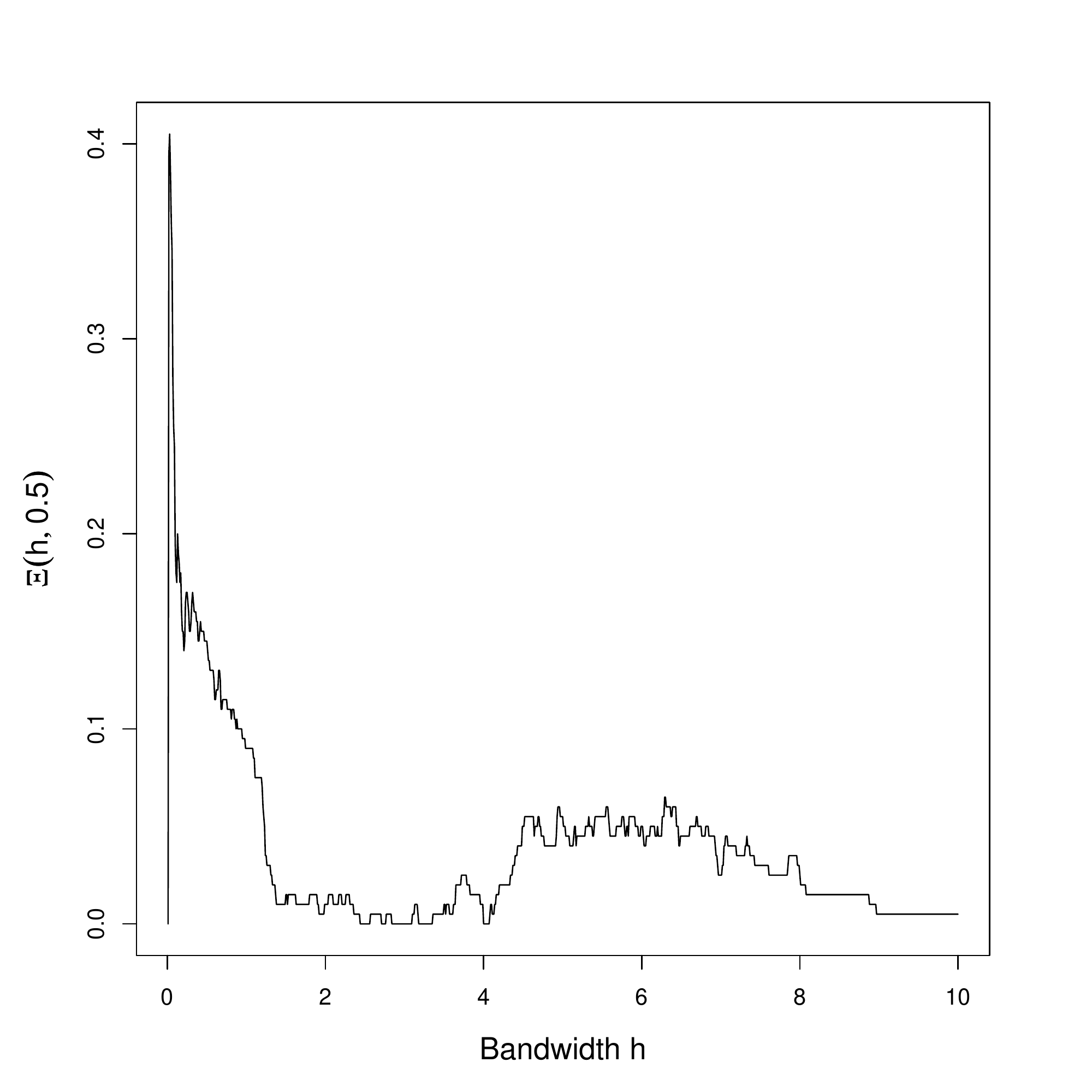}
\includegraphics[width=2.5in,height=2.5in]{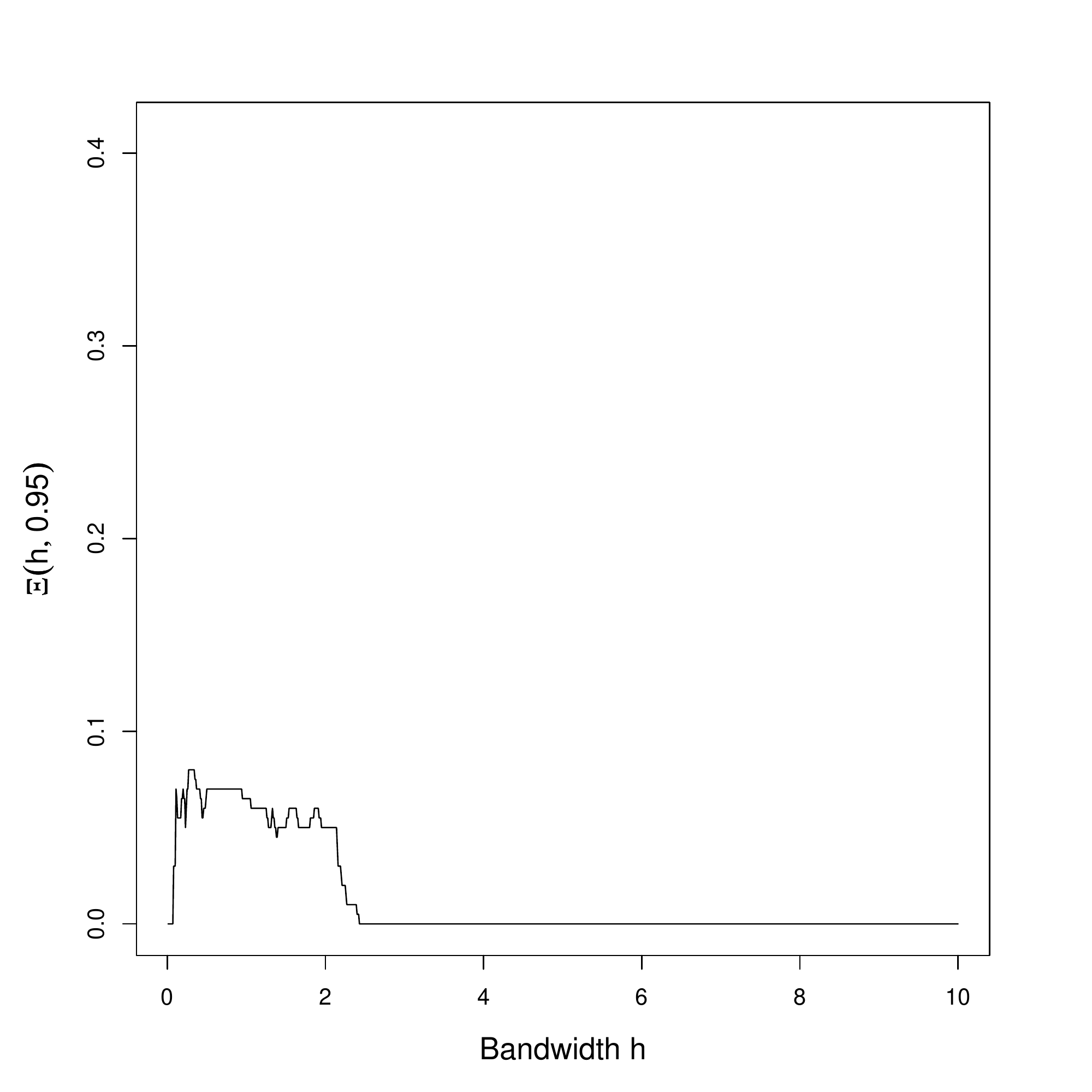}\\
\includegraphics[width=2.5in,height=2.5in]{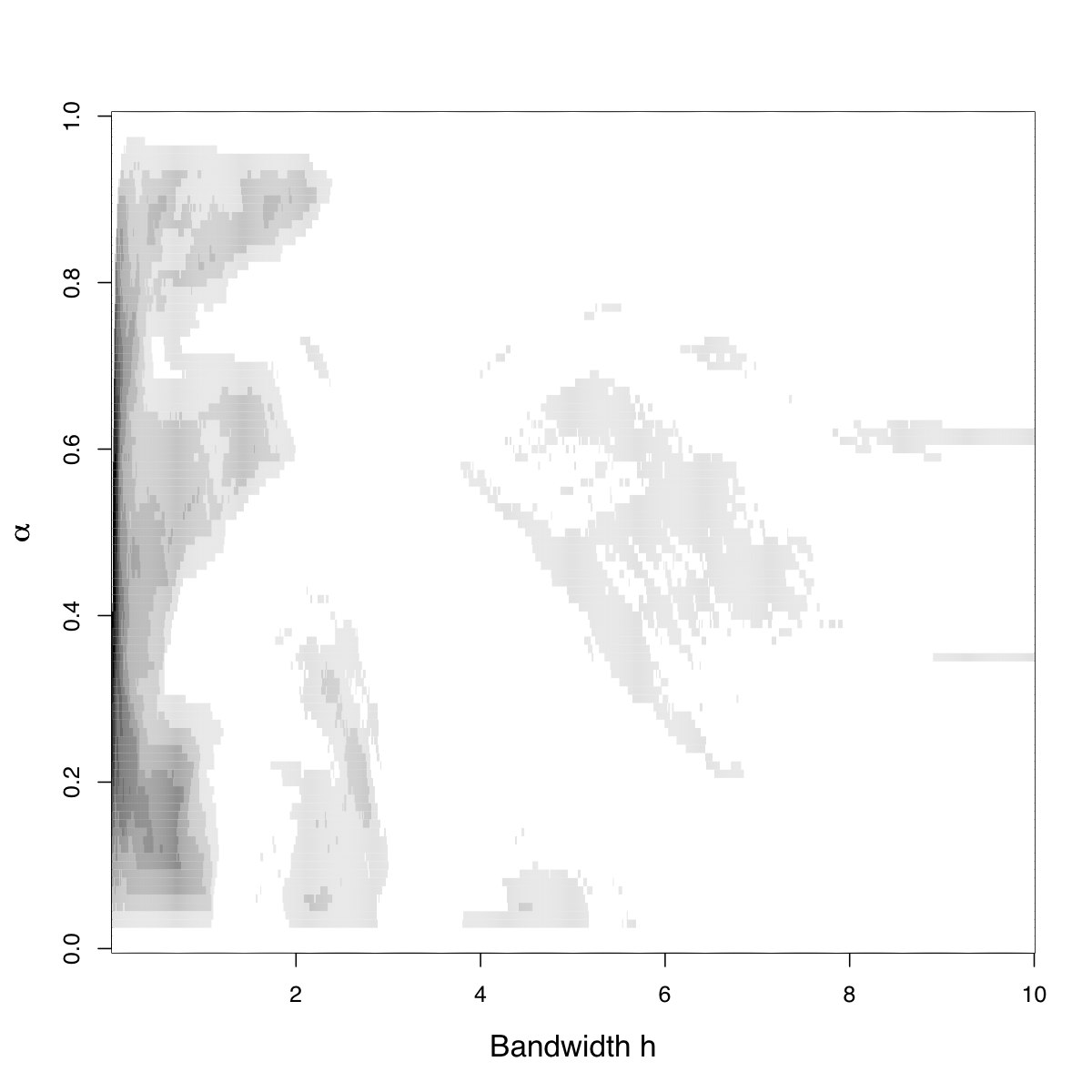}
\end{center}
\vspace{-5mm}
\caption{Top: $\Xi_n(h,\alpha=0.50)$ (left) and $\Xi_n(h,\alpha = 0.95)$ (right) as a function of $h$. Bottom: heat map of $\Xi_n(h,\alpha)$ as function of $h, \alpha$ for the example of Figure \ref{fig::intro}. The data were sampled from the mixture distribution of Figure \ref{fig::intro}.}
\label{fig::alpha}
\end{figure}

In Figure \ref{fig::alpha}, we examine $\Xi_n(h, \alpha)$ as a function of $h$ for $\alpha = 0.50, 0.95$.  For level sets that contain at least 50$\%$ probability content, i.e. $\hat{M}_{h,X}(0.50)$, the instability quickly drops as $h$ increases and then oscillates as $h$ approaches values that correspond to density estimates with uncertainty at those levels.  Again, this ambiguity occurs due to the presence of the second mode (we would see similar behavior with respect to the smallest mode if $\alpha \approx 0.80$).  As $h$ continues to increase, the density estimates become smooth enough that there is very little difference between $M_{h,X}(0.50)$, $M_{h,Y}(0.50)$.  This behavior also occurs when $\alpha = 0.95$ albeit more quickly (Figure \ref{fig::alpha}, top right) since level sets that contain at least 95$\%$ probability content occur at lower heights and are more stable.

Figure \ref{fig::alpha}c is the corresponding heat map for $\alpha = 0, 0.01, \ldots, 1.0$ and $h = 0, 0.01, \ldots,10$.  White sections indicate $\Xi_n(h,\alpha) \approx 0$; black sections indicate higher instability values.  In this particular example, the maximum instability of 0.425 is found at $h = 0.03, \alpha = 0.46$.  Note that around $h = 3$, we have very low instability values for almost all values of $\alpha$, and 
hence this value of kernel bandwidth would be a good choice that yields stable clustering.

\subsection{Instability Confidence Bands}
\label{subsection:examples-confbands}

\begin{figure}[!ht]
\begin{center}
\includegraphics[width=2.5in,height=2.5in]{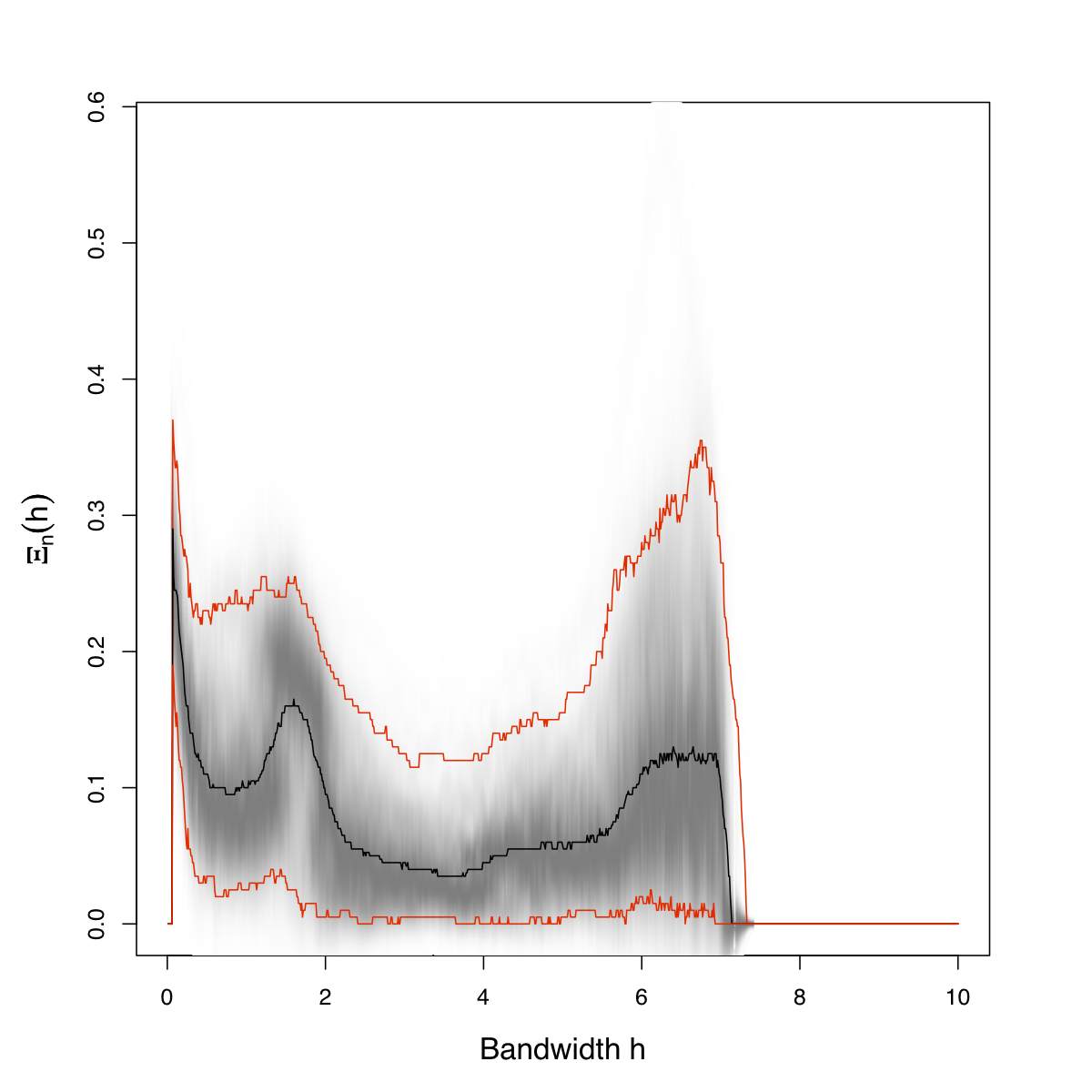}
\includegraphics[width=2.5in,height=2.5in]{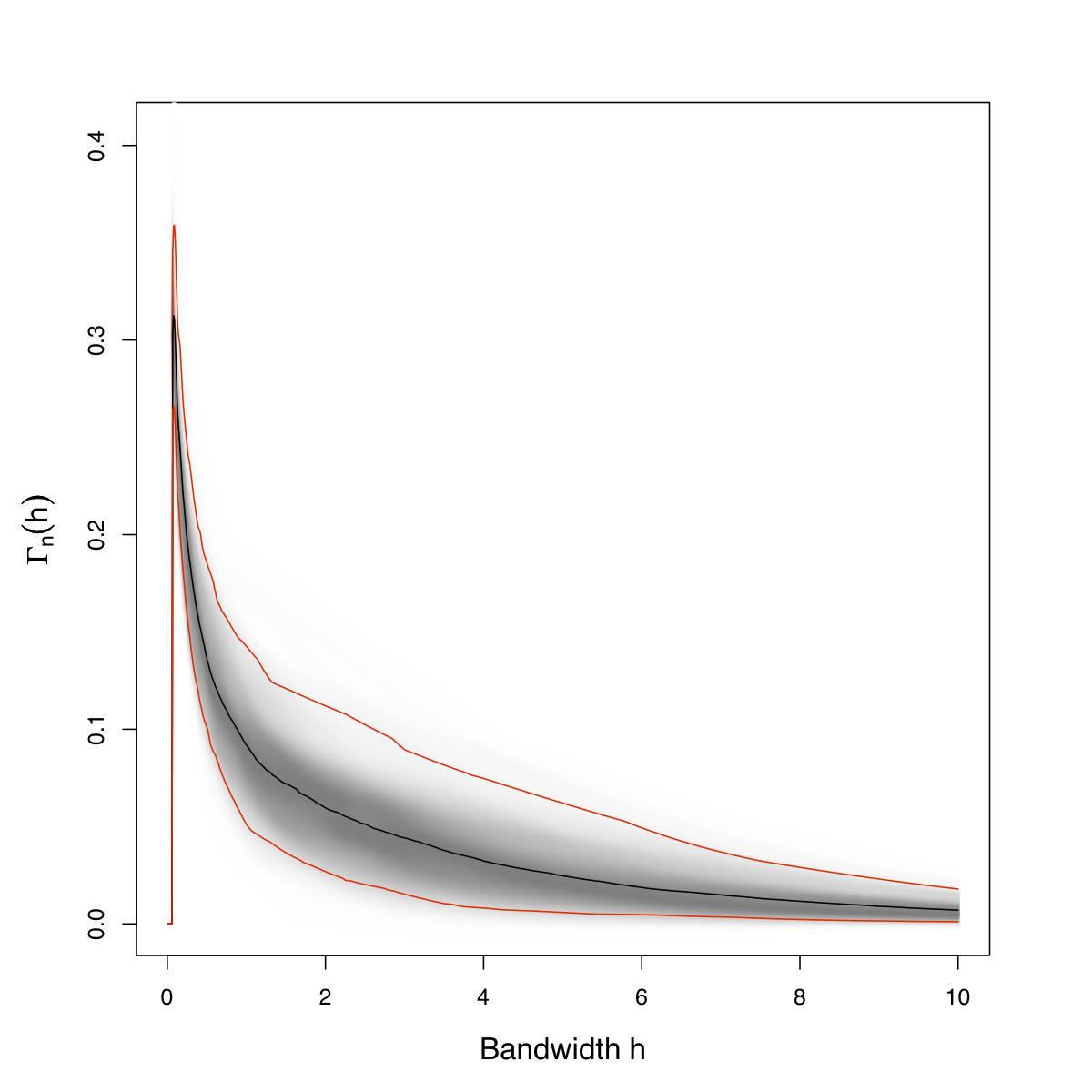}

\includegraphics[width=2.5in,height=2.5in]{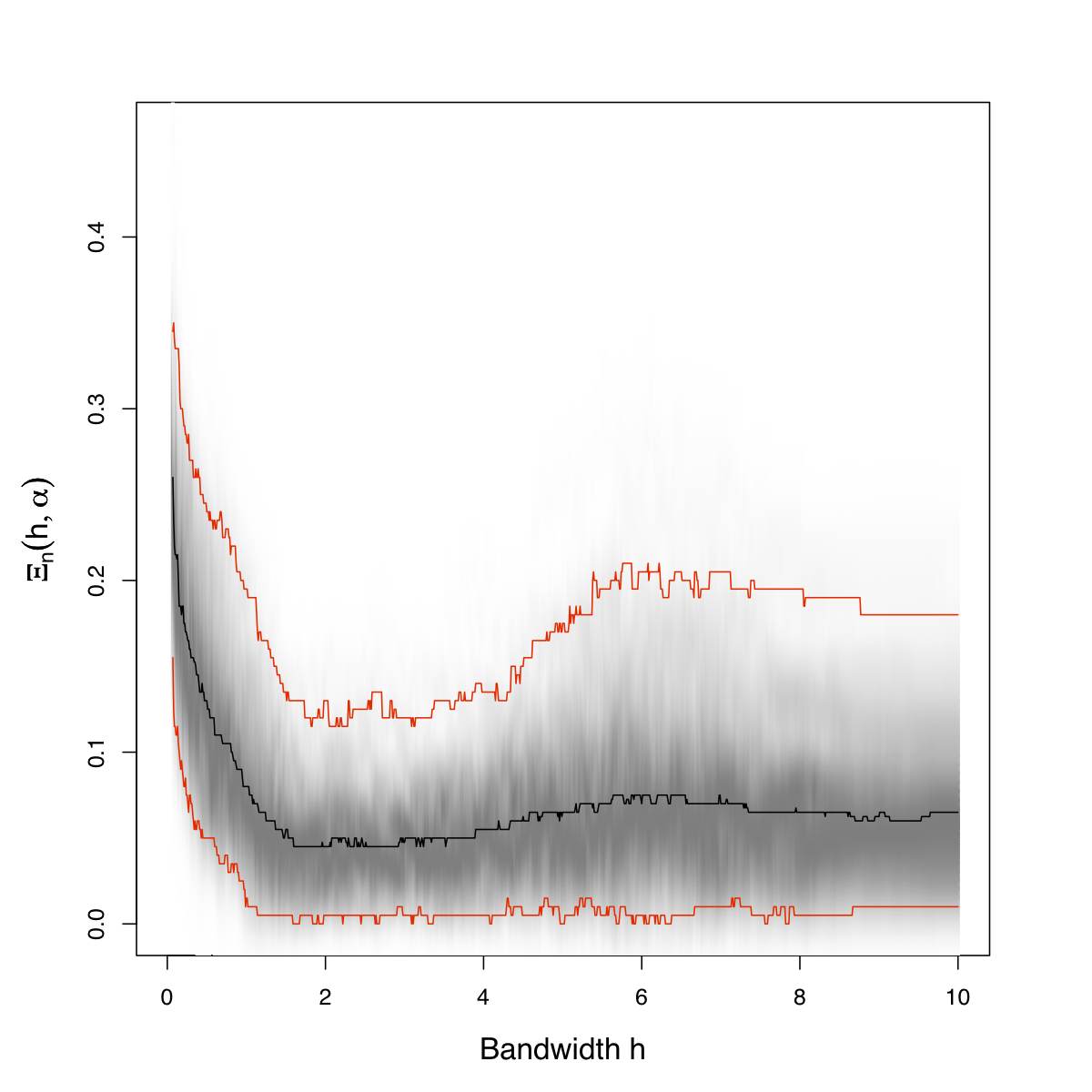}
\includegraphics[width=2.5in,height=2.5in]{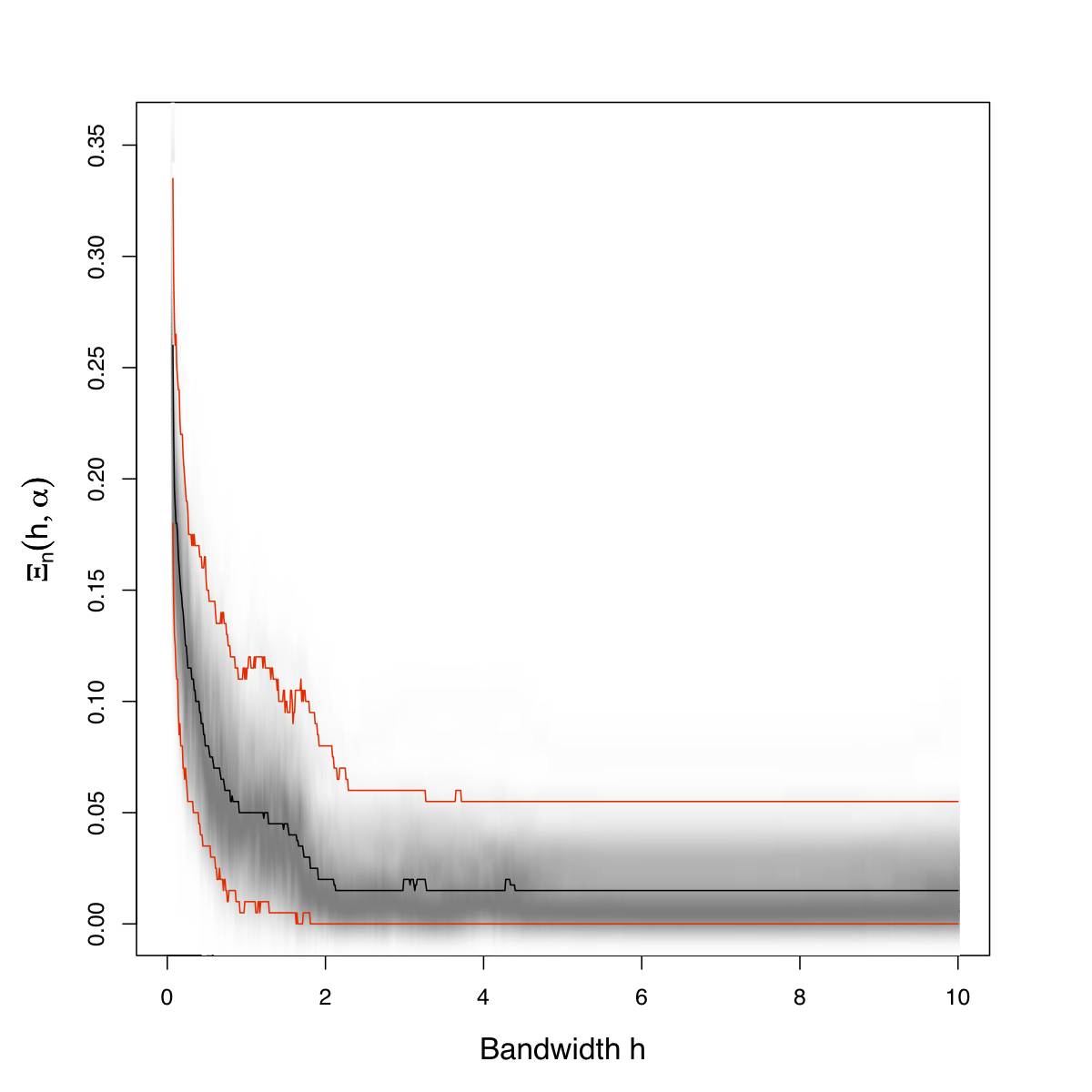}
\end{center}
\vspace{-5mm}
\caption{95$\%$ point-wise confidence bands for $\Xi_n(h)$ (top left), $\Gamma_n(h)$ (top right), $\Xi_n(h,\alpha = 0.50)$ (bottom left) and $\Xi_n(h,\alpha = 0.95)$ (bottom right) for data sampled from the mixture distribution of Figure \ref{fig::intro}.}
\label{fig::confbands}
\end{figure}

The results in the previous subsections were for splitting the original sample one time into three groups of 200 observations.  Here we briefly include a snapshot of what the distribution of our instability measures look like over repeated splits.  For computational reasons, we used the binned kernel density estimate, again with the Epanechikov kernel, and discretize the feature space over 200 bins; see \cite{wand}.  Increasing the number of bins improves the approximation to the kernel density estimate; the use of two hundred bins was found to give almost identical results to the original kernel density estimate (results not shown).  We split the original sample 200 times and find 95$\%$ point-wise confidence intervals for $\Xi_n(h)$, $\Gamma_n(h)$, and $\Xi_n(h,\alpha)$ for $\alpha = 0.50, 0.95$ and as a function of $h$. The results are depicted in Figure \ref{fig::confbands}.  The confidence bands are plotted in red, the medians in black.  The distribution of the instability measures for each value of $h$ is also plotted using density strips \citep[see][]{jackson}; on the grey-scale, darker colors indicate more common instability values.  The density strips allow us to see how the distribution changes (not just the 50, 95$\%$ percentiles). For example, for the plot on the top left in Figure \ref{fig::confbands}, note that right before $h=2$, the upper half of the distribution of $\Xi_n(h)$ is more concentrated.  This shift corresponds to the increase in instability in the presence of the additional modes.

\subsection{Bivariate Moons}
\label{subsection::moons}

\begin{figure}[!ht]
\begin{center}
\includegraphics[width=2.5in,height=2.5in]{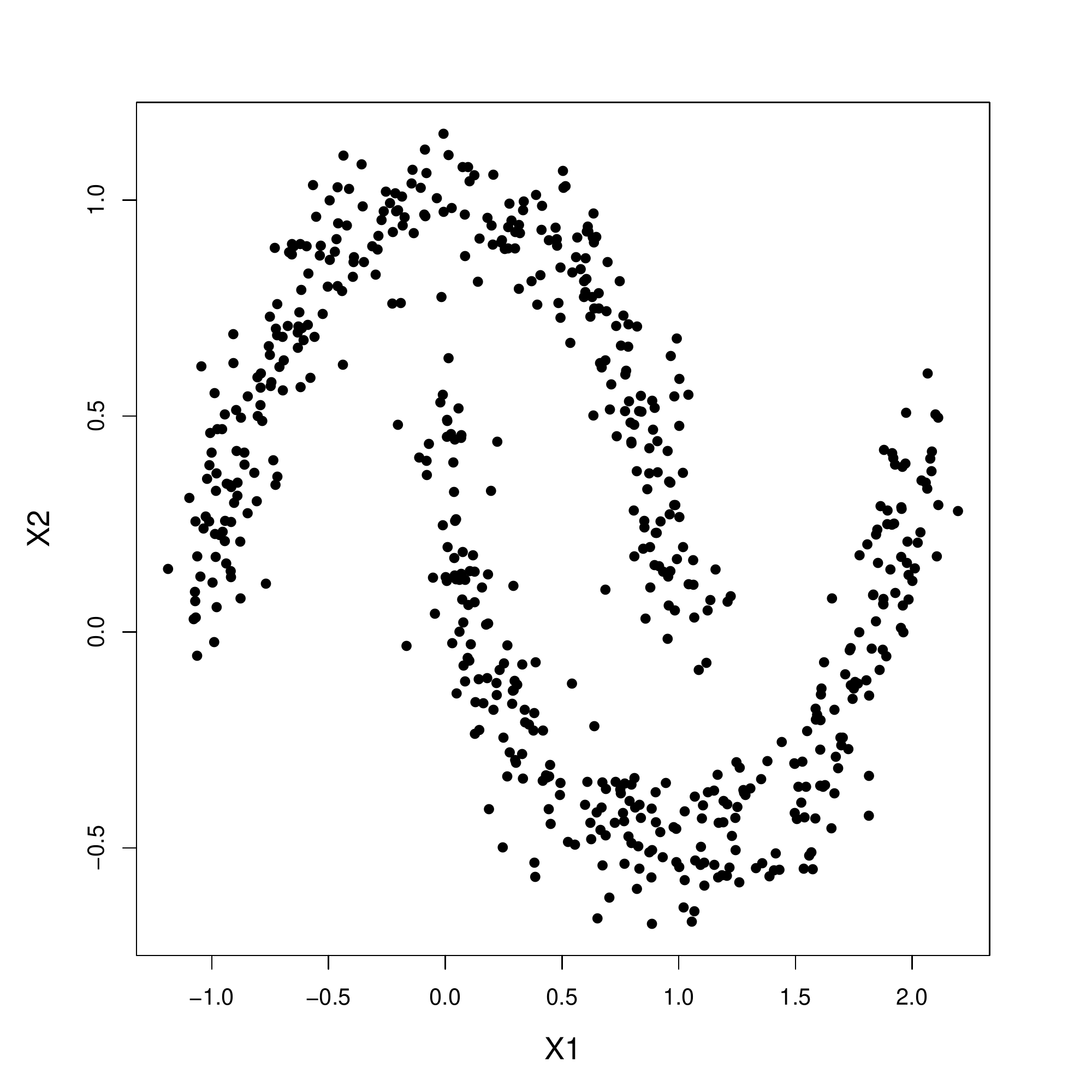}
\includegraphics[width=2.5in,height=2.5in]{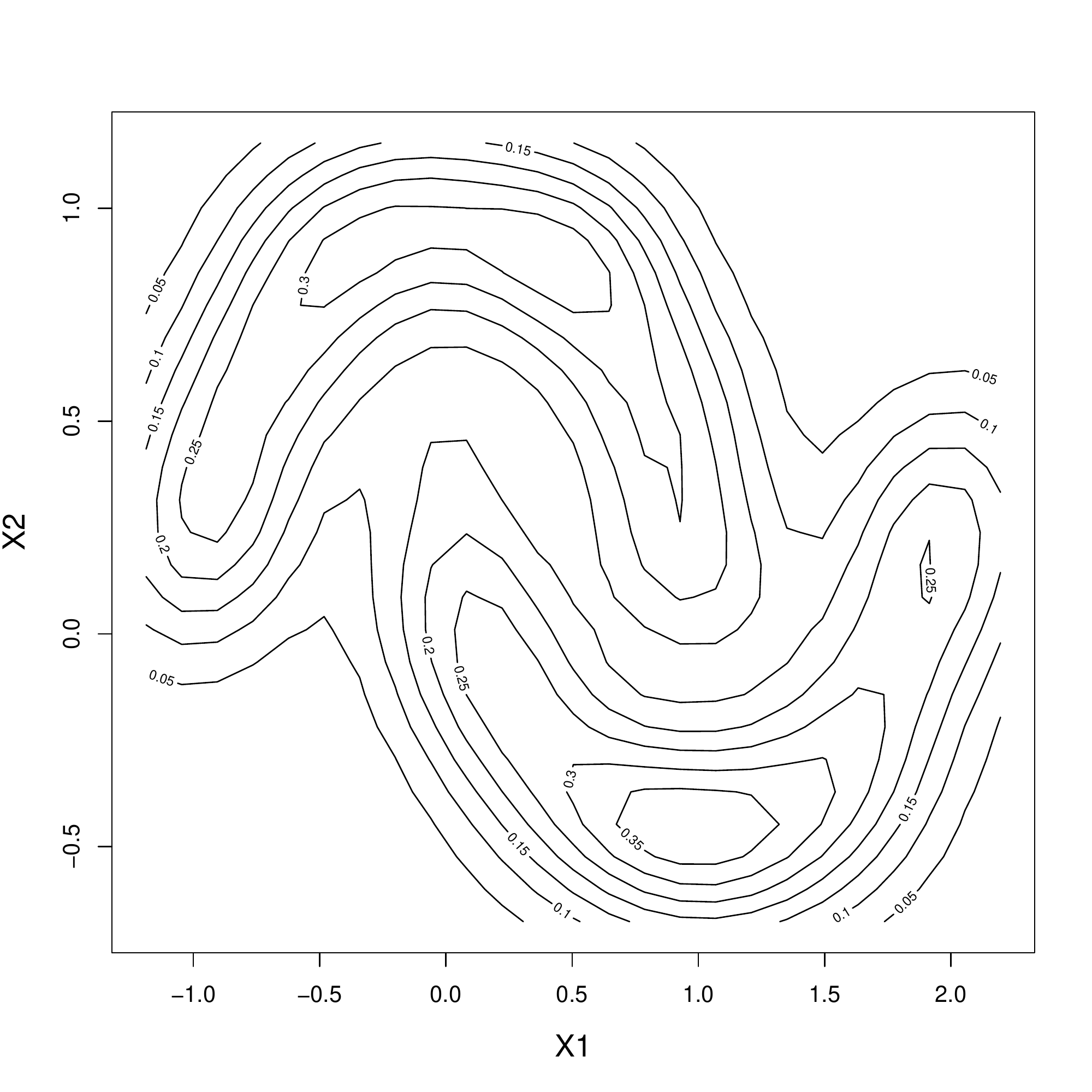}
\end{center}
\vspace{-5mm}
\caption{Bivariate moons (left) and contours of a Gaussian kernel density estimate (right) for the example discussed in Section \ref{subsection::moons}.}
\label{fig::moons}
\end{figure}

We also include a bivariate example with two equal-sized moons; this data set with seemingly simple structure can be quite difficult to analyze.  The scatterplot of the data on the left in Figure \ref{fig::moons} show two clusters, each shaped like a half moon. Each cluster contains 300 data points.  The plot on the right in Figure \ref{fig::moons}b shows a two-dimensional kernel density estimate (for illustrative purposes) using a Gaussian kernel with default bandwidth and evaluation points.  We can see that while levels around $\lambda = 0.30$ show clear multi-modality, the connectedness of the level sets around $\lambda = 0.15$ are less clear.

\begin{figure}[!ht]
\begin{center}
\includegraphics[width=2.5in,height=2.5in]{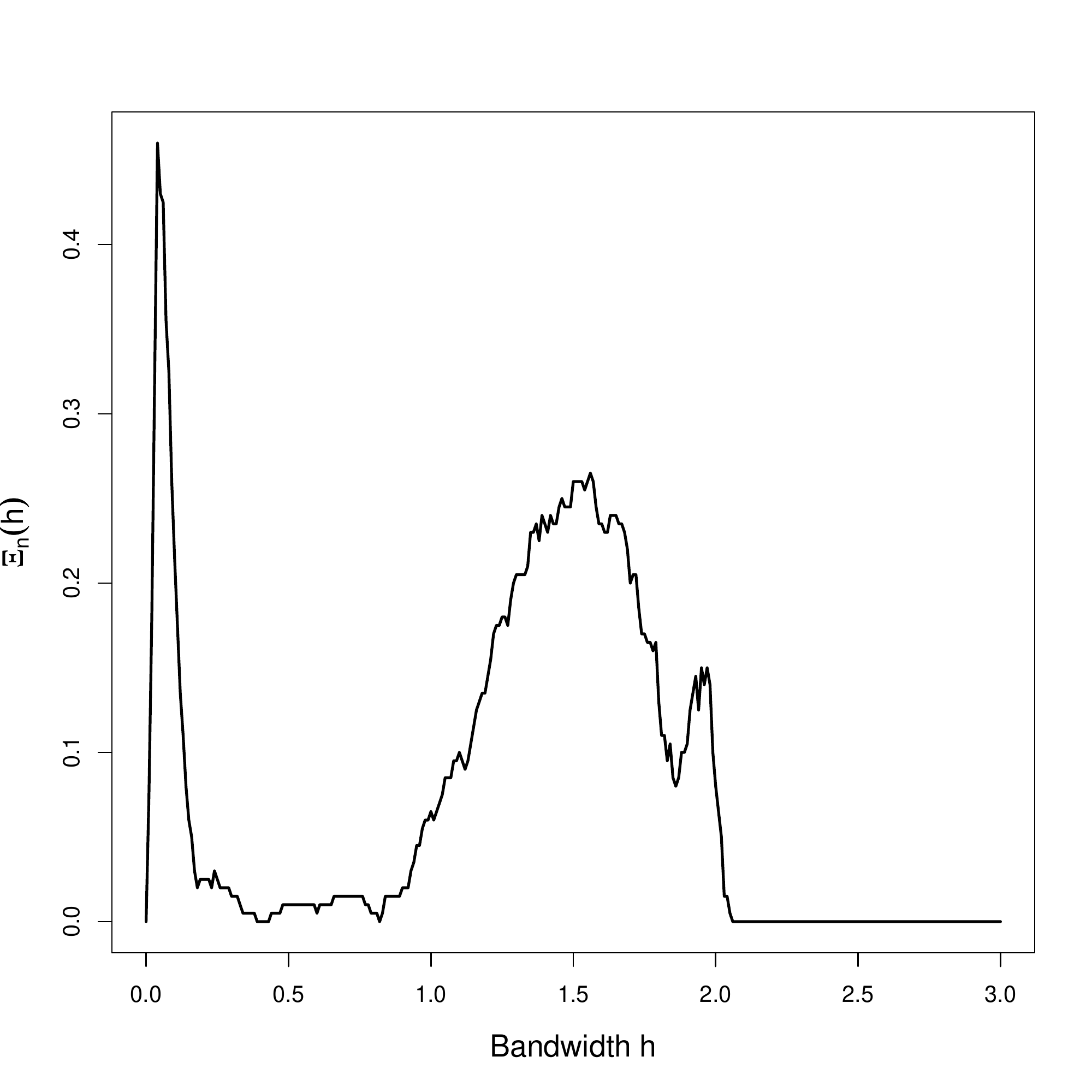}
\includegraphics[width=2.5in,height=2.5in]{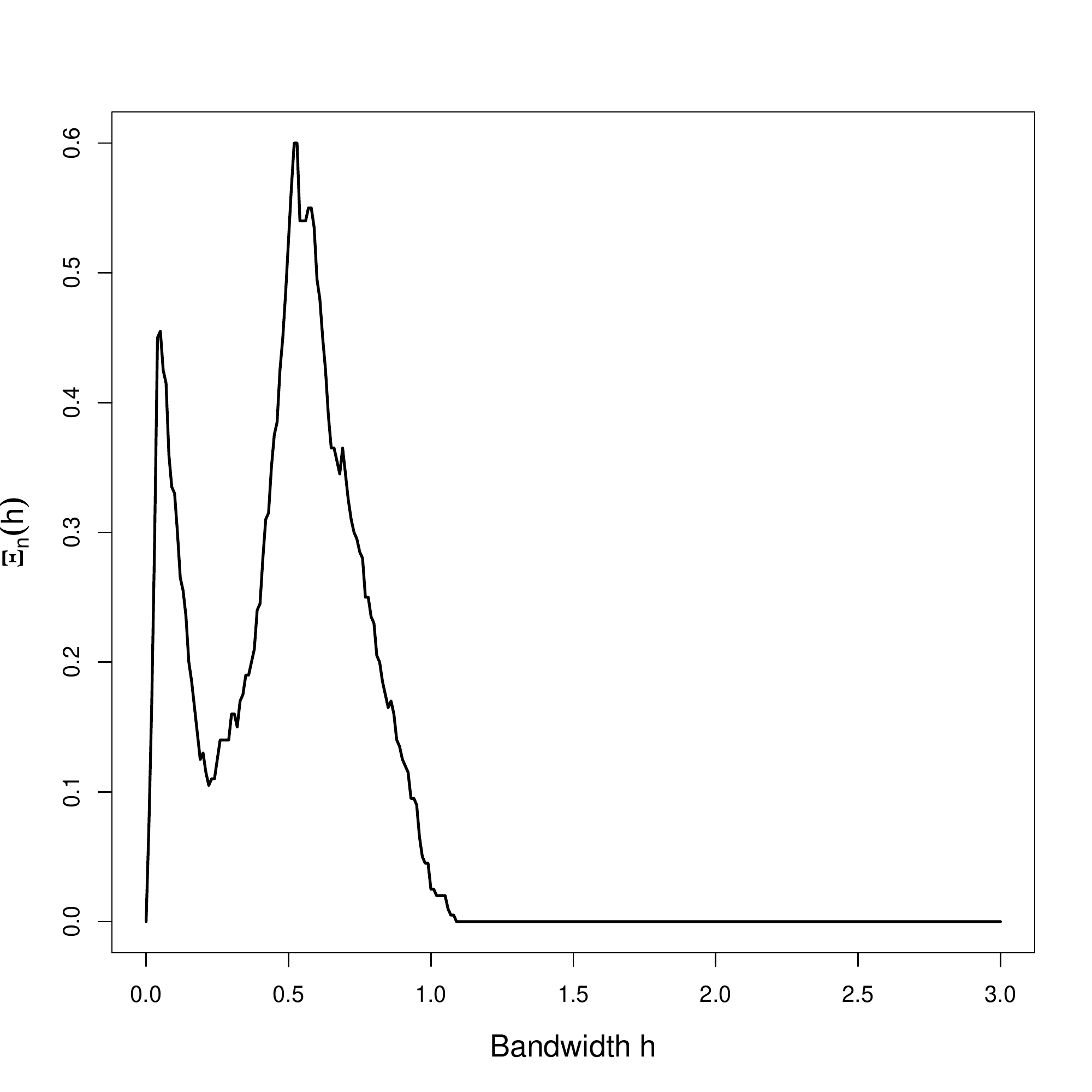}

\includegraphics[width=2.5in,height=2.5in]{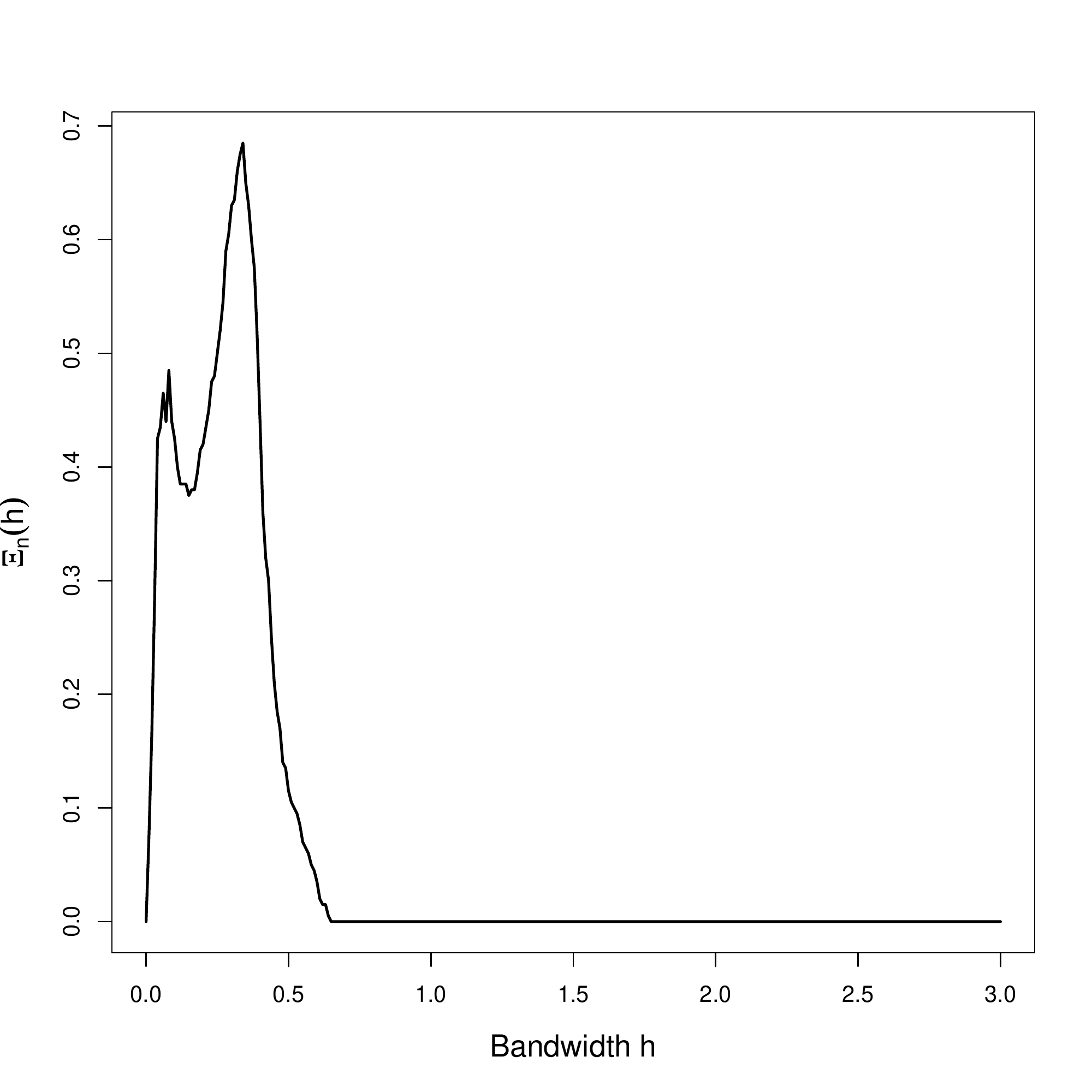}
\includegraphics[width=2.5in,height=2.5in]{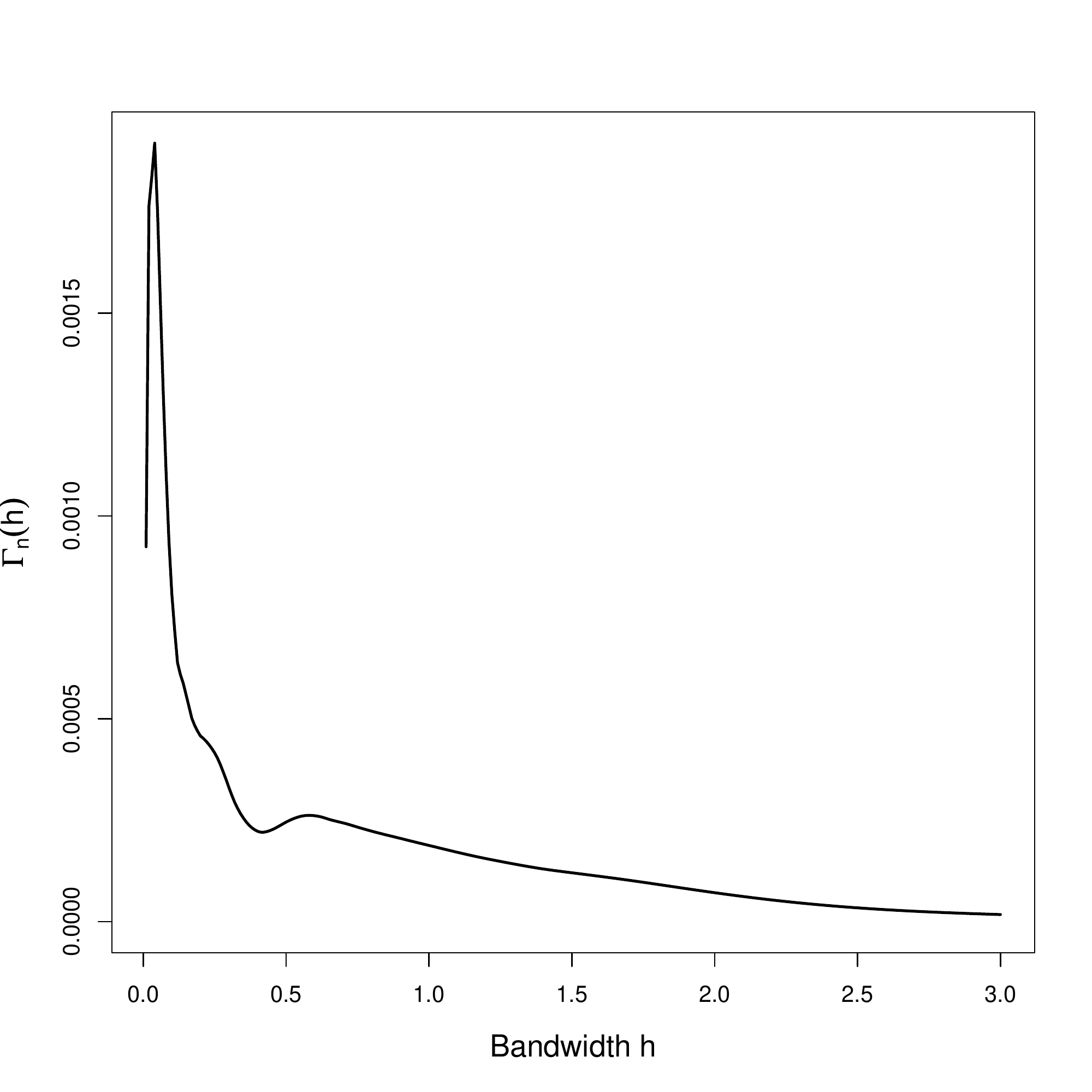}
\end{center}
\vspace{-5mm}
\caption{$\Xi_n(h)$ as a function of $h$ for $\lambda = 0.10$ (top left) $0.20$ (top right) and $0.30$ (bottom left). $\Gamma_n(h)$ as a function of $h$ (bottom right) for the data depicted in Figure \ref{fig::moons}.}
\label{fig::lambda2D}
\end{figure}

To examine instability, we use a product kernel density estimate with an Epanechikov kernel and the same bandwidth $h$ for both dimensions.  Figure \ref{fig::lambda2D} shows the sample instability $\Xi_n(h)$ as a function of $h$ for $\lambda = 0.10, 0.20, 0.30$ as well as the total variation instability $\Gamma_n(h)$ as a function of $h$.  As expected, the higher the $\lambda$, the more quickly the sample instability drops.  We also see the possible presence of multi-modality for all three values of $\lambda$ in $\Xi_n(h)$. On the other hand, the total variation instability drops smoothly as $h$ increases.

\begin{figure}[!ht]
\begin{center}
\includegraphics[width=2.5in,height=2.5in]{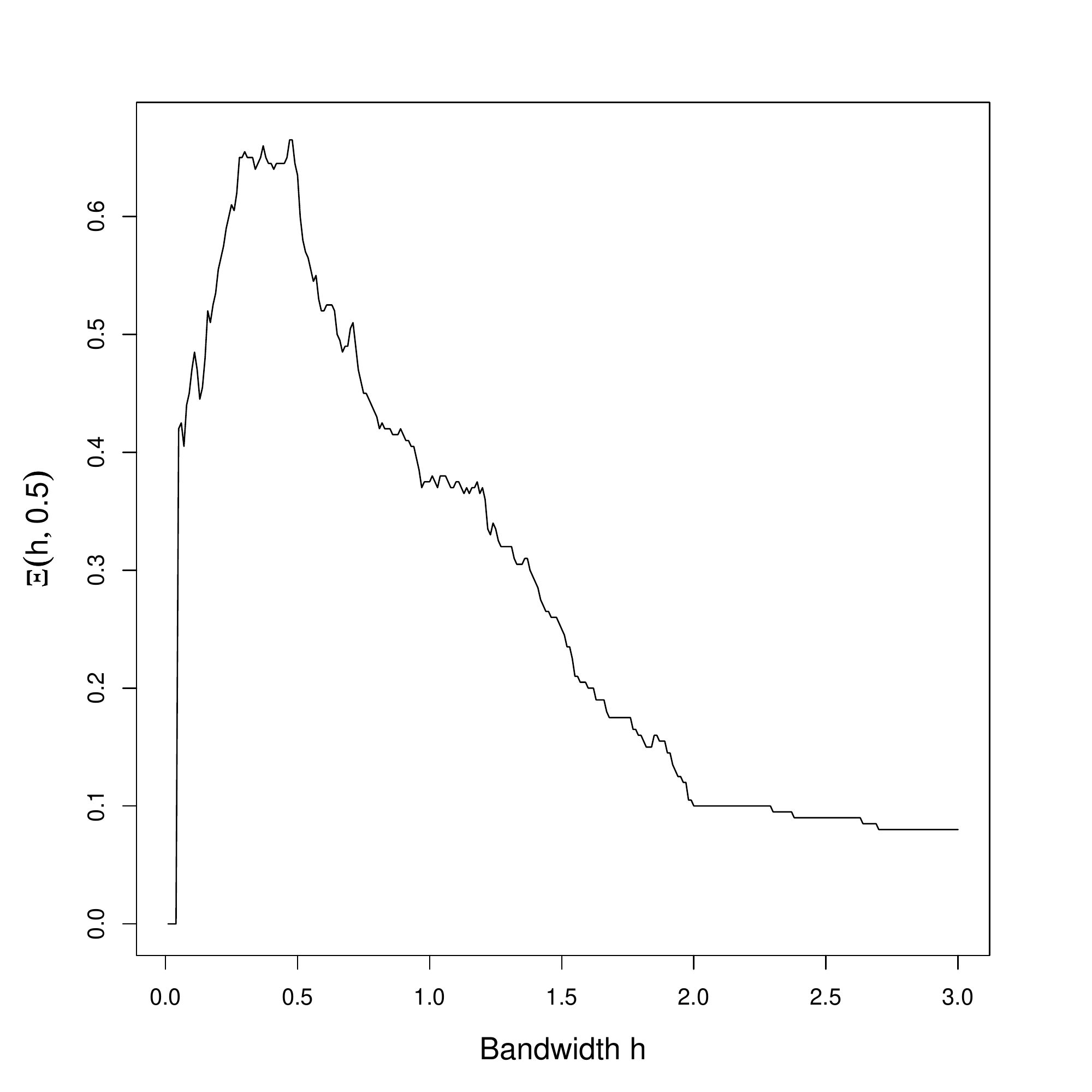}
\includegraphics[width=2.5in,height=2.5in]{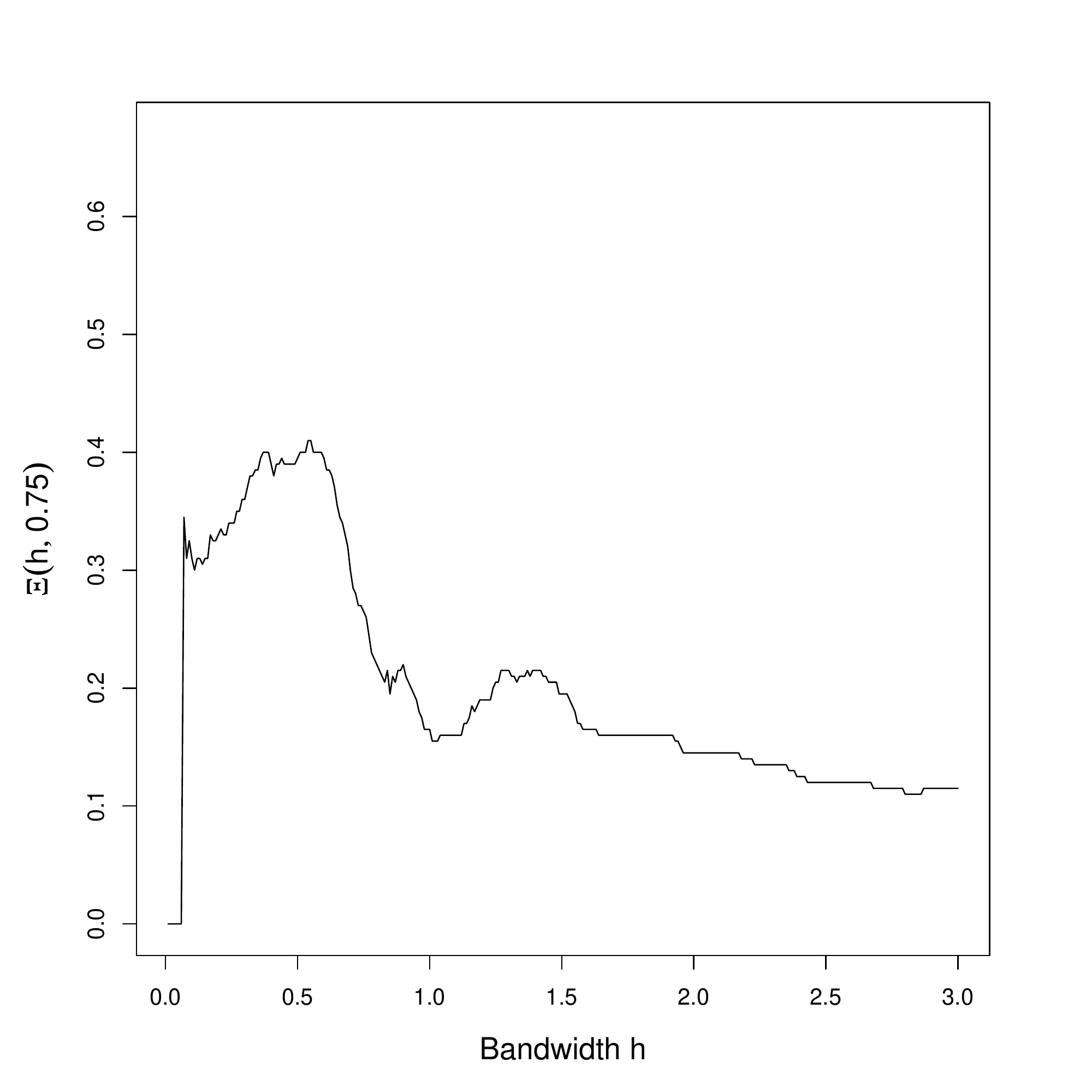}

\includegraphics[width=2.5in,height=2.5in]{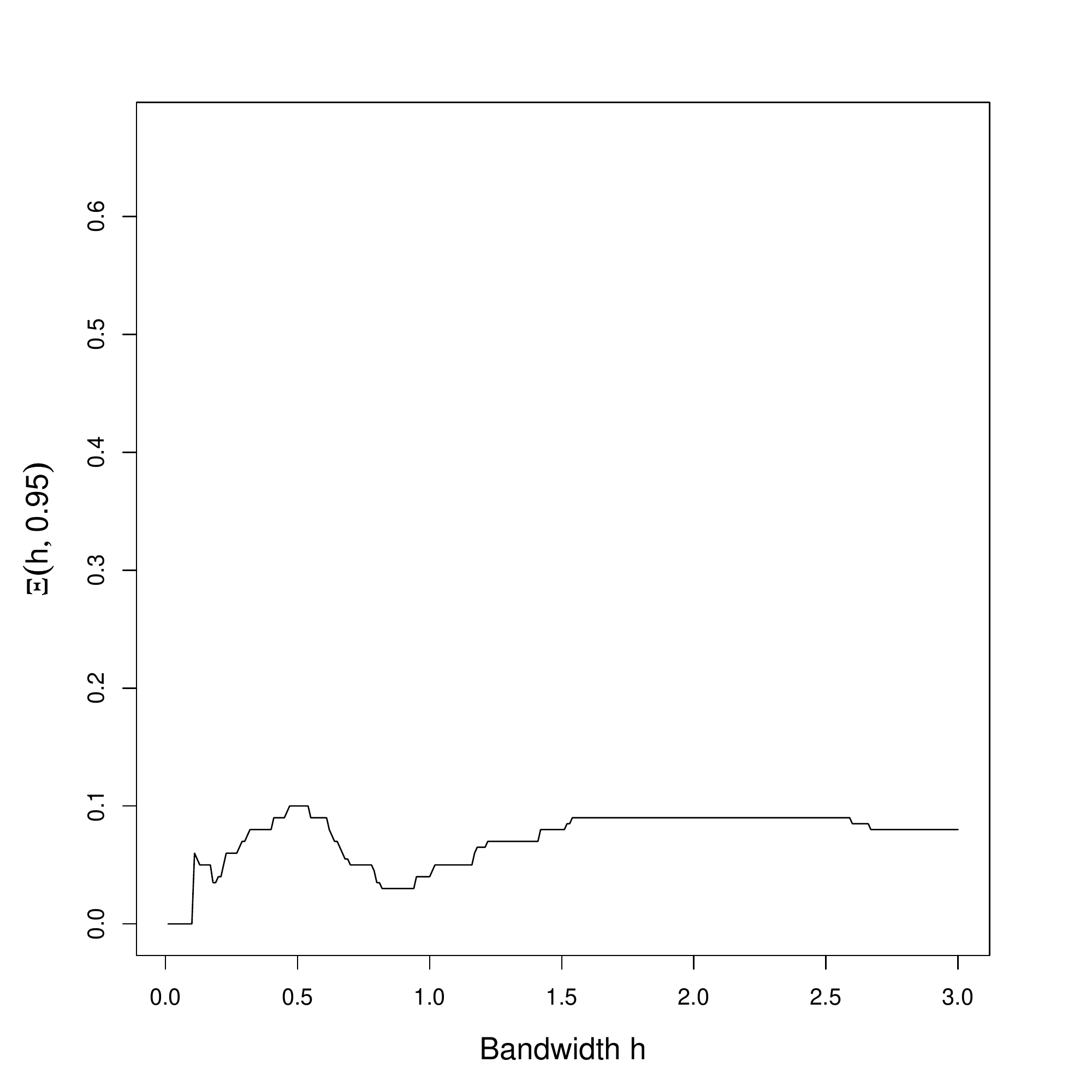}
\includegraphics[width=2.5in,height=2.5in]{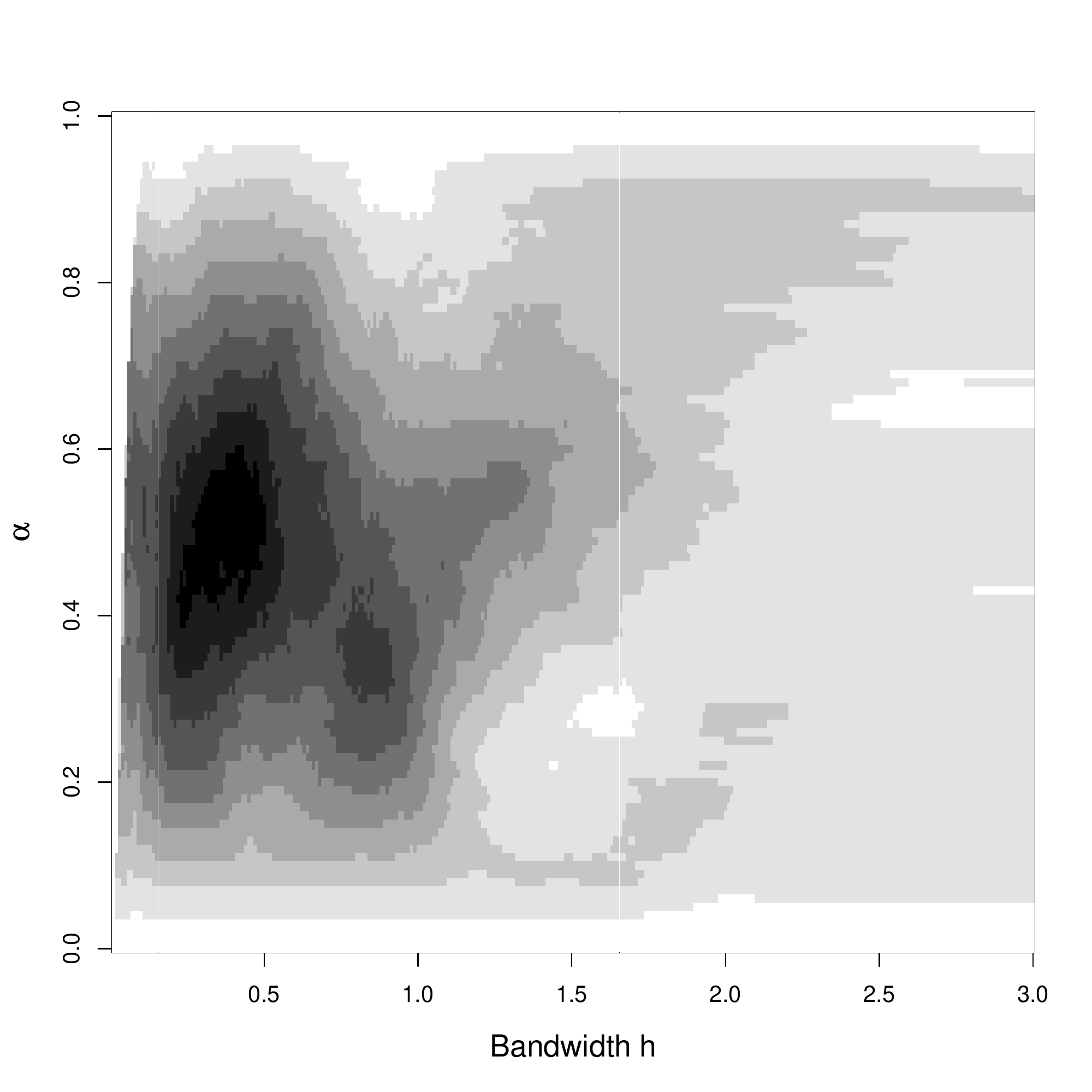}
\end{center}
\vspace{-5mm}
\caption{$\Xi_n(h,\alpha)$ as a function of $h$ for $\alpha = 0.50$ (top left) $0.75$ (top right) and $0.95$ (bottom left). Heat Map of $\Xi_n(h,\alpha)$ as function of $h, \alpha$ (bottom right).} 
\label{fig::alpha2D}
\end{figure}

Figure \ref{fig::alpha2D} contains the instability as a function of $h$ and probability content $\alpha$ for all values of $h$, $\alpha$ (Figure \ref{fig::alpha2D}d) and specifically for $\alpha = 0.50, 0.075, 0.95$.  Again, as expected, $\Xi_n(h, \alpha)$ drops as $h$ increases for smaller values of $\alpha$.  Note that for $\alpha = 0.95$, the instability remains relatively low regardless of the value of $h$.  When examining the heat map, we see that for small values of $h$, level sets corresponding to probability content around 0.4-0.6 are very unstable.  This behavior is not unexpected given that the moons are of equal sizes and difficult to separate due to the noise.  We would expect to have difficulty finding stable level sets ``in the middle''.

\section{Discussion}
\label{section::discussion}

We have  investigated the properties of the density level set and  density tree estimator based on kernel density estimates, and we have proposed and analyzed various  measures of instability for these quantities. We believe these measures of instability can provide useful guidelines for choosing the bandwidth parameter and also as explorative tools to gain insights into the properties and shape of the data-generating distribution.

Our analysis leaves some some open questions that we think deserve further attention. First, we have focused on kernel density estimators but the same ideas can be used with other density estimators or more, generally, with other clustering methods for which underlying tuning parameters have to be chosen in a data-driven fashion. See, for instance,  \cite{Meinshausen} for a related stability-based approach to clustering.

We have assumed the existence of the Lebesgue density $p$ but this 
assumption can be relaxed using methods in \cite{rinaldo} to allow for distributions supported on lower-dimensional, well-behaved subsets.
This extension is potentially important because it would allows us to include cases where the distribution has
positive mass on lower dimensional structures such as points and manifolds.

Finally, in computing the various measures of instability, we have considered just a single split of the data into non-overlapping sub-samples. In fact, one can randomly repeat the splitting process and combine over many splits, which is how we obtained the confidence bands of Figure \ref{fig::confbands}. Though the increase in the computational costs may be significant, repeated sub-sampling would yield a reliable estimate of the uncertainty of the chosen instability measures and would therefore be highly informative about the sample.  We believe that the properties of $\Xi_n$ can be established using the theory of U-statistics.



\section{Proofs}
\label{section::proofs}

\noindent \textbf{Proof of Theorem~\ref{thm:risk.lambda}:}
Let $\mathcal{A}_{h_n,\epsilon_n}$ denote the event that $\|\hat p_{h_n,X} - p_{h_n}\|_\infty \leq \epsilon_n$.
Then, for all $n\geq n_0$, by equation (\ref{eq:gine}), $\mathbb{P}_X(\mathcal{A}_{h_n,\epsilon_n}) \geq 1 - \frac{1}{n}$.  
Also observe that Assumption (A1) implies that, for any $h > 0$, the sup-norm density approximation error can be bounded as
\begin{eqnarray}
\label{eq:p_bias}
\|p_h - p\|_\infty & = & \sup_x \left|\int \frac1{h^d} K\left(\frac{\|x-y\|}{h}\right)p(y) dy - p(x)\right| \nonumber\\
& \leq &\sup_x \left|\int \frac1{h^d} K\left(\frac{\|x-y\|}{h}\right)A \|x-y\|dy\right|\nonumber\\
& = & A D h.
\end{eqnarray}
The second step in the previous display follows since $\int K(\|z\|)dz = 1$ and using Lipschitz assumption (A1) on the density, 
and the last step since $|\int \|z\|K(\|z\|)dz|<\infty$.
Putting the estimation and approximation error together, and using the triangle inequality, we obtain that, on the event $\mathcal{A}_{h_n,\epsilon_n}$,
\begin{equation}
\label{eq:supnorm}
\|\hat p_{h_n,X} - p\|_\infty \leq A D h_n + \epsilon_n, 
\end{equation}
for all $n\geq n_0$. 
Using  equation~(\ref{eq:supnorm}), we conclude that, on  $\mathcal{A}_{h_n,\epsilon_n}$ and for all $n \geq n_1(n_0,\lambda)$ so that $A D h_n + \epsilon_n < \lambda$,
\begin{eqnarray*}
L(\lambda) \Delta \hat L_{h_n,X}(\lambda) & = & \{u \colon p(u) >\lambda, \hat{p}_{h_n,X}(u) \leq \lambda\}
\cup \{u \colon p\leq\lambda, \hat{p}_{h_n,X}(u) > \lambda\}\\
& \subseteq & \{u \colon p(u) >\lambda, p(u) \leq \lambda+ A D h_n +  \epsilon_n\}
\cup \{u \colon p(u) \leq\lambda, p(u) > \lambda- A D h_n -  \epsilon_n\}\\
& = & \{ u \colon |p(u) - \lambda| < A D h_n +  \epsilon_n\}.
\end{eqnarray*}
Then, on  $\mathcal{A}_{h_n,\epsilon_n}$ and for all $n\geq n_1(n_0,\lambda)$ large enough
\[
{\cal L}(h_n,X,\lambda) = P(L(\lambda) \Delta \hat L_{h_n,X}(\lambda)) \leq r_{h_n,\epsilon_n,\lambda},
\]
so that,  $ \mathbb{P}_X \left( {\cal L}(h_n,X,\lambda) \leq r_n \right) \geq \mathbb{P}_X \left( \mathcal{A}_{h_n,\epsilon_n} \right) \geq 1 - \frac{1}{n}$, as claimed.

If (A2) is in force for the density level $\lambda$, then for all $n \geq n_2(n_0,\lambda,A,D,\epsilon_0)$ so that $A D h_n +  \epsilon_n \leq \epsilon_0$, we have $r_{h_n,\epsilon_n,\lambda} \leq \kappa_2 (A D h_n +  \epsilon_n)$, which proves the second claim.
\vspace{0.7cm}

\noindent \textbf{Proof of Lemma~\ref{lemma::lam_dev}:}
Using (A1) and the fact that $\int_{\mathbb{R}^d} K(\|z\|) dz =1$, Eq.(\ref{eq:p_bias})
states that for any $h>0$
$$
\|p_h - p\|_\infty \leq A D h.
$$
Then, for any $\alpha \in (0,1)$ and $h>0$,
$$
\{u \colon p(u)>\lambda_{h,\alpha} + ADh\} \subseteq \{u \colon p_h(u) > \lambda_{h,\alpha}\} \subseteq \{u \colon p(u) >\lambda_{h,\alpha} - ADh\}.
$$
And as a result, 
$$
P(\{u \colon p(u) >\lambda_{h,\alpha} + AD h\}) \leq P(\{u \colon p_h(u) >\lambda_{h,\alpha}\}) \leq P(\{u \colon p(u) >\lambda_{h,\alpha}- ADh\}).
$$
Since $P(\{u \colon p(u) >\lambda_\alpha\}) = \alpha = P(\{u \colon p_h(u) >\lambda_{h,\alpha}\})$, we have
$$
P(\{u \colon p(u) >\lambda_{h,\alpha} + AD h\}) \leq P(\{u \colon p(u) >\lambda_\alpha\}) \leq P(\{u \colon p(u) >\lambda_{h,\alpha}-AD h\}).
$$
Consequently,
$$
\lambda_{h,\alpha} + AD h \geq \lambda_{\alpha} \geq \lambda_{h,\alpha} - AD h.
$$
It follows that for any $\alpha \in (0,1)$ and $h>0$
$$
| \lambda_{h,\alpha}- \lambda_\alpha | \leq AD h.
$$
\vspace{0.3cm}

\noindent\textbf{Proof of Lemma \ref{lem:alpha}:}
Let  $\mathcal{C}_h = \big\{ \{ u \colon p_h(u) > \lambda\}, \lambda > 0 \big\}$ denote the class of level sets of $p_h$ and define the events 
\[
\mathcal{P}_{h,\epsilon} = \left\{ \sup_{C\in {\cal C}_{h}} |\hat{P}_X(C) - P(C)| \leq \epsilon \right\} \quad  \text{and} \quad \mathcal{A}_{h,\epsilon} = \left\{ ||\hat{p}_{h,X} - p_h||_\infty \leq \epsilon \right\}.
\]
Then, since the $n$-th shatter coefficients of ${\cal C}_{h}$ is $n$, 
\begin{equation}\label{eq:P.O}
\mathbb{P}_X(\mathcal{P}^c_{h,\epsilon}) \leq 8 n e^{ - n \epsilon^2/32 } \quad \text{and} \quad \mathbb{P}_X(\mathcal{A}^c_{h,\epsilon}) \leq  K_1 e^{ - K_2 n \epsilon^2 h^d },
\end{equation}
where the first inequality follows from the VC inequality and the second inequality is just \eqref{eq:gine_prob}.
Then, on $\mathcal{A}_{h,\epsilon}$, we obtain
$$
\{u \colon p_h(u) > \lambda + \epsilon\} \subseteq
\{u \colon \hat{p}_{h,X}(u) > \lambda \} \subseteq
\{u \colon p_h(u) > \lambda - \epsilon\}, \quad \forall \lambda > 0.
$$
Thus, on $\mathcal{A}_{h,\epsilon}$,
\[
\hat{P}_X(\{u \colon p_h(u) > \lambda + \epsilon\}) \leq
\hat{P}_X(\{ u \colon \hat{p}_{h,X}(u)  > \lambda \}) \leq
\hat{P}_X(\{ u \colon p_h(u) > \lambda - \epsilon\}),
\]
uniformly over all $\lambda > 0$ and any $h>0$. In particular, the previous inequality hold also for $\hat{\lambda}_{\alpha,h,X}$ 
(which is positive with probability one) for any $\alpha \in (0,1)$ and $h>0$. 

Recalling that, by definition,
\[
|\hat{P}_X(\{ u \colon \hat{p}_{h,X}(u)  > \hat{\lambda}_{h,\alpha,X} \}) -  \alpha| \leq 1/n,
\]
we obtain, on the events $\mathcal{P}_{h,\epsilon}$ and $\mathcal{A}_{h,\epsilon}$,
\begin{equation}\label{eq:two}
P(\{u \colon p_h(u) > \hat{\lambda}_{h,\alpha,X} + \epsilon\}) - \frac{1}{n} - \epsilon \leq \alpha \leq 
P\{u \colon p_h(u) > \hat{\lambda}_{h,\alpha,X} - \epsilon\}) + \frac{1}{n} + \epsilon.
\end{equation}
Since $\alpha = P(\{ u \colon p_h(u) > \lambda_{h,\alpha}\})$, the first inequality in \eqref{eq:two} can be written as
\[
\alpha + \frac{1}{n} + \epsilon = P(\{ u \colon p_h(u) > \lambda_{h,\alpha + \frac{1}{n} + \epsilon} \})  \geq 
P(\{u \colon p_h(u) > \hat{\lambda}_{h,\alpha,X} + \epsilon\})
\]
and the second one as
\[
\alpha - \frac{1}{n} - \epsilon = P(\{ u \colon p_h(u) > \lambda_{h,\alpha - \frac{1}{n} - \epsilon} \})  \leq P\{u \colon p_h(u) > \hat{\lambda}_{h,\alpha,X} - \epsilon\}),
\]
both holding on the events  $\mathcal{P}_{h,\epsilon}$ and $\mathcal{A}_{h,\epsilon}$.
Combining the last two expressions, we obtain, on the same events, for any $\alpha \in (0,1)$ and $h>0$
\begin{equation}\label{eq:almost}
\lambda_{h,\alpha + \frac{1}{n} + \epsilon } - \epsilon \leq \hat{\lambda}_{h,\alpha,X} \leq \lambda_{h,\alpha -  \frac{1}{n} - \epsilon } + \epsilon.
\end{equation}
We will now show that for level sets of $p_h$ indexed by $\alpha$ that satisfy (B3), 
for any $\eta \in (-\eta_0,\eta_0)$ and $0<h\leq H$,  we have 
\begin{equation}\label{eq:lambda_eta}
| \lambda_{h,\alpha + \eta} - \lambda_{h,\alpha} | \leq A\kappa_3 |\eta|.
\end{equation}
In fact, \eqref{eq:almost} and \eqref{eq:lambda_eta} will imply, on the events $\mathcal{P}_{h,\epsilon}$ and $\mathcal{A}_{h,\epsilon}$,  for level sets of $p_h$ indexed by $\alpha$ that satisfy (B3) since $\epsilon + 1/n < \eta_0$ and $0<h\leq H$,  we have
\[
\lambda_{h,\alpha} - A\kappa_3\left(\epsilon + \frac{1}{n}\right) - \epsilon \leq \hat\lambda_{h,\alpha,X} \leq \lambda_{h,\alpha} + A\kappa_3\left(\epsilon + \frac{1}{n}\right) + \epsilon,
\]
from which, using \eqref{eq:P.O}, 
the claim will follow.

In order to show \eqref{eq:lambda_eta}, for a set $A \subset \mathbb{R}^d$, let  $\partial A$ denote its boundary. Then, notice that, because $p_h$ is Lipschitz and hence continuous, for every $x \in \partial M_{h}(\alpha)$, $p_h(x) = \lambda_{h,\alpha}$ and, for every $y \in \partial M_{h}(\alpha + \eta)$, $p_h(y) = \lambda_{h,\alpha + \eta}$. Furthermore, for any  point $x \in \partial M_{h}(\alpha)$, there exists a point $y = y(x) = \inf_{z \in \partial M_{h}(\alpha + \eta)} \| x - z\|$.
Thus, for $|\eta| < \eta_0$,
\[
\| x - y\| \leq d_\infty(M_{h}(\alpha),M_{h}(\alpha+\eta)) \leq \kappa_3 |\eta|,\]
where the last inequality follows for level sets of $p_h$ indexed by $\alpha$ that satisfy (B3) and $0<h\leq H$. Therefore,
\[
| \lambda_{h,\alpha + \eta} - \lambda_{h,\alpha} | = |p_h(y) - p_h(x)| \leq A \| x - y\| \leq A \kappa_3 |\eta|,
\]
where in the first inequality we used the fact that, by (A1), $p_h$ is Lipschitz with constant $A$. Indeed, for any $x \neq y$, using the Lipschitz assumption (A1) on $p$,
\[
|p_h(x) - p_h(y)| \leq \int_{\mathbb{R}^d} \left| p(x + zh) - p(y+zh) \right|K(z) dz\leq A \| x - y \| \int_{\mathbb{R}^d} K(z) dz = A \| x - y \|.
\]
\vspace{0.7cm}

\noindent\textbf{Proof of Theorem~\ref{thm:risk.alpha}:}
Let $\mathcal{A}_{h_n,\epsilon_n}$ be event defined in the proof of Theorem \ref{thm:risk.lambda}, and recall that
for all $n\geq n_0$, by equation~(\ref{eq:gine}), $\mathbb{P}_X(\mathcal{A}^c_{h_n,\epsilon_n}) \leq 1/n$ and that,  equation \eqref{eq:supnorm} states that
\begin{equation}\label{eq:C1}
\|\hat p_{h,X} - p\|_\infty \leq C_{1,n}
\end{equation}
on that event, for all $n\geq n_0$. 
Also, let $\mathcal{P}_{h_n,\epsilon_n}$ be the event defined in Lemma \ref{lem:alpha} such that 
$\mathbb{P}_X(\mathcal{P}^c_{h_n,\epsilon_n}) \leq  8n e^{-n \epsilon_n^2/32}$.
Then from Lemma \ref{lem:alpha} proof, we have that on the event $\mathcal{A}_{h_n,\epsilon_n} 
\cap \mathcal{P}_{h_n,\epsilon_n}$, for $h_n = \omega((\log n/n)^{1/d})$ and $h_n \leq H $, 
\begin{equation}\label{eq:C2}
|\hat\lambda_{h_n,\alpha,X}- \lambda_{\alpha}| \leq C_{2,n}
\end{equation}
for all $n\geq n_3(n_0,\eta_0,K_3)$. Also, since $n$ is large enough, we have
\[
8n e^{-n \epsilon_n^2/32} \leq \frac{1}{n}.
\]
Therefore, for all such large $n$, both \eqref{eq:C1} and \eqref{eq:C2} hold with probability at least 
$ \mathbb{P}_X \left(\mathcal{A}_{h_n,\epsilon_n} \cap \mathcal{P}_{h_n,\epsilon_n} \right) \geq 1 - \frac{2}{n}$.
Thus, on $\mathcal{A}_{h_n,\epsilon_n} \cap \mathcal{P}_{h_n,\epsilon_n}$, for $h_n = \omega((\log n/n)^{1/d})$
and $h_n \leq H$, we have for all $n\geq n_3(n_0,\eta_0,K_3)$
\begin{eqnarray*}
M(\alpha) \Delta \hat M_{h,X}(\alpha) & = & \{u \colon p(u) >\lambda_\alpha, \hat{p}_{h,X}(u) \leq \hat{\lambda}_{h,\alpha,X}\}
\cup \{u \colon p(u) \leq \lambda_\alpha, \hat{p}_{h,X}(u) > \hat\lambda_{h,\alpha,X}\}\\
& \subseteq & \{u \colon p(u) >\lambda_\alpha, p(u) \leq  \hat\lambda_{h,\alpha,X}+ C_{1,n}\}
\cup \{u \colon p(u) \leq\lambda_\alpha, p(u) > \hat\lambda_{h,\alpha,X}-C_{1,n}\}\\
& \subseteq & \{u \colon p(u)>\lambda_\alpha, p(u) \leq \lambda_{\alpha}+C_{1,n} + C_{2,n}\}
\cup \{u \colon p(u) \leq\lambda_\alpha, p(u) > \lambda_{\alpha}-C_{1,n} - C_{2,n}\}\\
& = & \{ u \colon |p(u) - \lambda_\alpha| \leq C_{1,n} + C_{2,n}\}.
\end{eqnarray*}
Therefore, for for $h_n = \omega((\log n/n)^{1/d})$
and $h_n \leq H$, we have for all $n\geq n_3(n_0,\eta_0,K_3)$,
\[
\mathbb{P}_X \left( \mathcal{L}(h_n,X,\alpha) \leq r_{h_n,\epsilon_n,\alpha} \right) \geq  \mathbb{P}_X \left(\mathcal{A}_{h_n,\epsilon_n} \cap \mathcal{P}_{h_n,\epsilon_n} \right) \geq 1 - \frac{2}{n}.
\]

\vspace{1cm}

\noindent
\noindent{\bf Proof of Theorem~\ref{thm:xi}:}
\begin{enumerate}
\item Since $X$, $Y$ and $Z$ are independent samples from the same
  distribution, $\hat{p}_{h,X}(u)$ and $\hat{p}_{h,Y}(u)$ are
  independent and identically distributed, for any $u \in
  \mathbb{R}^d$ and $h > 0$. Also, notice that for every measurable
  set $A$, $\mathbb{E}_Z(\hat{P}_Z(A)) = P(A)$. Thus,
\begin{eqnarray}
\xi(h) & = & \mathbb{E}_{X,Y,Z}[\hat{P}_Z(\{u \colon \hat p_{h,X}(u) > \lambda\}\Delta\{ u \colon \hat p_{h,Y}(u) > \lambda\})] \nonumber\\
& = & \mathbb{E}_{X,Y}[P(\{ u \colon \hat p_{h,X}(u) > \lambda, \hat p_{h,Y}(u) \leq \lambda\}) + 
P(\{u \colon \hat p_{h,X}(u) \leq \lambda, \hat p_{h,Y}(u) > \lambda\})] \nonumber \\
& = & 2 \mathbb{E}_{X,Y} \left[ P(\{ u \colon \hat p_{h,X}(u) > \lambda, \hat p_{h,Y}(u) \leq \lambda\}) \right] \nonumber \\
& = & 2 \int_{\mathbb{R}^d}  \mathbb{P}_{X,Y} \left( \hat p_{h,X}(u) > \lambda, \hat p_{h,Y}(u) \leq \lambda \right) dP(u), \label{eq:xi.deriv2}
\end{eqnarray}
where the last identity follows from Fubini theorem.
The integrand in the last equation can be written as
\[
\begin{array}{rcl}
\mathbb{P}_{X,Y} \left( \hat p_{h,X}(u) > \lambda, \hat p_{h,Y}(u) \leq \lambda \right) & =  & \mathbb{P}_X \left( \hat p_{h,X}(u) > \lambda \right) \mathbb{P}_Y \left( \hat p_{h,Y}(u) \leq \lambda  \right)\\
& = & \mathbb{P}_X\left( \hat p_{h,X}(u) > \lambda \right) \mathbb{P}_X \left( \hat p_{h,X}(u) \leq \lambda  \right)\\
& = & \pi_h(u) ( 1- \pi_h(u)),
\end{array}
\]
from which \eqref{eq:piu}  follows.
 
\item Let $\mathcal{A}_{h,\epsilon}$ denote the event
\begin{equation}\label{eq:pinfity.X.Y}
\| p_h - \hat{p}_{h,X} \|_\infty \vee \| p_h - \hat{p}_{h,Y} \|_\infty \leq \epsilon.
\end{equation}
By \eqref{eq:gine_prob}, $\mathbb{P}_{X,Y}(\mathcal{A}^c_{h,\epsilon}) \leq 2 K_1 e^{- K_2 n h^d \epsilon^2}$.
Letting $1_{\mathcal{A}_{h,\epsilon}}$ denote the indicator function of the event $\mathcal{A}_{h,\epsilon}$,
\[
\xi(h) \leq \mathbb{E}_{X,Y,Z}[\hat{P}_Z(\{u \colon \hat p_{h,X}(u) > \lambda\}\Delta\{ u \colon \hat p_{h,Y}(u) > \lambda\}) 1_{\mathcal{A}_{h,\epsilon}}(X,Y)] + \mathbb{P}_{X,Y}(\mathcal{A}^c_{h,\epsilon}),
\]
and, using the same reasoning that led to \eqref{eq:xi.deriv2},
\[
\xi(h) \leq  2 \int_{\mathbb{R}^d}  \mathbb{P}_{X,Y} \left( \{ \hat p_{h,X}(u) > \lambda, \hat p_{h,Y}(u) \leq \lambda\} \cap {\cal A}_{h,\epsilon} \right) dP(u)  + \mathbb{P}_{X,Y}({\cal A}_{h,\epsilon}^c)\\
\]
Notice that, on $\mathcal{A}_{h,\epsilon}$,
\[
\{ u \colon \hat{p}_{h,X}(u) > \lambda, \hat{p}_{h,Y}(u) \leq \lambda \} \subseteq \{ u \colon  \lambda - \epsilon \leq p_h(u) \leq  \lambda + \epsilon\} = U_{h,\epsilon},
\]
and therefore, $\mathrm{sign}(\hat{p}_{h,X}(u) - \lambda) = \mathrm{sign}(p_h(u) - \lambda) $ for all $u \not \in U_{h,\epsilon}$.
Thus, the previous expression for $\xi(h)$ is upper bounded by
\[
2 \int_{U_{h,\epsilon}} \mathbb{P}_{X,Y} \left( \{ \hat p_{h,X}(u) > \lambda, \hat p_{h,Y}(u) \leq \lambda\} \cap {\cal A}_{h,\epsilon}  \right) dP(u) + 2K_1 e^{- K_2 n h^d \epsilon^2}
\]
which, in turn, using independence, is no larger than
\[
2 \int_{U_{h,\epsilon}} \pi_h(u)(1 - \pi_h(u)) dP(u) + 2K_1 e^{- K_2 n h^d \epsilon^2} \leq P(U_{h,\epsilon})  \overline{A}_{h,\epsilon} + 2K_1 e^{- K_2 n h^d \epsilon^2}.
\]
As for the lower bound, from \eqref{eq:xi.deriv2} we obtain, trivially,
$$
\xi(h)  \geq 2 \int_{U_{h,\epsilon}} \pi_h(u)(1 - \pi_h(u)) dP(u) \geq
P(U_{h,\epsilon}) \overline{A}_{h,\epsilon}.
$$
\end{enumerate}

\vspace{1cm}

\noindent
{\bf Proof of Lemma \ref{lemma:xi_smallh}:}
For simplicity, we will provide the proof for the case of a spherical kernel, i.e. $K(x) = 1_{\| x\| \leq 1}$, $x \in \mathbb{R}^d$. The extension to other compactly supported kernels is analogous.

By the minimal spacings theorem 
\citep[see][]{deheuvels}, for all $N$ large enough, there exists a constant $C>0$ such that, $P$-almost surely, the quantities
\[
\min_{i\neq j} ||X_i - X_j||, \quad \min_{i\neq j} ||Y_i - Y_j||  \quad {\rm and} \quad \min_{i, j} ||X_i - Y_j|| 
\]
are all larger than $ C \left(\frac{\log n}{n}\right)^{1/d}$.
Hence, by the compactness of the support of $K$, if $h <  C (\log n/n)^{1/d}/2$,
the sets 
$B(X_1,h),\ldots, B(X_n,h),B(Y_1,h),\ldots, B(Y_n,h)$
are disjoint. Therefore, $\hat p_{h,X}(u) = 1/(n h^d)$ if and only if $u \in B(X_i,h)$ for one $i$ and, similarly,  $\hat p_{h,Y}(u) = 1/(n h^d)$ if and only if $u \in B(Y_j,h)$ for one $j$.
Furthermore, 
\[
\hat L_{h,X}\Delta \hat L_{h,Y} = \left(\bigcup_i B(X_i,h) \right)\bigcup \left(\bigcup_j B(Y_j,h) \right).
\]
As a result, $\Xi_n(h)$ is the fraction of $Z_i$'s contained in
$\left(\cup_i B(X_i,h) \right)\bigcup \left(\cup_i B(Y_i,h) \right)$.
Thus, 
\[
\Xi_n(h) = \hat P_Z(\hat L_{h,X}\Delta \hat L_{h,Y}|X,Y) \stackrel{d}{=} B/n,
\]
where $\stackrel{d}{=}$ denotes equality in distribution and $B\sim {\rm Binomial}(n, p_0)$, with $0 \leq p_0 \leq 2n \ p_{\max} v_d h^d$ and $p_{\max} = \| p \|_\infty$. 
Therefore, 
$\mathbb{E}_{Z}[\Xi_n(h)|X,Y] \leq 2 p_{\max} v_d n h^d$ and hence it follows that
\[
\xi(h) = \mathbb{E}_{X,Y,Z}[\Xi_n(h)] \leq 2 p_{\max} v_d n h^d = O(h^d),
\]
as $h \rightarrow 0$.

\vspace{1cm}

\noindent
{\bf Proof of Lemma \ref{prop:lambda.berry-esseen}.}
If $K$ is the spherical kernel, note that
$\hat{p}_{h,X}(u)=n^{-1}\sum_{i=1}^n B_i(u)$,
where
\[
B_i = h^{-d}K\left(\frac{||u-X_i||}{h}\right) =
\frac{I_{B(u,h)}(X_i)}{(h^d v_d)},
\]
 with $I_{B(u,h)}(\cdot)$ denoting the indicator function of the ball $B(u,h)$.
Let $\sigma^2(u,h) = {\rm Var}(B_i(u))$ and
$\mu_3(u,h) = \mathbb{E} |B_i(u) - \mu(u,h)|^3$
where $\mu (u,h)= \mathbb{E}(B_i(u)) = p_h(u)$.
Finally, let $p_{u,h} = P(B(u,h))$. Then, 
\begin{equation}\label{eq:sigma2}
\sigma^2(u,h) = \frac{p_{u,h}(1 - p_{u,h})}{(h^{d}v_d)^2}
\end{equation}
and
\[
\mu_3(u,h) = \frac{p_{u,h}(1 - p_{u,h}) \left[ (1 - p_{u,h})^2 + p_{u,h}^2 \right]}{(h^{d}v_d)^3} \leq 
\frac{p_{u,h}(1 - p_{u,h}) }{(h^{d}v_d)^3}, 
\]
where the last inequality holds since $(1 - p_{u,h})^2 + p_{u,h}^2 \leq 1$, for all $u$ and $h$.
As a result,
\begin{equation}\label{eq:mu3sigma3}
\frac{\mu_3(u,h)}{\sigma^{3}(u,h)} \leq \left( p_{u,h}(1 - p_{u,h}) \right)^{-1/2}.
\end{equation}
By assumption, $h < h(\delta,\epsilon)$ and $\epsilon \leq \lambda/2$. 
In order to avoid trivialities, we further assume that $P(U_{h,\epsilon}) > 0$. Then, uniformly over all $u$ in $U_{h,\epsilon}$,
\[
(\lambda - \epsilon) v_d h^d \leq p_{u,h} \leq (\lambda + \epsilon) v_d h^d
\]
and
\[
(1 - p_{u,h}) \geq \delta.
\]
Thus, 
\[
\frac{\mu_3(u,h)}{\sigma^{3}(u,h)} \leq \sqrt{\frac{1}{\delta v_d h^d(\lambda - \epsilon)}} \leq \sqrt{\frac{2}{ h^d \delta v_d  \lambda }}, 
\]
with the last inequality holding because of our assumption  $\epsilon \leq \lambda/2$.
From \eqref{eq:sigma2}, we then obtain
\[
\frac{\delta (\lambda - \epsilon)}{v_d h^d} \leq \sigma^2(u,h) \leq \frac{(\lambda + \epsilon)}{v_d h^d}.
\]
Thus,
\[
\frac{a_1}{h^d} \leq \sigma^2(u,h) \leq \frac{a_2}{h^d} ,
\]
where 
\begin{equation}\label{eq:a1.a2}
a_1 = \frac{\delta \lambda}{2 v_d} \quad \text{and} \quad a_2 = \frac{3 \lambda}{2 v_d},
\end{equation}
 uniformly over $u \in U_{h,\epsilon}$.

Writing $\sigma^2(u,h) = a(u,h)/h^d$ and using the Berry-Ess\'een bound (\cite{olive} p 78), we obtain 
$$
\sup_t \left|P\left(\frac{\sqrt{n h^d}(\hat{p}_{h,X}(u) - p_h(u))}{a(u,h)} \leq t \right) - \Phi(t)\right| \leq
\frac{33}{4} \frac{\mu_3(u,h)}{\sigma^3(u,h)\sqrt{n}} =
\sqrt{\frac{C(\delta,\lambda)}{nh^d}},
$$
where $\Phi$ is the cumulative distribution function of the standard 
Normal distribution.

Now,
$$
\pi_h(u) = \mathbb{P}_X(\hat{p}_{h,X}(u) > \lambda)=
\mathbb{P}_X \left(  \frac{\sqrt{nh^d}(\hat{p}_{h,X}(u) -p_{h}(u))}{a(u,h)} > 
\frac{\sqrt{nh^d}(\lambda - p_h(u))}{a(u,h)}\right).
$$
Hence,
$$
1 - \Phi\left( \frac{ \sqrt{nh^d}(\lambda-p_h(u))}{a(u,h)}\right) - \frac{C(\delta,\lambda)}{\sqrt{nh^d}} \leq 
\pi_h(u) \leq
1 - \Phi\left( \frac{ \sqrt{nh^d}(\lambda-p_h(u))}{a(u,h)}\right) + \frac{C(\delta,\lambda)}{\sqrt{nh^d}}.
$$
Using the fact that $u\in {U}_{h,\epsilon}$, and taking advantage of the uniform bounds 
$a_1 \leq a(u,h) \leq a_2$, the previous inequalities imply
\[
1 - \Phi\left( \frac{ \sqrt{nh^d}\epsilon }{a_1}\right) - \frac{C(\delta,\lambda)}{\sqrt{nh^d }}  \leq  
\pi_h(u) \leq
1 - \Phi\left( -\frac{ \sqrt{nh^d}\epsilon }{a_2}\right) + \frac{C(\delta,\lambda)}{\sqrt{n h^d}}.
\]
Noting that 
\[
1 - \Phi\left( \frac{ \sqrt{nh^d}\epsilon }{a_1}\right) = \Phi \left( - \frac{ \sqrt{nh^d}\epsilon }{a_1}\right) \geq \Phi \left( - \frac{ \sqrt{nh^d}\epsilon }{a_2}\right)
\]
and
\[
1 - \Phi\left( -\frac{ \sqrt{nh^d}\epsilon }{a_2}\right) = \Phi\left( \frac{ \sqrt{nh^d}\epsilon }{a_2}\right) 
\leq \Phi\left( \frac{ \sqrt{nh^d}\epsilon }{a_1}\right),
\]
we obtain the bounds 
\begin{equation}\label{eq:b1}
\Phi \left( - \frac{ \sqrt{nh^d}\epsilon }{a_2}\right)  - \frac{C(\delta,\lambda)}{\sqrt{nh^d }}   \leq \pi_h(u) \leq  1 - \Phi\left( -\frac{ \sqrt{nh^d}\epsilon }{a_2}\right) + \frac{C}{\sqrt{n h^d}}
\end{equation}
and
\begin{equation}\label{eq:b2}
1 - \Phi\left( \frac{ \sqrt{nh^d}\epsilon }{a_1}\right) - \frac{C(\delta,\lambda)}{\sqrt{nh^d }}   \leq \pi_h(u) \leq  \Phi\left( \frac{ \sqrt{nh^d}\epsilon }{a_1}\right) + \frac{C}{\sqrt{n h^d}},
\end{equation}
respectively.
Thus, uniformly over all $\epsilon \leq \lambda/2$ and all $h < h(\delta,\epsilon)$, equation \eqref{eq:b1} and \eqref{eq:b2} yield 
\begin{eqnarray*}
\overline{A}_{h,\epsilon} = 2 \sup_{u \in U_{h,\epsilon}}\pi_h(u)(1-\pi_h(u))  &\leq &
2 \left(1 - \Phi\left( - \frac{ \sqrt{nh^d}\epsilon }{a_2}\right) + \frac{C(\delta,\lambda)}{\sqrt{nh^d }}\right)^2 ,
\end{eqnarray*}
and 
\begin{eqnarray*}
\underline{A}_{h,\epsilon} = 2 \inf_{u \in U_{h,\epsilon}}\pi_h(u)(1-\pi_h(u)) &\geq &
2 \left(1 - \Phi\left( \frac{ \sqrt{nh^d}\epsilon }{a_1}\right) - \frac{C(\delta,\lambda)}{\sqrt{nh^d }}\right)^2 ,
\end{eqnarray*}
respectively, where $a_1$ and $a_2$ are given in \eqref{eq:a1.a2}.


\vspace{1cm}

\noindent
{\bf Proof of Lemma \ref{lemma::boundvar}.}
Letting $1_i = 1_{\{ Z_i \in \hat{L}_{X,h} \Delta \hat{L}_{Y,h} \}}$, we have
\[
\Xi_n(h) = \frac{1}{n} \sum_{i=1}^n 1_i.
\]
where,  conditionally on $X$ and $Y$, the $1_i$'s are independent and identically distributed Bernoulli random variables with
$\mathbb{E}_{Z}[1_i|X,Y] = P(\hat{L}_{h,X} \Delta \hat{L}_{h,Y})$.
Thus
\[
\begin{array}{rcl}
\mathbb{V}\left[ \Xi_n(h)\right] & = & \mathbb{E}_{X,Y,Z}\left[ \Xi_n^2(h) \right] - \xi^2(h)\\
 & = & \frac{1}{n^2}\mathbb{E}_{XY} [\mathbb{E}_Z\left[ (\sum_{i=1}^n 1_i + \sum_{j \neq k} 1_j 1_k) | X,Y ]\right] - \xi^2(h)\\
& = & \frac{\xi(h)}{n} + \frac{n-1}{2n} \mathbb{E}_{X,Y} \left[P^2(\hat{L}_{h,X} \Delta \hat{L}_{h,Y})\right]  - \xi^2(h)\\
& \leq & \frac{\xi(h)}{n} + \frac{n-1}{2n} \mathbb{E}_{X,Y} \left[P(\hat{L}_{h,X} \Delta \hat{L}_{h,Y})\right]  - \xi^2(h)\\
& = &  \frac{\xi(h)}{n} + \frac{n-1}{2n} \xi(h)  - \xi^2(h)\\
& = & \xi(h) \left( \frac{n+1}{2n} - \xi(h) \right).\\
\end{array}
\]

\vspace{1cm}

\noindent
{\bf Proof of Lemma \ref{lem:conc}.}  

Let $\xi(h,X,Y) =
\mathbb{E}_Z[\Xi_n(h) | X,Y]$ and let $A_{h,\epsilon}$ be the event
given in \eqref{eq:pinfity.X.Y}, where $\epsilon,h > 0$, so that
$\mathbb{P}_{X,Y}(\mathcal{A}^c_{h,\epsilon}) \leq 2 K_1 \exp \left\{
  - n K_2 h^d \epsilon^2 \right\}$ by \eqref{eq:gine_prob}.  Then, we
can write
\[
\mathbb{P}_{X,Y,Z} \left( \left| \Xi_n(h) - \xi(h) \right| > t \right) = \mathbb{P}_{X,Y,Z} \left( \left| \Xi_n(h) - \xi(h,X,Y) + \xi(h,X,Y)  - \xi(h) \right| > t \right),
\]
which is therefore upper bounded by
\[
\mathbb{P}_{X,Y,Z} \left( \left| \Xi_n(h) - \xi(h,X,Y) + \xi(h,X,Y)  - \xi(h) \right| > t ; \mathcal{A}_{h,\epsilon}\right) + 2 K_1 \exp \left\{  - n K_2 h^d \epsilon^2 \right\}.
\]
The first term in the previous expression is no larger than
\[
\mathbb{E}_{X,Y} \left[ \mathbb{P}_{Z} \left( \left| \Xi_n(h) - \xi(h,X,Y)  \right| > t \eta \Big | X,Y \right)  ; \mathcal{A}_{h,\epsilon} \right] + \mathbb{P}_{X,Y}\left( \left|\xi(h,X,Y)  - \xi(h) \right| > t (1 - \eta); \mathcal{A}_{h,\epsilon} \right),
\]
for any $\eta \in (0,1)$. 
We will first show that, if \eqref{eq:t} is satisfied,
\[
\mathbb{P}_{X,Y}\left( \left|\xi(h,X,Y)  - \xi(h) \right| > t (1 - \eta); \mathcal{A}_{h,\epsilon} \right) = 0.
\]
Indeed, first observe that
\[
\mathbb{E}_Z[\Xi_n(h) | X,Y] = P(\hat{L}_{h,X} \Delta \hat{L}_{h,Y})
\]
and that, on $\mathcal{A}_{h,\epsilon}$,
\begin{eqnarray*}
\hat{L}_{h,X}\Delta\hat{L}_{h,Y} & = & \{u \colon \hat p_{h,X}(u) >\lambda,\hat p_{h,Y}(u) \leq \lambda\}\cup\{u \colon \hat p_{h,X}(u) \leq \lambda, \hat p_{h,Y}(u) >\lambda\}\\
& \subseteq &  \{u \colon p_h(u) >\lambda-\epsilon, p_h(u) \leq \lambda+\epsilon\}\\
& = & \{u \colon |p_h(u) -\lambda| \leq \epsilon\} \\
& = & U_{h,\epsilon},
\end{eqnarray*}
Therefore,  on $\mathcal{A}_{h,\epsilon}$,
\begin{equation}\label{eq:xi.bound}
\xi(h,X,Y) = \mathbb{E}_Z[\Xi_n(h) | X,Y] \leq r_{h,\epsilon} \leq t(1 - \eta).
\end{equation}
By part 2 of Theorem \ref{thm:xi}, \eqref{eq:t} further implies that $t ( 1- \eta)   \geq \xi(h)$. As a result, on $\mathcal{A}_{h,\epsilon}$, $\left|\xi(h,X,Y)  - \xi(h) \right| \leq t ( 1- \eta) $, which yields 
\[
\mathbb{P}_{X,Y}\left( \left|\xi(h,X,Y)  - \xi(h) \right| > t ( 1- \eta) ; \mathcal{A}_{h,\epsilon} \right) = 0,
\]
as claimed.

We now proceed to bound from above 
\begin{equation}\label{eq:needthisone}
\mathbb{E}_{X,Y} \left[ \mathbb{P}_{Z} \left( \left| \Xi_n(h) - \xi(h,X,Y)  \right| > t \eta \Big | X,Y \right)  ; \mathcal{A}_{h,\epsilon} \right].
\end{equation}
Since 
\[
\Xi_n(h) = \frac{1}{n} \sum_{i =1}^n 1_{\{Z_i \in \hat{L}_{h,X}\Delta\hat{L}_{h,Y}\}},
\]
Bernstein's inequality \citep[see, for instance,][Proposition 2.9]{massart} yields that, for any $t>0$ and conditionally on $X$ and $Y$, 
\begin{equation}\label{eq:bernstein}
\mathbb{P}_Z \left( \left| \Xi_n(h) - \xi(h,X,Y) \right| > t \eta \Big | X,Y \right) \leq \exp \left\{ - 9 \sigma^2(X,Y,h)  g\left( \frac{n  t \eta}{ 3 \sigma^2(X,Y,h)} \right)\right\}
\end{equation}
where  $g(u) = 1 + u - \sqrt{1 + 2u}$ for all $u>0$, and 
\[
\sigma^2(X,Y,h) = \mathrm{Var}_Z[\Xi_n(h)|X,Y].
\] 
It is easy to see that
\[
 \sigma^2(X,Y,h) \leq \mathbb{E}_{Z} \left[  \Xi_n(h) |X,Y\right] = n \xi(h,X,Y) 
\]
and, therefore, restricting to the event $\mathcal{A}_{h,\epsilon}$, $\sigma^2(X,Y,h) \leq n t(1 - \eta)$, just like in \eqref{eq:xi.bound}.

Using the fact that $e^{-9x g\left( \frac{n  t }{ 3x} \right)}$ is increasing in $x$ for $x>0$, we conclude that, on the event $\mathcal{A}_{h,\epsilon}$, the right hand side of \eqref{eq:bernstein} is bounded from above by  
\[
\exp \left\{ - 9 n t (1 -\eta)  g\left( \frac{  \eta}{ 3 (1-\eta) } \right)\right\},
\]
which is independent of $X$ and $Y$. 
Thus, the previous expression is an upper bound for \eqref{eq:needthisone} and, therefore, for $\mathbb{P}_{X,Y,Z} \left( \left| \Xi_n(h) - \xi(h) \right| > t \right)$. The claim now follows from simple algebra.
\vspace{1cm}

\noindent
{\bf Proof of Theorem \ref{thm:instab.alpha}.}
\begin{enumerate}
\item The proof is almost the same as the proof of part 1 of Theorem \ref{thm:xi} and is therefore omitted.
\item Let $\mathcal{A}_{h,\tilde{\epsilon}}$ denote the event
\begin{equation}\label{eq:Atilde}
\max \left\{ ||\hat{p}_{h,X} - p_h||_\infty, |\lambda_{h,\alpha} - \hat{\lambda}_{h,\alpha,X}|, 
||\hat{p}_{h,Y} - p_h||_\infty, |\lambda_{h,\alpha} - \hat{\lambda}_{h,\alpha,Y}| \right\} \leq \tilde{\epsilon},
\end{equation}
where $\tilde{\epsilon} = \epsilon(A\kappa_3 + 1) + A\kappa_3/n$.
Then, using \eqref{eq:gine_prob}, \eqref{eq:alpha.conc} and the fact that $\epsilon < \tilde{\epsilon}$,  the union bound yields
\begin{equation}\label{eq:PA}
\mathbb{P}_{X,Y}(\mathcal{A}_{h,\tilde{\epsilon}}^c) \leq 4 K_1 e^{ - K_2 n h^d \epsilon^2} + 16 n e^{-n \epsilon^2/32} \equiv C(h,\epsilon,n)
\end{equation}
Now, on $\mathcal{A}_{h,\tilde{\epsilon}}$,
\begin{eqnarray*}
\{u:\hat{p}_{h,X}(u) > \hat{\lambda}_{h,\alpha,X}, \hat{p}_{h,Y}(u) \leq \hat{\lambda}_{h,\alpha,Y}\}
& \subseteq & \{u:p_h(u) > \hat{\lambda}_{h,\alpha,X}-\tilde{\epsilon}, p_h(u) \leq \hat{\lambda}_{h,\alpha,Y}+\tilde{\epsilon}\}\\
& \subseteq & \{u:p_h(u) > \lambda_{h,\alpha}-2\tilde{\epsilon}, p_h(u) \leq \lambda_{h,\alpha}+2\tilde{\epsilon}\}\\
& = & \{u:|p_h(u) - {\lambda}_{h,\alpha}| \leq 2\tilde{\epsilon}\}\\
& = & U_{h, \tilde{\epsilon},\alpha}.
\end{eqnarray*}
and therefore, $\mathrm{sign}(\widehat p_{h,X}(u)-\widehat \lambda_{h,\alpha,X}) = \mathrm{sign}(p_h(u)-\lambda_{h,\alpha})$ for all $u\notin U_{h,2\tilde\epsilon,\alpha}$.
Next, just  like in the proof of part 2 of theorem \ref{thm:xi}, using this fact and the result of the first part we can write
\begin{eqnarray*}
\xi(h,\alpha) & \leq &  \mathbb{E}_{X,Y,Z}[\hat{P}_Z(\{u \colon \hat p_{h,X}(u) > \hat{\lambda}_{h,\alpha,X}\}\Delta\{ u \colon \hat p_{h,Y}(u) > \hat{\lambda}_{h,\alpha,Y}\}) 1_{\mathcal{A}_{h,\tilde\epsilon}}(X,Y)] + \mathbb{P}_{X,Y}(\mathcal{A}_{h,\tilde{\epsilon}}^c)\\
& = & 2 \int_{\mathbb{R}^d} \mathbb{P}_{X,Y}(\{\hat{p}_{h,X}(u) > \hat{\lambda}_{h,\alpha,X}, \hat{p}_{h,Y}(u) \leq \hat{\lambda}_{h,\alpha,Y}\} \cap \mathcal{A}_{h,\tilde{\epsilon}}) dP(u) + \mathbb{P}_{X,Y}(\mathcal{A}_{h,\tilde{\epsilon}}^c)\\
& \leq & 2 \int_{U_{h,2\tilde{\epsilon},\alpha}} \mathbb{P}_{X,Y}(\{ \hat{p}_{h,X}(u) > \hat{\lambda}_{h,\alpha,X}, \hat{p}_{h,Y}(u) \leq \hat{\lambda}_{h,\alpha,Y}\} \cap \mathcal{A}_{h,\tilde{\epsilon}}) dP(u) + C(h,\epsilon,n)\\
& \leq & 2 \int_{U_{h,2\tilde{\epsilon},\alpha}} \mathbb{P}_{X,Y}(\hat{p}_{h,X}(u) > \hat{\lambda}_{h,\alpha,X}, \hat{p}_{h,Y}(u) \leq \hat{\lambda}_{h,\alpha,Y}) dP(u) +  C(h,\epsilon,n)\\
& \leq & 2 \int_{U_{h,2\tilde{\epsilon},\alpha}} \pi_{h,\alpha}(u)(1 - \pi_{h,\alpha}(u)) dP(u) +  C(h,\epsilon,n)\\
& \leq & P(U_{h,2\tilde{\epsilon},\alpha})\overline{A}_{h,\epsilon,\alpha} +  C(h,\epsilon,n).
\end{eqnarray*}

As for the lower bound, from the result of first part we obtain, trivially,
\[
\begin{array}{rcl}
\xi(h,\alpha) & \geq & 2\int_{U_{h,2\tilde{\epsilon},\alpha}} \pi_{h,\alpha}(u)(1 - \pi_{h,\alpha}(u)) dP(u)\\
& \geq & P(U_{h,2\tilde{\epsilon},\alpha}) \overline{A}_{h,\epsilon,\alpha}.\\
\end{array}
\]

\item To compute an upper bound for $\overline{A}_{h,\epsilon,\alpha}$
and a lower bound for $\underline{A}_{h,\epsilon,\alpha}$, we use
the Berry-Ess\'een bound and the stated assumptions. The proof is
very similar to the proof of lemma
\ref{prop:lambda.berry-esseen}, except that the result holds only on
the event $\mathcal{A}_{h,\tilde{\epsilon}}$. Therefore, we only
provide a sketch of the arguments.

The assumptions that $\tilde{\epsilon} \leq \inf_h
\frac{\lambda_{\alpha,h}}{4}$, implies that, for any $u \in
U_{h,2\tilde{\epsilon},\alpha}$,
\[
\frac{1}{h^d} \frac{\delta \lambda_{\alpha,h}}{2 v_d} \leq \frac{\delta (\lambda_{\alpha,h} - 2\tilde{\epsilon})}{h^d v_d} \leq \sigma^2(u,h) \leq \frac{(\lambda_{\alpha,h} + 2\tilde{\epsilon})}{h^d v_d} \leq \frac{1}{h^d}\frac{3 \lambda_{\alpha,h}}{2 v_d}.
\]
Because of this and the fact that, on $\mathcal{A}_{h,\tilde{\epsilon}}$, $|p_h(u) - \hat{\lambda}_{h,\alpha,X}| \leq 3 \tilde{\epsilon}$ for all $u \in U_{h,2\tilde{\epsilon},\alpha}$, the same Berry-Esseen arguments used in the proof of lemma \ref{prop:lambda.berry-esseen}  yield
\[
1 - \Phi\left( \frac{ 3 \tilde{\epsilon} \sqrt{nh^d} }{a_1}\right) - \frac{C(\delta,\lambda_{h,\alpha})}{\sqrt{nh^d }} \leq
\pi_{h,\alpha,\tilde{\epsilon}}(u) \leq
1 - \Phi\left( -\frac{ 3 \tilde{\epsilon} \sqrt{nh^d} }{a_2}\right) + \frac{C(\delta,\lambda_{h,\alpha})}{\sqrt{n h^d}}.
\]
where $\pi_{h,\alpha,\tilde{\epsilon}}(u) = \mathbb{P}_X\left( \{ \hat{p}_{h,X}(u) > \hat{\lambda}_{h,\alpha,X}\} \cap  \mathcal{A}_{h,\tilde{\epsilon}}\right)$, $a_1 = \delta\lambda_{h,\alpha}/(2v_d)$, $a_2=3\lambda_{h,\alpha}/(2v_d)$, and $C(\delta,\lambda_{h,\alpha}) = \frac{33}{4}\sqrt{\frac{2}{ \delta v_d \lambda_{h,\alpha}}}$.
Now notice that
\[
\pi_{h,\alpha}(u) \geq \pi_{h,\alpha,\tilde{\epsilon}}(u) \geq 1 - \Phi\left( \frac{ 3 \tilde{\epsilon} \sqrt{nh^d} }{a_1}\right) - \frac{C(\delta,\lambda_{h,\alpha})}{\sqrt{nh^d }}
\]
and
\[
\pi_{h,\alpha}(u) \leq \pi_{h,\alpha,\tilde{\epsilon}}(u) + P(\mathcal{A}_{h,\tilde{\epsilon}}^c) \leq 1 - \Phi\left( -\frac{ 3 \tilde{\epsilon} \sqrt{nh^d} }{a_2}\right) + \frac{C(\delta,\lambda_{h,\alpha})}{\sqrt{n h^d}} + C(h,\epsilon,n).
\]
where $C(h,\epsilon,n)$ is defined in \eqref{eq:PA}.
Therefore,
\[
\overline{A}_{h,\epsilon,\alpha} = 2\sup_{u\in U_{h,2\tilde{\epsilon},\alpha}} \pi_{h,\alpha}(u)(1- \pi_{h,\alpha}(u)) \leq 2\left(1 - \Phi\left( -\frac{ 3 \tilde{\epsilon} \sqrt{nh^d} }{a_2}\right) + \frac{C(\delta,\lambda_{h,\alpha})}{\sqrt{n h^d}} + C(h,\epsilon,n)\right)^2,
\]
and
\[
\underline{A}_{h,\epsilon,\alpha} = 2\inf_{u\in U_{h,2\tilde{\epsilon},\alpha}} \pi_{h,\alpha}(u)(1- \pi_{h,\alpha}(u)) \geq 2\left( 1 - \Phi\left( \frac{3 \tilde{\epsilon} \sqrt{n h^d}}{a_1} \right) - \frac{C(\delta,\lambda_{h,\alpha})}{\sqrt{n h^d}}- C(h,\epsilon,n)\right)^2.
\]
\end{enumerate}

\vspace{1cm}

\noindent
{\bf Proof of Theorem \ref{thm::total-var}.}
(1) Since the sample space is compact, $\mu(S) < \infty$, where $S$ denotes the support of $P$ 
and $\mu$ denotes the Lebesgue measure. Therefore, we obtain the inequality
\begin{eqnarray*}
\Gamma_n(h) &\leq &\frac{\mu(S)}{2} ||\hat{p}_{h,X}-\hat{p}_{h,Y}||_\infty
\leq 
\frac{\mu(S)}{2} ||\hat{p}_{h,X}-p_h||_\infty + \frac{\mu(S)}{2} ||\hat{p}_{h,Y}-p_h||_\infty\\
& \stackrel{d}{=} &
\mu(S) ||\hat{p}_{h,X}-p_h||_\infty.
\end{eqnarray*}
Next, let $C = \frac{(\mu(S))^2(a+2)}{K_2}$, so that for $n>K_1$ 
\[
t_h > \sqrt{\frac{\mu(S)^2 \log (n^{a+1}K_1)}{K_2 n h^d}}.
\]
Then,
\begin{eqnarray*}
\mathbb{P}_{X,Y}\left( \Gamma_n(h) > t_h \ \ {\rm for\ some\ }h\in {\cal H}_n \right) & \leq & 
\mathbb{P}_{X}\left(  ||\hat{p}_{h,X}-p_h||_\infty > \frac{t_h}{\mu(S)} \ \ {\rm for\ some\ }h\in {\cal H}_n \right)\\
& \leq & \sum_{h \in \mathcal{H}_n} \mathbb{P}_{X}\left(  ||\hat{p}_{h,X}-p_h||_\infty > \frac{t_h}{\mu(S)} \right)\\
& \leq & \sum_{h \in \mathcal{H}_n} K_1 \exp \{ - K_2 n t_h^2 h^d/(\mu(S)^2)  \}\\
& \leq & A n^a \frac{1}{n^{a + 1}} = \frac{A}{n}\\
& \leq & \delta,  
\end{eqnarray*}
where the third inequality stems from \eqref{eq:gine_prob} and the
assumption that $n$ is large enough, and the last inequality follows
from the assumed condition on $\delta$.

\vspace{.5cm}

\noindent
(2)
Consider any $h \leq h_*$.
Note that
$$
\Gamma_n(h) \geq \Gamma_{n,S}(h) \equiv \frac{1}{2}\int_S | \hat{p}_{h,X}(u)- \hat{p}_{h,Y}(u)| du.
$$
Let
$$
D(u) =\sqrt{n h^d}(\hat{p}_{h,X}(u) - \hat{p}_{h,Y}(u)).
$$
The variance of $D(u)$
is
\begin{eqnarray*}
{\rm Var}\left( \sqrt{n h^d}(\hat{p}_{h,X}(u) - \hat{p}_{h,Y}(u))\right) &=&
n h^d \left( {\rm Var} (\hat{p}_{h,X}(u)) + {\rm Var} (\hat{p}_{h,Y}(u))\right)\\
&=&
2n h^d  {\rm Var} (\hat{p}_{h,X}(u)) \\
&=&
2n h^d  {\rm Var} \left(\frac{1}{n h^d v_d }\sum_{i=1}^n I(||X_i-u||\leq h)\right)\\
&=& \frac{2 n^2 h^d }{n^2 h^{2d} v_d^2}{\rm Var} ( I(||X_i -u|| \leq h))\\
&=& \frac{2}{v_d^2 h^d} P(B(u,h))(1-P(B(u,h))).
\end{eqnarray*}
Now, for $u\in S$, by \eqref{eq:a1a2},
$$
P(B(u,h))(1-P(B(u,h))) \leq  P(B(u,h)) \leq a_2 h^d v_d
$$
and
$$
P(B(u,h))(1-P(B(u,h))) \geq  P(B(u,h))\delta \geq a_1 h^d v_d \delta.
$$
Hence,
$$
2 a_1 v_d \delta \leq 
{\rm Var}(D(u)) \leq 2 a_2 v_d, \quad \forall u \in S,
$$
which shows that the variance of $D(u)$ is bounded above and below by positive functions that do not depend on $h$.
By a similar calculation,
${\rm Cov}(D(u),D(v))$
is bounded above and below by functions that do not depend on $h$, for all $u,v \in S$.

Now, for any $u$,
$$
D(u) = D_1(u) - D_2(u) \equiv \sqrt{n h^d}(P_n - P)(f_u) - \sqrt{n h^d}(Q_n - P)(f_u)
$$
where
$P_n$ is the empirical measure based on $X_1,\ldots, X_n$,
$Q_n$ is the empirical measure based on $Y_1,\ldots, Y_n$,
and
$f_u (\cdot)= h^{-d}K(||u-\cdot||/h)$.
Note that $D_1$ and $D_2$ are independent, mean 0 stochastic processes.
We can regard
$\{ \sqrt{n h^d} (P_n - P) (f):\ f\in {\cal F}\}$
as an empirical process, where
${\cal F}=\{f_u:\ u\in S\}$
and similarly for
$\{ \sqrt{n h^d} (Q_n - P) (f):\ f\in {\cal F}\}$.
For fixed $h$,
the collection ${\cal F}$ is a Donsker class.
Hence, for every $u \in S$,
$D_1(u)$ and $D_2(u)$ converge to two independent mean 0 Gaussian processes.
By the continuous mapping theorem, for every $u \in S$,
$D(u)$ converges to a mean 0 Gaussian process $\mathbb{G}$ with
some covariance kernel $\kappa$.
By the calculations above, there exist positive bounded functions
$r(u,v)\leq s(u,v)$ such that
$r(u,v) \leq \kappa(u,v) \leq s(u,v)$ and such that
neither $r$ nor $s$ depend on $h$.
Hence
\begin{eqnarray*}
\mathbb{P}_{X,Y}\left( \Gamma_n(h) \geq t \sqrt{\frac{1 }{n h^d}}\right) & \geq &
\mathbb{P}_{X,Y}\left( \Gamma_{n,S}(h) \geq t \sqrt{\frac{1 }{n h^d}}\right)=
\mathbb{P}_{X,Y}\left( \sqrt{n h^d}\Gamma_{n,S}(h) \geq t \right)\\
&=&
\mathbb{P}_{X,Y}\left( \frac1{2}\int_S |D(u)|du \geq t \right)\\
&=& \mathbb{P}\left( \frac{1}{2}\int|\mathbb{G}(u)|du \geq t \right) + o(1),
\end{eqnarray*}
where the last probability is the law of the Gaussian process $\mathbb{G}$. 
Since $\mathbb{G}$ has strictly positive variance,
$\mathbb{P}\left( \int|\mathbb{G}| \geq 0 \right)=1$.
Clearly,
$\mathbb{P}\left( \int|\mathbb{G}| \geq 2t \right)$ is decreasing in $t$.
Hence, for each $\delta$, there is a positive $t$ such that
$\mathbb{P}\left( \frac1{2}\int|\mathbb{G}| \geq t \right) \geq 1-\delta/2$.

\vspace{.5cm}

\noindent
(3) The proof of this part is straightforward and is omitted.

\section*{Acknowledgments}

Research supported by
NSF grant CCF-0625879, NSF grant DMS-0631589
and AFOSR contract
FA9550-09-1-0373.

\vskip 0.2in
\bibliography{Rinaldo_Singh_Nugent_Wasserman}

\end{document}